%% file: rlscope_mlsys2021.tex
\documentclass{article}

\input{tex/config}
\input{tex/custom_imports}

\RequirePackage{microtype}
\RequirePackage{graphicx}
\RequirePackage{booktabs} 

\RequirePackage{hyperref}

\input{tex/custom_commands}
\ifx\ShowAuthors\undefined
\RequirePackage{styles/mlsys2021_style/mlsys2021}
\else
\RequirePackage[accepted]{styles/mlsys2021_style/mlsys2021}
\fi 

\mlsystitlerunning{\TitleRunning}

\begin{document}

\input{tex/post_begin_document}

\twocolumn[
\mlsystitle{\PaperTitle}



\mlsyssetsymbol{equal}{*}

\begin{mlsysauthorlist}
\mlsysauthor{James Gleeson}{vector,uoft}
\mlsysauthor{Srivatsan Krishnan}{harvard}
\mlsysauthor{Moshe Gabel}{uoft}
\mlsysauthor{Vijay Janapa Reddi}{harvard}
\mlsysauthor{Eyal de Lara}{uoft}
\mlsysauthor{Gennady Pekhimenko}{vector,uoft}

\end{mlsysauthorlist}

\mlsysaffiliation{vector}{Vector Institute}
\mlsysaffiliation{uoft}{Department of Computer Science, University of Toronto, Toronto, Ontario, Canada}
\mlsysaffiliation{harvard}{School of Engineering and Applied Sciences, Harvard, Cambridge, Massachusetts, USA}

\mlsyscorrespondingauthor{James Gleeson}{jgleeson@cs.toronto.edu}

\mlsyskeywords{Machine Learning, MLSys}

\vskip 0.3in

\input{tex/abstract}
]



\ifx\ShowAuthors\undefined
\else
\printAffiliationsAndNotice{}  
\fi

\input{tex/intro}
\input{tex/background}

\input{tex/rlscope_profiler}

\input{tex/case_studies}

\input{tex/related_work}

\input{tex/conclusion}

\input{tex/acknowledgements}

\label{page:last-page}


\bibliography{rlscope_mlsys2021}
\bibliographystyle{styles/mlsys2021_style/mlsys2021}

\appendix
\input{tex/appendix}


\end{document}

%% file: tex/config.tex
\def\RebuttalEffectOfOverheadCorrection{1}

\def\PaperFormatWithAppendix{1}

\def\ShowAuthors{1}

%% file: tex/custom_commands.tex
\newcommand{\rlscope}{RL-Scope\xspace}
\newcommand{\autograph}{Autograph\xspace}
\newcommand{\graph}{Graph\xspace}
\newcommand{\eager}{Eager\xspace}
\newcommand{\minigo}{Minigo\xspace}

\newcommand{\githubURL}{\url{https://github.com/UofT-EcoSystem/rlscope}}
\newcommand{\ecosys}{\href{http://www.cs.toronto.edu/ecosystem}{EcoSys group}\xspace}

\ifx\TexMetricsUncorrected\undefined %
\newcommand{\PaperTitle}{\rlscope: Cross-Stack Profiling for\\Deep Reinforcement Learning Workloads}
\newcommand{\TitleRunning}{\rlscope: Cross-Stack Profiling for Deep Reinforcement Learning Workloads}
\else %
\newcommand{\PaperTitle}{(DBG) \rlscope: Cross-Stack Profiling for\\Deep Reinforcement Learning Workloads}
\newcommand{\TitleRunning}{(DBG) \rlscope: Cross-Stack Profiling for Deep Reinforcement Learning Workloads}
\fi

\newcommand{\CiteExtra}[1]{\xspace}

\newcommand{\librlscope}{\texttt{librlscope.so}\xspace}

\newcommand{\asPercent}[1]{\pgfmathparse{100*#1}\num[round-mode=places,round-precision=1]{\pgfmathresult}\%}

\newcommand{\asPercentNoDecimal}[1]{\pgfmathparse{100*#1}\num[round-mode=places,round-precision=0]{\pgfmathresult}\%}

\NewDocumentCommand{\codeword}{v}{\texttt{#1}}

\newcommand{\irect}[1]{%
  \tikz[baseline=(char.base)]\node[anchor=south west, draw,rectangle, rounded corners=0.5mm, inner sep=1.5pt, minimum size=3mm,outer xsep=0pt,column sep=0pt,row sep=0pt,
    text height=2.2mm](char){#1} ;}

\newcounter{FndQualCounter}
\newcommand{\FndQual}[1]{%
  \refstepcounter{FndQualCounter}
  \label{#1}
  \irect{\textit{Q}.\theFndQualCounter}\xspace
}

\newcounter{FndSurpCounter}
\newcommand{\FndSurp}[1]{%
  \refstepcounter{FndSurpCounter}
  \label{#1}
  \irect{\textit{S}.\theFndSurpCounter}\xspace
}

\newcounter{FndCounter}
\newcommand{\Fnd}[1]{%
  \refstepcounter{FndCounter}
  \label{#1}
  \irect{\textit{F}.\theFndCounter}\xspace
}
\newcommand{\RefFnd}[1]{%
  \irect{\hyperref[#1]{\textit{F}.\ref{#1}}}\xspace
}

\newcommand{\IntroRefFnd}[1]{}

\newcommand{\EvalRefFnd}[1]{}

\newcommand{\BoxGpu}{\legendbox{green}{GPU}}
\newcommand{\BoxCpu}{\legendbox{red}{CPU}}
\newcommand{\BoxCpuPlusGpu}{\legendbox{blue}{CPU + GPU}}

\newenvironment{rlscope-finding-qual}[1]
{
    \begin{tcolorbox}[size=fbox]
    \FndQual{#1} 
    \bfseries
}
{
    \end{tcolorbox}
}

\newenvironment{rlscope-finding-surp}[1]
{
    \begin{tcolorbox}[size=fbox]
    \FndSurp{#1} 
    \bfseries
}
{
    \end{tcolorbox}
}

\newenvironment{rlscope-finding}[1]
{
    \begin{tcolorbox}[size=fbox,after skip=0.4\baselineskip]
    \Fnd{#1} 
    \bfseries
}
{
    \end{tcolorbox}
}

\newtcbox{\legendbox}[1]{on line, colback=#1!10!white, colframe=#1!50!black, size=fbox, fontupper=\sffamily\scriptsize, arc=0mm, boxsep=2pt}

\newtcbox{\barbox}[1]{on line, colback=#1!30!white, colframe=#1!50!black, size=fbox, fontupper=\sffamily\scriptsize, arc=0mm, boxsep=2pt}

%% file: tex/post_begin_document.tex
\def\TexMetricsEnabled{1}

\input{tex/autogenerated/ddpg_consecutive_steps_1000/FrameworkChoiceMetrics.manual}

\input{tex/autogenerated/uncorrected/FrameworkChoiceMetrics.uncorrected}

\input{tex/autogenerated/uncorrected/AlgoChoiceMetrics.uncorrected}

\input{tex/autogenerated/uncorrected/FrameworkChoiceMetricsUncorrected}

\ifx\TexMetricsUncorrected\undefined %
\input{tex/autogenerated/FrameworkChoiceMetrics}

\input{tex/autogenerated/AlgoChoiceMetrics}

\newcommand{\AlgorithmChoicePdf}{fig/12_algorithm_choice.pdf}
\newcommand{\FrameworkChoiceTdPdf}{fig/framework_choice.pdf}
\newcommand{\FrameworkChoiceDdpgPdf}{fig/framework_choice_ddpg.pdf}

\else %
\input{tex/autogenerated/uncorrected/FrameworkChoiceMetrics}

\input{tex/autogenerated/uncorrected/AlgoChoiceMetrics}

\newcommand{\AlgorithmChoicePdf}{fig/uncorrected/12_algorithm_choice.pdf}
\newcommand{\FrameworkChoiceTdPdf}{fig/uncorrected/framework_choice.pdf}
\newcommand{\FrameworkChoiceDdpgPdf}{fig/uncorrected/framework_choice_ddpg.pdf}

\fi

\ifx\TexMetricsEnabled\undefined
    \newcommand{\FCAsTimes}[1]{$#1\times$}
    \newcommand{\FCAsPercent}[1]{$#1\%$}
\else
    \newcommand{\FCAsTimes}[1]{\pgfmathparse{#1}\num[round-mode=places,round-precision=1]{\pgfmathresult}$\times$}
    \newcommand{\FCAsPercent}[1]{\pgfmathparse{100*#1}\num[round-mode=places,round-precision=1]{\pgfmathresult}\%}
\fi

\input{tex/hardcoded_numbers}

%% file: tex/autogenerated/ddpg_consecutive_steps_1000/FrameworkChoiceMetrics.manual.tex
\ifx\TexMetricsEnabled\undefined

  \newcommand{\DDPGHyperParamFindSurpAutographInflatesPythonMaxRatioSimulationPythonAutographToEager}{a}
  \newcommand{\DDPGHyperParamFindSurpAutographInflatesPythonMeanRatioSimulationPythonAutographToEagerDDPG}{b}
  \newcommand{\DDPGHyperParamFindSurpAutographInflatesPythonMeanRatioSimulationPythonAutographToEagerTD}{c}

\else

  \newcommand{\DDPGHyperParamFindSurpAutographInflatesPythonMaxRatioSimulationPythonAutographToEager}{1.6228075349881232}
  \newcommand{\DDPGHyperParamFindSurpAutographInflatesPythonMeanRatioSimulationPythonAutographToEagerDDPG}{1.148768086483055}
  \newcommand{\DDPGHyperParamFindSurpAutographInflatesPythonMeanRatioSimulationPythonAutographToEagerTD}{1.6228075349881232}

\fi

%% file: tex/autogenerated/uncorrected/FrameworkChoiceMetrics.uncorrected.tex
\ifx\TexMetricsEnabled\undefined
  \newcommand{\UncorrectedFindQualEagerMoreTransMinEagerSlowdown}{a}
  \newcommand{\UncorrectedFindQualEagerMoreTransMaxEagerSlowdown}{b}
  \newcommand{\UncorrectedFindQualEagerMoreTransMinRatioInferenceFrameworkTransTFToPyTorch}{c}
  \newcommand{\UncorrectedFindQualEagerMoreTransMaxRatioInferenceFrameworkTransTFToPyTorch}{d}
  \newcommand{\UncorrectedFindQualEagerMoreTransMeanRatioInferenceFrameworkTransTFToPyTorch}{e}
  \newcommand{\UncorrectedFindQualEagerMoreTransGeomeanRatioInferenceFrameworkTransTFToPyTorch}{f}
  \newcommand{\UncorrectedFindQualEagerMoreTransStdRatioInferenceFrameworkTransTFToPyTorch}{g}
  \newcommand{\UncorrectedFindQualEagerMoreTransMinRatioBackpropagationFrameworkTransTFToPyTorch}{h}
  \newcommand{\UncorrectedFindQualEagerMoreTransMaxRatioBackpropagationFrameworkTransTFToPyTorch}{i}
  \newcommand{\UncorrectedFindQualEagerMoreTransMeanRatioBackpropagationFrameworkTransTFToPyTorch}{j}
  \newcommand{\UncorrectedFindQualEagerMoreTransGeomeanRatioBackpropagationFrameworkTransTFToPyTorch}{k}
  \newcommand{\UncorrectedFindQualEagerMoreTransStdRatioBackpropagationFrameworkTransTFToPyTorch}{l}
  \newcommand{\UncorrectedFindQualEagerMoreTransMinRatioInferenceFrameworkTFToPyTorch}{m}
  \newcommand{\UncorrectedFindQualEagerMoreTransMaxRatioInferenceFrameworkTFToPyTorch}{n}
  \newcommand{\UncorrectedFindQualEagerMoreTransMeanRatioInferenceFrameworkTFToPyTorch}{o}
  \newcommand{\UncorrectedFindQualEagerMoreTransGeomeanRatioInferenceFrameworkTFToPyTorch}{p}
  \newcommand{\UncorrectedFindQualEagerMoreTransStdRatioInferenceFrameworkTFToPyTorch}{q}
  \newcommand{\UncorrectedFindQualEagerMoreTransMinRatioBackpropagationFrameworkTFToPyTorch}{r}
  \newcommand{\UncorrectedFindQualEagerMoreTransMaxRatioBackpropagationFrameworkTFToPyTorch}{s}
  \newcommand{\UncorrectedFindQualEagerMoreTransMeanRatioBackpropagationFrameworkTFToPyTorch}{t}
  \newcommand{\UncorrectedFindQualEagerMoreTransGeomeanRatioBackpropagationFrameworkTFToPyTorch}{u}
  \newcommand{\UncorrectedFindQualEagerMoreTransStdRatioBackpropagationFrameworkTFToPyTorch}{v}
  \newcommand{\UncorrectedFindQualAutographReducesPythonMinAutographPythonPercentOfOp}{a}
  \newcommand{\UncorrectedFindQualAutographReducesPythonMaxAutographPythonPercentOfOp}{b}
  \newcommand{\UncorrectedFindQualAutographReducesPythonMinGraphPythonPercentOfOp}{c}
  \newcommand{\UncorrectedFindQualAutographReducesPythonMaxGraphPythonPercentOfOp}{d}
  \newcommand{\UncorrectedFindQualAutographReducesPythonMinRatioPythonBackpropagationDDPG}{e}
  \newcommand{\UncorrectedFindQualAutographReducesPythonMaxRatioPythonBackpropagationDDPG}{f}
  \newcommand{\UncorrectedFindQualAutographReducesPythonMeanRatioPythonBackpropagationDDPG}{g}
  \newcommand{\UncorrectedFindQualAutographReducesPythonGeomeanRatioPythonBackpropagationDDPG}{h}
  \newcommand{\UncorrectedFindQualAutographReducesPythonStdRatioPythonBackpropagationDDPG}{i}
  \newcommand{\UncorrectedFindQualAutographReducesPythonMinRatioPythonInferenceDDPG}{j}
  \newcommand{\UncorrectedFindQualAutographReducesPythonMaxRatioPythonInferenceDDPG}{k}
  \newcommand{\UncorrectedFindQualAutographReducesPythonMeanRatioPythonInferenceDDPG}{l}
  \newcommand{\UncorrectedFindQualAutographReducesPythonGeomeanRatioPythonInferenceDDPG}{m}
  \newcommand{\UncorrectedFindQualAutographReducesPythonStdRatioPythonInferenceDDPG}{n}
  \newcommand{\UncorrectedFindQualAutographReducesPythonMinRatioPythonBackpropagationTD}{o}
  \newcommand{\UncorrectedFindQualAutographReducesPythonMaxRatioPythonBackpropagationTD}{p}
  \newcommand{\UncorrectedFindQualAutographReducesPythonMeanRatioPythonBackpropagationTD}{q}
  \newcommand{\UncorrectedFindQualAutographReducesPythonGeomeanRatioPythonBackpropagationTD}{r}
  \newcommand{\UncorrectedFindQualAutographReducesPythonStdRatioPythonBackpropagationTD}{s}
  \newcommand{\UncorrectedFindQualAutographReducesPythonMinRatioPythonInferenceTD}{t}
  \newcommand{\UncorrectedFindQualAutographReducesPythonMaxRatioPythonInferenceTD}{u}
  \newcommand{\UncorrectedFindQualAutographReducesPythonMeanRatioPythonInferenceTD}{v}
  \newcommand{\UncorrectedFindQualAutographReducesPythonGeomeanRatioPythonInferenceTD}{w}
  \newcommand{\UncorrectedFindQualAutographReducesPythonStdRatioPythonInferenceTD}{x}
  \newcommand{\UncorrectedFindSurpDdpgBackpropSlowMeanRatioAutographToGraphBackpropagationDDPG}{a}
  \newcommand{\UncorrectedFindSurpDdpgBackpropSlowMeanRatioAutographToGraphInferenceDDPG}{b}
  \newcommand{\UncorrectedFindSurpDdpgBackpropSlowMeanRatioAutographToGraphBackpropagationTD}{c}
  \newcommand{\UncorrectedFindSurpDdpgBackpropSlowMeanRatioAutographToGraphInferenceTD}{d}
  \newcommand{\UncorrectedFindSurpDdpgBackpropSlowMeanRatioAutographToGraphCUDABackpropagationDDPG}{e}
  \newcommand{\UncorrectedFindSurpDdpgBackpropSlowMeanRatioAutographToGraphPythonBackpropagationDDPG}{f}
  \newcommand{\UncorrectedFindSurpDdpgBackpropSlowMeanRatioAutographToGraphCUDAInferenceDDPG}{g}
  \newcommand{\UncorrectedFindSurpDdpgBackpropSlowMeanRatioAutographToGraphPythonInferenceDDPG}{h}
  \newcommand{\UncorrectedFindSurpDdpgBackpropSlowMeanRatioAutographToGraphCUDABackpropagationTD}{i}
  \newcommand{\UncorrectedFindSurpDdpgBackpropSlowMeanRatioAutographToGraphPythonBackpropagationTD}{j}
  \newcommand{\UncorrectedFindSurpDdpgBackpropSlowMeanRatioAutographToGraphCUDAInferenceTD}{k}
  \newcommand{\UncorrectedFindSurpDdpgBackpropSlowMeanRatioAutographToGraphPythonInferenceTD}{l}
  \newcommand{\UncorrectedFindQualPytorchEagerBetterMinPyTorchEagerSpeedup}{a}
  \newcommand{\UncorrectedFindQualPytorchEagerBetterMaxPyTorchEagerSpeedup}{b}
  \newcommand{\UncorrectedFindQualAutographGraphSimilarMaxAutographSpeedup}{a}
  \newcommand{\UncorrectedFindQualAutographGraphSimilarMaxGraphSpeedup}{b}
  \newcommand{\UncorrectedFindQualAutographGraphSimilarMaxSpeedup}{c}
  \newcommand{\UncorrectedFindSurpExecModelComparisonMinTFSpeedup}{a}
  \newcommand{\UncorrectedFindSurpExecModelComparisonMaxTFSpeedup}{b}
  \newcommand{\UncorrectedFindSurpTotalGpuTimeMinGPUPercent}{a}
  \newcommand{\UncorrectedFindSurpTotalGpuTimeMaxGPUPercent}{b}
  \newcommand{\UncorrectedFindSurpCudaApiDominatesMinRatioCUDAToGPU}{a}
  \newcommand{\UncorrectedFindSurpCudaApiDominatesMaxRatioCUDAToGPU}{b}
  \newcommand{\UncorrectedFindSurpCudaApiDominatesMeanRatioCUDAToGPU}{c}
  \newcommand{\UncorrectedFindSurpCudaApiDominatesGeomeanRatioCUDAToGPU}{d}
  \newcommand{\UncorrectedFindSurpCudaApiDominatesStdRatioCUDAToGPU}{e}
  \newcommand{\UncorrectedFindSurpAutographNoGpuMinRatioGPUAutographToNonautograph}{a}
  \newcommand{\UncorrectedFindSurpAutographNoGpuMaxRatioGPUAutographToNonautograph}{b}
  \newcommand{\UncorrectedFindSurpAutographNoGpuMeanRatioGPUAutographToNonautograph}{c}
  \newcommand{\UncorrectedFindSurpAutographNoGpuGeomeanRatioGPUAutographToNonautograph}{d}
  \newcommand{\UncorrectedFindSurpAutographNoGpuStdRatioGPUAutographToNonautograph}{e}
  \newcommand{\UncorrectedFindSurpAutographInflatesInferenceMinRatioInferenceFrameworkAutographToGraph}{a}
  \newcommand{\UncorrectedFindSurpAutographInflatesInferenceMaxRatioInferenceFrameworkAutographToGraph}{b}
  \newcommand{\UncorrectedFindSurpAutographInflatesInferenceMeanRatioInferenceFrameworkAutographToGraph}{c}
  \newcommand{\UncorrectedFindSurpAutographInflatesInferenceGeomeanRatioInferenceFrameworkAutographToGraph}{d}
  \newcommand{\UncorrectedFindSurpAutographInflatesInferenceStdRatioInferenceFrameworkAutographToGraph}{e}
  \newcommand{\UncorrectedFindSurpAutographInflatesInferenceMeanRatioInferenceFrameworkDDPGAutographToGraph}{f}
  \newcommand{\UncorrectedFindSurpAutographInflatesInferenceMeanRatioInferenceFrameworkTDAutographToGraph}{g}
  \newcommand{\UncorrectedFindSurpAutographInflatesPythonMaxRatioSimulationPythonAutographToEager}{a}
  \newcommand{\UncorrectedFindSurpAutographInflatesPythonMeanRatioSimulationPythonAutographToEagerDDPG}{b}
  \newcommand{\UncorrectedFindSurpAutographInflatesPythonMeanRatioSimulationPythonAutographToEagerTD}{c}
\else
  \newcommand{\UncorrectedFindQualEagerMoreTransMinEagerSlowdown}{1.7305056439463715}
  \newcommand{\UncorrectedFindQualEagerMoreTransMaxEagerSlowdown}{4.598368342263544}
  \newcommand{\UncorrectedFindQualEagerMoreTransMinRatioInferenceFrameworkTransTFToPyTorch}{3.155013832244816}
  \newcommand{\UncorrectedFindQualEagerMoreTransMaxRatioInferenceFrameworkTransTFToPyTorch}{3.155013832244816}
  \newcommand{\UncorrectedFindQualEagerMoreTransMeanRatioInferenceFrameworkTransTFToPyTorch}{3.155013832244816}
  \newcommand{\UncorrectedFindQualEagerMoreTransGeomeanRatioInferenceFrameworkTransTFToPyTorch}{3.155013832244816}
  \newcommand{\UncorrectedFindQualEagerMoreTransStdRatioInferenceFrameworkTransTFToPyTorch}{nan}
  \newcommand{\UncorrectedFindQualEagerMoreTransMinRatioBackpropagationFrameworkTransTFToPyTorch}{1.599702380952381}
  \newcommand{\UncorrectedFindQualEagerMoreTransMaxRatioBackpropagationFrameworkTransTFToPyTorch}{1.599702380952381}
  \newcommand{\UncorrectedFindQualEagerMoreTransMeanRatioBackpropagationFrameworkTransTFToPyTorch}{1.599702380952381}
  \newcommand{\UncorrectedFindQualEagerMoreTransGeomeanRatioBackpropagationFrameworkTransTFToPyTorch}{1.599702380952381}
  \newcommand{\UncorrectedFindQualEagerMoreTransStdRatioBackpropagationFrameworkTransTFToPyTorch}{nan}
  \newcommand{\UncorrectedFindQualEagerMoreTransMinRatioInferenceFrameworkTFToPyTorch}{11.762905935923435}
  \newcommand{\UncorrectedFindQualEagerMoreTransMaxRatioInferenceFrameworkTFToPyTorch}{11.762905935923435}
  \newcommand{\UncorrectedFindQualEagerMoreTransMeanRatioInferenceFrameworkTFToPyTorch}{11.762905935923435}
  \newcommand{\UncorrectedFindQualEagerMoreTransGeomeanRatioInferenceFrameworkTFToPyTorch}{11.762905935923435}
  \newcommand{\UncorrectedFindQualEagerMoreTransStdRatioInferenceFrameworkTFToPyTorch}{nan}
  \newcommand{\UncorrectedFindQualEagerMoreTransMinRatioBackpropagationFrameworkTFToPyTorch}{3.206866615000169}
  \newcommand{\UncorrectedFindQualEagerMoreTransMaxRatioBackpropagationFrameworkTFToPyTorch}{3.206866615000169}
  \newcommand{\UncorrectedFindQualEagerMoreTransMeanRatioBackpropagationFrameworkTFToPyTorch}{3.206866615000169}
  \newcommand{\UncorrectedFindQualEagerMoreTransGeomeanRatioBackpropagationFrameworkTFToPyTorch}{3.206866615000169}
  \newcommand{\UncorrectedFindQualEagerMoreTransStdRatioBackpropagationFrameworkTFToPyTorch}{nan}
  \newcommand{\UncorrectedFindQualAutographReducesPythonMinAutographPythonPercentOfOp}{0.05237954768430697}
  \newcommand{\UncorrectedFindQualAutographReducesPythonMaxAutographPythonPercentOfOp}{0.06309627838641839}
  \newcommand{\UncorrectedFindQualAutographReducesPythonMinGraphPythonPercentOfOp}{0.19297854945330573}
  \newcommand{\UncorrectedFindQualAutographReducesPythonMaxGraphPythonPercentOfOp}{0.5632848879878798}
  \newcommand{\UncorrectedFindQualAutographReducesPythonMinRatioPythonBackpropagationDDPG}{11.547763677104035}
  \newcommand{\UncorrectedFindQualAutographReducesPythonMaxRatioPythonBackpropagationDDPG}{11.547763677104035}
  \newcommand{\UncorrectedFindQualAutographReducesPythonMeanRatioPythonBackpropagationDDPG}{11.547763677104035}
  \newcommand{\UncorrectedFindQualAutographReducesPythonGeomeanRatioPythonBackpropagationDDPG}{11.547763677104035}
  \newcommand{\UncorrectedFindQualAutographReducesPythonStdRatioPythonBackpropagationDDPG}{nan}
  \newcommand{\UncorrectedFindQualAutographReducesPythonMinRatioPythonInferenceDDPG}{2.584217546193086}
  \newcommand{\UncorrectedFindQualAutographReducesPythonMaxRatioPythonInferenceDDPG}{2.584217546193086}
  \newcommand{\UncorrectedFindQualAutographReducesPythonMeanRatioPythonInferenceDDPG}{2.584217546193086}
  \newcommand{\UncorrectedFindQualAutographReducesPythonGeomeanRatioPythonInferenceDDPG}{2.584217546193086}
  \newcommand{\UncorrectedFindQualAutographReducesPythonStdRatioPythonInferenceDDPG}{nan}
  \newcommand{\UncorrectedFindQualAutographReducesPythonMinRatioPythonBackpropagationTD}{3.546300417853453}
  \newcommand{\UncorrectedFindQualAutographReducesPythonMaxRatioPythonBackpropagationTD}{3.546300417853453}
  \newcommand{\UncorrectedFindQualAutographReducesPythonMeanRatioPythonBackpropagationTD}{3.546300417853453}
  \newcommand{\UncorrectedFindQualAutographReducesPythonGeomeanRatioPythonBackpropagationTD}{3.5463004178534536}
  \newcommand{\UncorrectedFindQualAutographReducesPythonStdRatioPythonBackpropagationTD}{nan}
  \newcommand{\UncorrectedFindQualAutographReducesPythonMinRatioPythonInferenceTD}{3.8988723680744646}
  \newcommand{\UncorrectedFindQualAutographReducesPythonMaxRatioPythonInferenceTD}{3.8988723680744646}
  \newcommand{\UncorrectedFindQualAutographReducesPythonMeanRatioPythonInferenceTD}{3.8988723680744646}
  \newcommand{\UncorrectedFindQualAutographReducesPythonGeomeanRatioPythonInferenceTD}{3.898872368074465}
  \newcommand{\UncorrectedFindQualAutographReducesPythonStdRatioPythonInferenceTD}{nan}
  \newcommand{\UncorrectedFindSurpDdpgBackpropSlowMeanRatioAutographToGraphBackpropagationDDPG}{3.1455914542436902}
  \newcommand{\UncorrectedFindSurpDdpgBackpropSlowMeanRatioAutographToGraphInferenceDDPG}{0.7014258660919}
  \newcommand{\UncorrectedFindSurpDdpgBackpropSlowMeanRatioAutographToGraphBackpropagationTD}{0.8619596568959995}
  \newcommand{\UncorrectedFindSurpDdpgBackpropSlowMeanRatioAutographToGraphInferenceTD}{0.39768689368950083}
  \newcommand{\UncorrectedFindSurpDdpgBackpropSlowMeanRatioAutographToGraphCUDABackpropagationDDPG}{2.389004121179645}
  \newcommand{\UncorrectedFindSurpDdpgBackpropSlowMeanRatioAutographToGraphPythonBackpropagationDDPG}{11.547763677104035}
  \newcommand{\UncorrectedFindSurpDdpgBackpropSlowMeanRatioAutographToGraphCUDAInferenceDDPG}{0.7730556223566809}
  \newcommand{\UncorrectedFindSurpDdpgBackpropSlowMeanRatioAutographToGraphPythonInferenceDDPG}{2.584217546193086}
  \newcommand{\UncorrectedFindSurpDdpgBackpropSlowMeanRatioAutographToGraphCUDABackpropagationTD}{0.8817469934175527}
  \newcommand{\UncorrectedFindSurpDdpgBackpropSlowMeanRatioAutographToGraphPythonBackpropagationTD}{3.546300417853453}
  \newcommand{\UncorrectedFindSurpDdpgBackpropSlowMeanRatioAutographToGraphCUDAInferenceTD}{0.10319444996724406}
  \newcommand{\UncorrectedFindSurpDdpgBackpropSlowMeanRatioAutographToGraphPythonInferenceTD}{3.8988723680744646}
  \newcommand{\UncorrectedFindQualPytorchEagerBetterMinPyTorchEagerSpeedup}{1.8726833861689574}
  \newcommand{\UncorrectedFindQualPytorchEagerBetterMaxPyTorchEagerSpeedup}{1.8726833861689574}
  \newcommand{\UncorrectedFindQualAutographGraphSimilarMaxAutographSpeedup}{0.3525108512249642}
  \newcommand{\UncorrectedFindQualAutographGraphSimilarMaxGraphSpeedup}{0.41894766894746993}
  \newcommand{\UncorrectedFindQualAutographGraphSimilarMaxSpeedup}{0.41894766894746993}
  \newcommand{\UncorrectedFindSurpExecModelComparisonMinTFSpeedup}{1.7305056439463715}
  \newcommand{\UncorrectedFindSurpExecModelComparisonMaxTFSpeedup}{2.4554969495781442}
  \newcommand{\UncorrectedFindSurpTotalGpuTimeMinGPUPercent}{0.01535945516084529}
  \newcommand{\UncorrectedFindSurpTotalGpuTimeMaxGPUPercent}{0.10384931583912607}
  \newcommand{\UncorrectedFindSurpCudaApiDominatesMinRatioCUDAToGPU}{1.195889849143247}
  \newcommand{\UncorrectedFindSurpCudaApiDominatesMaxRatioCUDAToGPU}{10.752731571655387}
  \newcommand{\UncorrectedFindSurpCudaApiDominatesMeanRatioCUDAToGPU}{5.6524353682159925}
  \newcommand{\UncorrectedFindSurpCudaApiDominatesGeomeanRatioCUDAToGPU}{4.832639520309643}
  \newcommand{\UncorrectedFindSurpCudaApiDominatesStdRatioCUDAToGPU}{2.6554692038731345}
  \newcommand{\UncorrectedFindSurpAutographNoGpuMinRatioGPUAutographToNonautograph}{0.49360871145526874}
  \newcommand{\UncorrectedFindSurpAutographNoGpuMaxRatioGPUAutographToNonautograph}{1.2645218294738434}
  \newcommand{\UncorrectedFindSurpAutographNoGpuMeanRatioGPUAutographToNonautograph}{0.8826934214998957}
  \newcommand{\UncorrectedFindSurpAutographNoGpuGeomeanRatioGPUAutographToNonautograph}{0.843152637068627}
  \newcommand{\UncorrectedFindSurpAutographNoGpuStdRatioGPUAutographToNonautograph}{0.28302394416734444}
  \newcommand{\UncorrectedFindSurpAutographInflatesInferenceMinRatioInferenceFrameworkAutographToGraph}{2.212317755094514}
  \newcommand{\UncorrectedFindSurpAutographInflatesInferenceMaxRatioInferenceFrameworkAutographToGraph}{3.9326091608115323}
  \newcommand{\UncorrectedFindSurpAutographInflatesInferenceMeanRatioInferenceFrameworkAutographToGraph}{3.072463457953023}
  \newcommand{\UncorrectedFindSurpAutographInflatesInferenceGeomeanRatioInferenceFrameworkAutographToGraph}{2.9496069348831364}
  \newcommand{\UncorrectedFindSurpAutographInflatesInferenceStdRatioInferenceFrameworkAutographToGraph}{1.216429718599442}
  \newcommand{\UncorrectedFindSurpAutographInflatesInferenceMeanRatioInferenceFrameworkDDPGAutographToGraph}{2.212317755094514}
  \newcommand{\UncorrectedFindSurpAutographInflatesInferenceMeanRatioInferenceFrameworkTDAutographToGraph}{3.9326091608115323}
  \newcommand{\UncorrectedFindSurpAutographInflatesPythonMaxRatioSimulationPythonAutographToEager}{1.5836380826253553}
  \newcommand{\UncorrectedFindSurpAutographInflatesPythonMeanRatioSimulationPythonAutographToEagerDDPG}{1.5836380826253553}
  \newcommand{\UncorrectedFindSurpAutographInflatesPythonMeanRatioSimulationPythonAutographToEagerTD}{1.1591156215570124}
\fi

%% file: tex/autogenerated/uncorrected/AlgoChoiceMetrics.uncorrected.tex
\ifx\TexMetricsEnabled\undefined
  \newcommand{\UncorrectedFindAlgoChoiceMinRatioPercentOnPolicyToOffPolicyBackpropagation}{a}
  \newcommand{\UncorrectedFindAlgoChoiceMaxRatioPercentOnPolicyToOffPolicyBackpropagation}{b}
  \newcommand{\UncorrectedFindAlgoChoiceMeanRatioPercentOnPolicyToOffPolicyBackpropagation}{c}
  \newcommand{\UncorrectedFindAlgoChoiceMinRatioPercentOnPolicyToOffPolicyInference}{d}
  \newcommand{\UncorrectedFindAlgoChoiceMaxRatioPercentOnPolicyToOffPolicyInference}{e}
  \newcommand{\UncorrectedFindAlgoChoiceMeanRatioPercentOnPolicyToOffPolicyInference}{f}
  \newcommand{\UncorrectedFindAlgoChoiceMinRatioPercentOnPolicyToOffPolicySimulation}{g}
  \newcommand{\UncorrectedFindAlgoChoiceMaxRatioPercentOnPolicyToOffPolicySimulation}{h}
  \newcommand{\UncorrectedFindAlgoChoiceMeanRatioPercentOnPolicyToOffPolicySimulation}{i}
  \newcommand{\UncorrectedFindAlgoChoiceACBackpropagationOpPercentMean}{j}
  \newcommand{\UncorrectedFindAlgoChoiceACInferenceOpPercentMean}{k}
  \newcommand{\UncorrectedFindAlgoChoiceACSimulationOpPercentMean}{l}
  \newcommand{\UncorrectedFindAlgoChoiceDDPGBackpropagationOpPercentMean}{m}
  \newcommand{\UncorrectedFindAlgoChoiceDDPGInferenceOpPercentMean}{n}
  \newcommand{\UncorrectedFindAlgoChoiceDDPGSimulationOpPercentMean}{o}
  \newcommand{\UncorrectedFindAlgoChoicePPOBackpropagationOpPercentMean}{p}
  \newcommand{\UncorrectedFindAlgoChoicePPOInferenceOpPercentMean}{q}
  \newcommand{\UncorrectedFindAlgoChoicePPOSimulationOpPercentMean}{r}
  \newcommand{\UncorrectedFindAlgoChoiceSACBackpropagationOpPercentMean}{s}
  \newcommand{\UncorrectedFindAlgoChoiceSACInferenceOpPercentMean}{t}
  \newcommand{\UncorrectedFindAlgoChoiceSACSimulationOpPercentMean}{u}
  \newcommand{\UncorrectedFindAlgoChoiceACCpuResourcePercentMean}{v}
  \newcommand{\UncorrectedFindAlgoChoiceACGpuResourcePercentMean}{w}
  \newcommand{\UncorrectedFindAlgoChoiceDDPGCpuResourcePercentMean}{x}
  \newcommand{\UncorrectedFindAlgoChoiceDDPGGpuResourcePercentMean}{y}
  \newcommand{\UncorrectedFindAlgoChoicePPOCpuResourcePercentMean}{z}
  \newcommand{\UncorrectedFindAlgoChoicePPOGpuResourcePercentMean}{A}
  \newcommand{\UncorrectedFindAlgoChoiceSACCpuResourcePercentMean}{B}
  \newcommand{\UncorrectedFindAlgoChoiceSACGpuResourcePercentMean}{C}
  \newcommand{\UncorrectedFindAlgoChoiceACBackpropagationCpuResourcePercentMean}{D}
  \newcommand{\UncorrectedFindAlgoChoiceACBackpropagationGpuResourcePercentMean}{E}
  \newcommand{\UncorrectedFindAlgoChoiceACInferenceCpuResourcePercentMean}{F}
  \newcommand{\UncorrectedFindAlgoChoiceACInferenceGpuResourcePercentMean}{G}
  \newcommand{\UncorrectedFindAlgoChoiceDDPGBackpropagationCpuResourcePercentMean}{H}
  \newcommand{\UncorrectedFindAlgoChoiceDDPGBackpropagationGpuResourcePercentMean}{I}
  \newcommand{\UncorrectedFindAlgoChoiceDDPGInferenceCpuResourcePercentMean}{J}
  \newcommand{\UncorrectedFindAlgoChoiceDDPGInferenceGpuResourcePercentMean}{K}
  \newcommand{\UncorrectedFindAlgoChoicePPOBackpropagationCpuResourcePercentMean}{L}
  \newcommand{\UncorrectedFindAlgoChoicePPOBackpropagationGpuResourcePercentMean}{M}
  \newcommand{\UncorrectedFindAlgoChoicePPOInferenceCpuResourcePercentMean}{N}
  \newcommand{\UncorrectedFindAlgoChoicePPOInferenceGpuResourcePercentMean}{O}
  \newcommand{\UncorrectedFindAlgoChoiceSACBackpropagationCpuResourcePercentMean}{P}
  \newcommand{\UncorrectedFindAlgoChoiceSACBackpropagationGpuResourcePercentMean}{Q}
  \newcommand{\UncorrectedFindAlgoChoiceSACInferenceCpuResourcePercentMean}{R}
  \newcommand{\UncorrectedFindAlgoChoiceSACInferenceGpuResourcePercentMean}{S}
  \newcommand{\UncorrectedFindAlgoChoiceACBackendCategoryPercentMean}{T}
  \newcommand{\UncorrectedFindAlgoChoiceACCUDACategoryPercentMean}{U}
  \newcommand{\UncorrectedFindAlgoChoiceACPythonCategoryPercentMean}{V}
  \newcommand{\UncorrectedFindAlgoChoiceACSimulatorCategoryPercentMean}{W}
  \newcommand{\UncorrectedFindAlgoChoiceDDPGBackendCategoryPercentMean}{X}
  \newcommand{\UncorrectedFindAlgoChoiceDDPGCUDACategoryPercentMean}{Y}
  \newcommand{\UncorrectedFindAlgoChoiceDDPGPythonCategoryPercentMean}{Z}
  \newcommand{\UncorrectedFindAlgoChoiceDDPGSimulatorCategoryPercentMean}{aa}
  \newcommand{\UncorrectedFindAlgoChoicePPOBackendCategoryPercentMean}{bb}
  \newcommand{\UncorrectedFindAlgoChoicePPOCUDACategoryPercentMean}{cc}
  \newcommand{\UncorrectedFindAlgoChoicePPOPythonCategoryPercentMean}{dd}
  \newcommand{\UncorrectedFindAlgoChoicePPOSimulatorCategoryPercentMean}{ee}
  \newcommand{\UncorrectedFindAlgoChoiceSACBackendCategoryPercentMean}{ff}
  \newcommand{\UncorrectedFindAlgoChoiceSACCUDACategoryPercentMean}{gg}
  \newcommand{\UncorrectedFindAlgoChoiceSACPythonCategoryPercentMean}{hh}
  \newcommand{\UncorrectedFindAlgoChoiceSACSimulatorCategoryPercentMean}{ii}
\else
  \newcommand{\UncorrectedFindAlgoChoiceMinRatioPercentOnPolicyToOffPolicyBackpropagation}{0.115270974748685}
  \newcommand{\UncorrectedFindAlgoChoiceMaxRatioPercentOnPolicyToOffPolicyBackpropagation}{0.5480925369272535}
  \newcommand{\UncorrectedFindAlgoChoiceMeanRatioPercentOnPolicyToOffPolicyBackpropagation}{0.3176344050148199}
  \newcommand{\UncorrectedFindAlgoChoiceMinRatioPercentOnPolicyToOffPolicyInference}{0.8209371165030651}
  \newcommand{\UncorrectedFindAlgoChoiceMaxRatioPercentOnPolicyToOffPolicyInference}{2.591661152637387}
  \newcommand{\UncorrectedFindAlgoChoiceMeanRatioPercentOnPolicyToOffPolicyInference}{1.6168360509757207}
  \newcommand{\UncorrectedFindAlgoChoiceMinRatioPercentOnPolicyToOffPolicySimulation}{3.4267244965196086}
  \newcommand{\UncorrectedFindAlgoChoiceMaxRatioPercentOnPolicyToOffPolicySimulation}{5.161615929856052}
  \newcommand{\UncorrectedFindAlgoChoiceMeanRatioPercentOnPolicyToOffPolicySimulation}{4.290052048847504}
  \newcommand{\UncorrectedFindAlgoChoiceACBackpropagationOpPercentMean}{0.09042924978769398}
  \newcommand{\UncorrectedFindAlgoChoiceACInferenceOpPercentMean}{0.19748697610972946}
  \newcommand{\UncorrectedFindAlgoChoiceACSimulationOpPercentMean}{0.7120837741025766}
  \newcommand{\UncorrectedFindAlgoChoiceDDPGBackpropagationOpPercentMean}{0.6782135328510118}
  \newcommand{\UncorrectedFindAlgoChoiceDDPGInferenceOpPercentMean}{0.18382893916173343}
  \newcommand{\UncorrectedFindAlgoChoiceDDPGSimulationOpPercentMean}{0.13795752798725483}
  \newcommand{\UncorrectedFindAlgoChoicePPOBackpropagationOpPercentMean}{0.3717237757987062}
  \newcommand{\UncorrectedFindAlgoChoicePPOInferenceOpPercentMean}{0.15091199924525084}
  \newcommand{\UncorrectedFindAlgoChoicePPOSimulationOpPercentMean}{0.4773642249560429}
  \newcommand{\UncorrectedFindAlgoChoiceSACBackpropagationOpPercentMean}{0.7844928004196096}
  \newcommand{\UncorrectedFindAlgoChoiceSACInferenceOpPercentMean}{0.07620092461113528}
  \newcommand{\UncorrectedFindAlgoChoiceSACSimulationOpPercentMean}{0.13930627496925513}
  \newcommand{\UncorrectedFindAlgoChoiceACCpuResourcePercentMean}{0.9868115765868581}
  \newcommand{\UncorrectedFindAlgoChoiceACGpuResourcePercentMean}{0.013188423413141902}
  \newcommand{\UncorrectedFindAlgoChoiceDDPGCpuResourcePercentMean}{0.9316787860351122}
  \newcommand{\UncorrectedFindAlgoChoiceDDPGGpuResourcePercentMean}{0.06832121396488774}
  \newcommand{\UncorrectedFindAlgoChoicePPOCpuResourcePercentMean}{0.9669810818460296}
  \newcommand{\UncorrectedFindAlgoChoicePPOGpuResourcePercentMean}{0.03301891815397045}
  \newcommand{\UncorrectedFindAlgoChoiceSACCpuResourcePercentMean}{0.9323911957551164}
  \newcommand{\UncorrectedFindAlgoChoiceSACGpuResourcePercentMean}{0.06760880424488361}
  \newcommand{\UncorrectedFindAlgoChoiceACBackpropagationCpuResourcePercentMean}{0.966780697406984}
  \newcommand{\UncorrectedFindAlgoChoiceACBackpropagationGpuResourcePercentMean}{0.033219302593015886}
  \newcommand{\UncorrectedFindAlgoChoiceACInferenceCpuResourcePercentMean}{0.948429881292393}
  \newcommand{\UncorrectedFindAlgoChoiceACInferenceGpuResourcePercentMean}{0.051570118707606916}
  \newcommand{\UncorrectedFindAlgoChoiceDDPGBackpropagationCpuResourcePercentMean}{0.9204480048320827}
  \newcommand{\UncorrectedFindAlgoChoiceDDPGBackpropagationGpuResourcePercentMean}{0.07955199516791737}
  \newcommand{\UncorrectedFindAlgoChoiceDDPGInferenceCpuResourcePercentMean}{0.9218405200931563}
  \newcommand{\UncorrectedFindAlgoChoiceDDPGInferenceGpuResourcePercentMean}{0.07815947990684373}
  \newcommand{\UncorrectedFindAlgoChoicePPOBackpropagationCpuResourcePercentMean}{0.9278692449398025}
  \newcommand{\UncorrectedFindAlgoChoicePPOBackpropagationGpuResourcePercentMean}{0.07213075506019746}
  \newcommand{\UncorrectedFindAlgoChoicePPOInferenceCpuResourcePercentMean}{0.958875360721341}
  \newcommand{\UncorrectedFindAlgoChoicePPOInferenceGpuResourcePercentMean}{0.04112463927865907}
  \newcommand{\UncorrectedFindAlgoChoiceSACBackpropagationCpuResourcePercentMean}{0.9168183252447575}
  \newcommand{\UncorrectedFindAlgoChoiceSACBackpropagationGpuResourcePercentMean}{0.0831816747552425}
  \newcommand{\UncorrectedFindAlgoChoiceSACInferenceCpuResourcePercentMean}{0.9691161323230669}
  \newcommand{\UncorrectedFindAlgoChoiceSACInferenceGpuResourcePercentMean}{0.03088386767693316}
  \newcommand{\UncorrectedFindAlgoChoiceACBackendCategoryPercentMean}{0.09743167385931133}
  \newcommand{\UncorrectedFindAlgoChoiceACCUDACategoryPercentMean}{0.0827962568560451}
  \newcommand{\UncorrectedFindAlgoChoiceACPythonCategoryPercentMean}{0.5338958489939756}
  \newcommand{\UncorrectedFindAlgoChoiceACSimulatorCategoryPercentMean}{0.28587622029066795}
  \newcommand{\UncorrectedFindAlgoChoiceDDPGBackendCategoryPercentMean}{0.27053612194711674}
  \newcommand{\UncorrectedFindAlgoChoiceDDPGCUDACategoryPercentMean}{0.36851833948624235}
  \newcommand{\UncorrectedFindAlgoChoiceDDPGPythonCategoryPercentMean}{0.29628420023537266}
  \newcommand{\UncorrectedFindAlgoChoiceDDPGSimulatorCategoryPercentMean}{0.06466133833126828}
  \newcommand{\UncorrectedFindAlgoChoicePPOBackendCategoryPercentMean}{0.17409155354135128}
  \newcommand{\UncorrectedFindAlgoChoicePPOCUDACategoryPercentMean}{0.21494529463320539}
  \newcommand{\UncorrectedFindAlgoChoicePPOPythonCategoryPercentMean}{0.4128270457928353}
  \newcommand{\UncorrectedFindAlgoChoicePPOSimulatorCategoryPercentMean}{0.19813610603260806}
  \newcommand{\UncorrectedFindAlgoChoiceSACBackendCategoryPercentMean}{0.2638748187586117}
  \newcommand{\UncorrectedFindAlgoChoiceSACCUDACategoryPercentMean}{0.35068845199229054}
  \newcommand{\UncorrectedFindAlgoChoiceSACPythonCategoryPercentMean}{0.3268570078458032}
  \newcommand{\UncorrectedFindAlgoChoiceSACSimulatorCategoryPercentMean}{0.058579721403294704}
\fi

%% file: tex/autogenerated/uncorrected/FrameworkChoiceMetricsUncorrected.tex
\ifx\TexMetricsEnabled\undefined
  \newcommand{\FindUncorrectedTrainingTimeInflationMinRatio}{a}
  \newcommand{\FindUncorrectedTrainingTimeInflationMaxRatio}{b}
  \newcommand{\FindUncorrectedTrainingTimeInflationMeanRatio}{c}
\else
  \newcommand{\FindUncorrectedTrainingTimeInflationMinRatio}{1.558473083578901}
  \newcommand{\FindUncorrectedTrainingTimeInflationMaxRatio}{2.1654849733110866}
  \newcommand{\FindUncorrectedTrainingTimeInflationMeanRatio}{1.8370112649090875}
\fi

%% file: tex/autogenerated/FrameworkChoiceMetrics.tex
\ifx\TexMetricsEnabled\undefined
  \newcommand{\FindQualEagerMoreTransMinEagerSlowdown}{a}
  \newcommand{\FindQualEagerMoreTransMaxEagerSlowdown}{b}
  \newcommand{\FindQualEagerMoreTransMinRatioInferenceFrameworkTransTFToPyTorch}{c}
  \newcommand{\FindQualEagerMoreTransMaxRatioInferenceFrameworkTransTFToPyTorch}{d}
  \newcommand{\FindQualEagerMoreTransMeanRatioInferenceFrameworkTransTFToPyTorch}{e}
  \newcommand{\FindQualEagerMoreTransGeomeanRatioInferenceFrameworkTransTFToPyTorch}{f}
  \newcommand{\FindQualEagerMoreTransStdRatioInferenceFrameworkTransTFToPyTorch}{g}
  \newcommand{\FindQualEagerMoreTransMinRatioBackpropagationFrameworkTransTFToPyTorch}{h}
  \newcommand{\FindQualEagerMoreTransMaxRatioBackpropagationFrameworkTransTFToPyTorch}{i}
  \newcommand{\FindQualEagerMoreTransMeanRatioBackpropagationFrameworkTransTFToPyTorch}{j}
  \newcommand{\FindQualEagerMoreTransGeomeanRatioBackpropagationFrameworkTransTFToPyTorch}{k}
  \newcommand{\FindQualEagerMoreTransStdRatioBackpropagationFrameworkTransTFToPyTorch}{l}
  \newcommand{\FindQualEagerMoreTransMinRatioInferenceFrameworkTFToPyTorch}{m}
  \newcommand{\FindQualEagerMoreTransMaxRatioInferenceFrameworkTFToPyTorch}{n}
  \newcommand{\FindQualEagerMoreTransMeanRatioInferenceFrameworkTFToPyTorch}{o}
  \newcommand{\FindQualEagerMoreTransGeomeanRatioInferenceFrameworkTFToPyTorch}{p}
  \newcommand{\FindQualEagerMoreTransStdRatioInferenceFrameworkTFToPyTorch}{q}
  \newcommand{\FindQualEagerMoreTransMinRatioBackpropagationFrameworkTFToPyTorch}{r}
  \newcommand{\FindQualEagerMoreTransMaxRatioBackpropagationFrameworkTFToPyTorch}{s}
  \newcommand{\FindQualEagerMoreTransMeanRatioBackpropagationFrameworkTFToPyTorch}{t}
  \newcommand{\FindQualEagerMoreTransGeomeanRatioBackpropagationFrameworkTFToPyTorch}{u}
  \newcommand{\FindQualEagerMoreTransStdRatioBackpropagationFrameworkTFToPyTorch}{v}
  \newcommand{\FindQualAutographReducesPythonMinAutographPythonPercentOfOp}{a}
  \newcommand{\FindQualAutographReducesPythonMaxAutographPythonPercentOfOp}{b}
  \newcommand{\FindQualAutographReducesPythonMinGraphPythonPercentOfOp}{c}
  \newcommand{\FindQualAutographReducesPythonMaxGraphPythonPercentOfOp}{d}
  \newcommand{\FindQualAutographReducesPythonMinRatioPythonBackpropagationDDPG}{e}
  \newcommand{\FindQualAutographReducesPythonMaxRatioPythonBackpropagationDDPG}{f}
  \newcommand{\FindQualAutographReducesPythonMeanRatioPythonBackpropagationDDPG}{g}
  \newcommand{\FindQualAutographReducesPythonGeomeanRatioPythonBackpropagationDDPG}{h}
  \newcommand{\FindQualAutographReducesPythonStdRatioPythonBackpropagationDDPG}{i}
  \newcommand{\FindQualAutographReducesPythonMinRatioPythonInferenceDDPG}{j}
  \newcommand{\FindQualAutographReducesPythonMaxRatioPythonInferenceDDPG}{k}
  \newcommand{\FindQualAutographReducesPythonMeanRatioPythonInferenceDDPG}{l}
  \newcommand{\FindQualAutographReducesPythonGeomeanRatioPythonInferenceDDPG}{m}
  \newcommand{\FindQualAutographReducesPythonStdRatioPythonInferenceDDPG}{n}
  \newcommand{\FindQualAutographReducesPythonMinRatioPythonBackpropagationTD}{o}
  \newcommand{\FindQualAutographReducesPythonMaxRatioPythonBackpropagationTD}{p}
  \newcommand{\FindQualAutographReducesPythonMeanRatioPythonBackpropagationTD}{q}
  \newcommand{\FindQualAutographReducesPythonGeomeanRatioPythonBackpropagationTD}{r}
  \newcommand{\FindQualAutographReducesPythonStdRatioPythonBackpropagationTD}{s}
  \newcommand{\FindQualAutographReducesPythonMinRatioPythonInferenceTD}{t}
  \newcommand{\FindQualAutographReducesPythonMaxRatioPythonInferenceTD}{u}
  \newcommand{\FindQualAutographReducesPythonMeanRatioPythonInferenceTD}{v}
  \newcommand{\FindQualAutographReducesPythonGeomeanRatioPythonInferenceTD}{w}
  \newcommand{\FindQualAutographReducesPythonStdRatioPythonInferenceTD}{x}
  \newcommand{\FindSurpDdpgBackpropSlowMeanRatioAutographToGraphBackpropagationDDPG}{a}
  \newcommand{\FindSurpDdpgBackpropSlowMeanRatioAutographToGraphInferenceDDPG}{b}
  \newcommand{\FindSurpDdpgBackpropSlowMeanRatioAutographToGraphBackpropagationTD}{c}
  \newcommand{\FindSurpDdpgBackpropSlowMeanRatioAutographToGraphInferenceTD}{d}
  \newcommand{\FindSurpDdpgBackpropSlowMeanRatioAutographToGraphCUDABackpropagationDDPG}{e}
  \newcommand{\FindSurpDdpgBackpropSlowMeanRatioAutographToGraphPythonBackpropagationDDPG}{f}
  \newcommand{\FindSurpDdpgBackpropSlowMeanRatioAutographToGraphCUDAInferenceDDPG}{g}
  \newcommand{\FindSurpDdpgBackpropSlowMeanRatioAutographToGraphPythonInferenceDDPG}{h}
  \newcommand{\FindSurpDdpgBackpropSlowMeanRatioAutographToGraphCUDABackpropagationTD}{i}
  \newcommand{\FindSurpDdpgBackpropSlowMeanRatioAutographToGraphPythonBackpropagationTD}{j}
  \newcommand{\FindSurpDdpgBackpropSlowMeanRatioAutographToGraphCUDAInferenceTD}{k}
  \newcommand{\FindSurpDdpgBackpropSlowMeanRatioAutographToGraphPythonInferenceTD}{l}
  \newcommand{\FindQualPytorchEagerBetterMinPyTorchEagerSpeedup}{a}
  \newcommand{\FindQualPytorchEagerBetterMaxPyTorchEagerSpeedup}{b}
  \newcommand{\FindQualAutographGraphSimilarMaxAutographSpeedup}{a}
  \newcommand{\FindQualAutographGraphSimilarMaxGraphSpeedup}{b}
  \newcommand{\FindQualAutographGraphSimilarMaxSpeedup}{c}
  \newcommand{\FindSurpExecModelComparisonMinTFSpeedup}{a}
  \newcommand{\FindSurpExecModelComparisonMaxTFSpeedup}{b}
  \newcommand{\FindSurpTotalGpuTimeMinGPUPercent}{a}
  \newcommand{\FindSurpTotalGpuTimeMaxGPUPercent}{b}
  \newcommand{\FindSurpCudaApiDominatesMinRatioCUDAToGPU}{a}
  \newcommand{\FindSurpCudaApiDominatesMaxRatioCUDAToGPU}{b}
  \newcommand{\FindSurpCudaApiDominatesMeanRatioCUDAToGPU}{c}
  \newcommand{\FindSurpCudaApiDominatesGeomeanRatioCUDAToGPU}{d}
  \newcommand{\FindSurpCudaApiDominatesStdRatioCUDAToGPU}{e}
  \newcommand{\FindSurpAutographNoGpuMinRatioGPUAutographToNonautograph}{a}
  \newcommand{\FindSurpAutographNoGpuMaxRatioGPUAutographToNonautograph}{b}
  \newcommand{\FindSurpAutographNoGpuMeanRatioGPUAutographToNonautograph}{c}
  \newcommand{\FindSurpAutographNoGpuGeomeanRatioGPUAutographToNonautograph}{d}
  \newcommand{\FindSurpAutographNoGpuStdRatioGPUAutographToNonautograph}{e}
  \newcommand{\FindSurpAutographInflatesInferenceMinRatioInferenceFrameworkAutographToGraph}{a}
  \newcommand{\FindSurpAutographInflatesInferenceMaxRatioInferenceFrameworkAutographToGraph}{b}
  \newcommand{\FindSurpAutographInflatesInferenceMeanRatioInferenceFrameworkAutographToGraph}{c}
  \newcommand{\FindSurpAutographInflatesInferenceGeomeanRatioInferenceFrameworkAutographToGraph}{d}
  \newcommand{\FindSurpAutographInflatesInferenceStdRatioInferenceFrameworkAutographToGraph}{e}
  \newcommand{\FindSurpAutographInflatesInferenceMeanRatioInferenceFrameworkDDPGAutographToGraph}{f}
  \newcommand{\FindSurpAutographInflatesInferenceMeanRatioInferenceFrameworkTDAutographToGraph}{g}
  \newcommand{\FindSurpAutographInflatesPythonMaxRatioSimulationPythonAutographToEager}{a}
  \newcommand{\FindSurpAutographInflatesPythonMeanRatioSimulationPythonAutographToEagerDDPG}{b}
  \newcommand{\FindSurpAutographInflatesPythonMeanRatioSimulationPythonAutographToEagerTD}{c}
\else
  \newcommand{\FindQualEagerMoreTransMinEagerSlowdown}{1.8532326839977415}
  \newcommand{\FindQualEagerMoreTransMaxEagerSlowdown}{4.761963170388129}
  \newcommand{\FindQualEagerMoreTransMinRatioInferenceFrameworkTransTFToPyTorch}{3.155013832244816}
  \newcommand{\FindQualEagerMoreTransMaxRatioInferenceFrameworkTransTFToPyTorch}{3.155013832244816}
  \newcommand{\FindQualEagerMoreTransMeanRatioInferenceFrameworkTransTFToPyTorch}{3.155013832244816}
  \newcommand{\FindQualEagerMoreTransGeomeanRatioInferenceFrameworkTransTFToPyTorch}{3.155013832244816}
  \newcommand{\FindQualEagerMoreTransStdRatioInferenceFrameworkTransTFToPyTorch}{nan}
  \newcommand{\FindQualEagerMoreTransMinRatioBackpropagationFrameworkTransTFToPyTorch}{1.599702380952381}
  \newcommand{\FindQualEagerMoreTransMaxRatioBackpropagationFrameworkTransTFToPyTorch}{1.599702380952381}
  \newcommand{\FindQualEagerMoreTransMeanRatioBackpropagationFrameworkTransTFToPyTorch}{1.599702380952381}
  \newcommand{\FindQualEagerMoreTransGeomeanRatioBackpropagationFrameworkTransTFToPyTorch}{1.599702380952381}
  \newcommand{\FindQualEagerMoreTransStdRatioBackpropagationFrameworkTransTFToPyTorch}{nan}
  \newcommand{\FindQualEagerMoreTransMinRatioInferenceFrameworkTFToPyTorch}{59.480646245433}
  \newcommand{\FindQualEagerMoreTransMaxRatioInferenceFrameworkTFToPyTorch}{59.480646245433}
  \newcommand{\FindQualEagerMoreTransMeanRatioInferenceFrameworkTFToPyTorch}{59.480646245433}
  \newcommand{\FindQualEagerMoreTransGeomeanRatioInferenceFrameworkTFToPyTorch}{59.48064624543302}
  \newcommand{\FindQualEagerMoreTransStdRatioInferenceFrameworkTFToPyTorch}{nan}
  \newcommand{\FindQualEagerMoreTransMinRatioBackpropagationFrameworkTFToPyTorch}{3.4068920585775015}
  \newcommand{\FindQualEagerMoreTransMaxRatioBackpropagationFrameworkTFToPyTorch}{3.4068920585775015}
  \newcommand{\FindQualEagerMoreTransMeanRatioBackpropagationFrameworkTFToPyTorch}{3.4068920585775015}
  \newcommand{\FindQualEagerMoreTransGeomeanRatioBackpropagationFrameworkTFToPyTorch}{3.4068920585775015}
  \newcommand{\FindQualEagerMoreTransStdRatioBackpropagationFrameworkTFToPyTorch}{nan}
  \newcommand{\FindQualAutographReducesPythonMinAutographPythonPercentOfOp}{0.055828119863115144}
  \newcommand{\FindQualAutographReducesPythonMaxAutographPythonPercentOfOp}{0.07154774455058878}
  \newcommand{\FindQualAutographReducesPythonMinGraphPythonPercentOfOp}{0.19747385693349523}
  \newcommand{\FindQualAutographReducesPythonMaxGraphPythonPercentOfOp}{0.59236488217148}
  \newcommand{\FindQualAutographReducesPythonMinRatioPythonBackpropagationDDPG}{13.497151452448529}
  \newcommand{\FindQualAutographReducesPythonMaxRatioPythonBackpropagationDDPG}{13.497151452448529}
  \newcommand{\FindQualAutographReducesPythonMeanRatioPythonBackpropagationDDPG}{13.497151452448529}
  \newcommand{\FindQualAutographReducesPythonGeomeanRatioPythonBackpropagationDDPG}{13.497151452448529}
  \newcommand{\FindQualAutographReducesPythonStdRatioPythonBackpropagationDDPG}{nan}
  \newcommand{\FindQualAutographReducesPythonMinRatioPythonInferenceDDPG}{1.8140409472650212}
  \newcommand{\FindQualAutographReducesPythonMaxRatioPythonInferenceDDPG}{1.8140409472650212}
  \newcommand{\FindQualAutographReducesPythonMeanRatioPythonInferenceDDPG}{1.8140409472650212}
  \newcommand{\FindQualAutographReducesPythonGeomeanRatioPythonInferenceDDPG}{1.8140409472650212}
  \newcommand{\FindQualAutographReducesPythonStdRatioPythonInferenceDDPG}{nan}
  \newcommand{\FindQualAutographReducesPythonMinRatioPythonBackpropagationTD}{5.140804189241857}
  \newcommand{\FindQualAutographReducesPythonMaxRatioPythonBackpropagationTD}{5.140804189241857}
  \newcommand{\FindQualAutographReducesPythonMeanRatioPythonBackpropagationTD}{5.140804189241857}
  \newcommand{\FindQualAutographReducesPythonGeomeanRatioPythonBackpropagationTD}{5.140804189241857}
  \newcommand{\FindQualAutographReducesPythonStdRatioPythonBackpropagationTD}{nan}
  \newcommand{\FindQualAutographReducesPythonMinRatioPythonInferenceTD}{4.429123672492049}
  \newcommand{\FindQualAutographReducesPythonMaxRatioPythonInferenceTD}{4.429123672492049}
  \newcommand{\FindQualAutographReducesPythonMeanRatioPythonInferenceTD}{4.429123672492049}
  \newcommand{\FindQualAutographReducesPythonGeomeanRatioPythonInferenceTD}{4.429123672492049}
  \newcommand{\FindQualAutographReducesPythonStdRatioPythonInferenceTD}{nan}
  \newcommand{\FindSurpDdpgBackpropSlowMeanRatioAutographToGraphBackpropagationDDPG}{3.7425521095138428}
  \newcommand{\FindSurpDdpgBackpropSlowMeanRatioAutographToGraphInferenceDDPG}{0.5128501413461403}
  \newcommand{\FindSurpDdpgBackpropSlowMeanRatioAutographToGraphBackpropagationTD}{1.2344689632580377}
  \newcommand{\FindSurpDdpgBackpropSlowMeanRatioAutographToGraphInferenceTD}{0.4459416048181845}
  \newcommand{\FindSurpDdpgBackpropSlowMeanRatioAutographToGraphCUDABackpropagationDDPG}{2.6411493822527876}
  \newcommand{\FindSurpDdpgBackpropSlowMeanRatioAutographToGraphPythonBackpropagationDDPG}{13.497151452448529}
  \newcommand{\FindSurpDdpgBackpropSlowMeanRatioAutographToGraphCUDAInferenceDDPG}{0.6126382808073244}
  \newcommand{\FindSurpDdpgBackpropSlowMeanRatioAutographToGraphPythonInferenceDDPG}{1.8140409472650212}
  \newcommand{\FindSurpDdpgBackpropSlowMeanRatioAutographToGraphCUDABackpropagationTD}{1.2196412813058282}
  \newcommand{\FindSurpDdpgBackpropSlowMeanRatioAutographToGraphPythonBackpropagationTD}{5.140804189241857}
  \newcommand{\FindSurpDdpgBackpropSlowMeanRatioAutographToGraphCUDAInferenceTD}{0.09472264722622846}
  \newcommand{\FindSurpDdpgBackpropSlowMeanRatioAutographToGraphPythonInferenceTD}{4.429123672492049}
  \newcommand{\FindQualPytorchEagerBetterMinPyTorchEagerSpeedup}{2.342969522180876}
  \newcommand{\FindQualPytorchEagerBetterMaxPyTorchEagerSpeedup}{2.342969522180876}
  \newcommand{\FindQualAutographGraphSimilarMaxAutographSpeedup}{0.19745142483862074}
  \newcommand{\FindQualAutographGraphSimilarMaxGraphSpeedup}{0.09670403679968885}
  \newcommand{\FindQualAutographGraphSimilarMaxSpeedup}{0.19745142483862074}
  \newcommand{\FindSurpExecModelComparisonMinTFSpeedup}{1.8532326839977415}
  \newcommand{\FindSurpExecModelComparisonMaxTFSpeedup}{2.032447765669445}
  \newcommand{\FindSurpTotalGpuTimeMinGPUPercent}{0.02157154435549628}
  \newcommand{\FindSurpTotalGpuTimeMaxGPUPercent}{0.1414541625849027}
  \newcommand{\FindSurpCudaApiDominatesMinRatioCUDAToGPU}{0.7419781155262749}
  \newcommand{\FindSurpCudaApiDominatesMaxRatioCUDAToGPU}{9.131803034499567}
  \newcommand{\FindSurpCudaApiDominatesMeanRatioCUDAToGPU}{3.615545639104065}
  \newcommand{\FindSurpCudaApiDominatesGeomeanRatioCUDAToGPU}{3.0739398237227396}
  \newcommand{\FindSurpCudaApiDominatesStdRatioCUDAToGPU}{1.9243313441954812}
  \newcommand{\FindSurpAutographNoGpuMinRatioGPUAutographToNonautograph}{0.510243460351047}
  \newcommand{\FindSurpAutographNoGpuMaxRatioGPUAutographToNonautograph}{1.0643760270632519}
  \newcommand{\FindSurpAutographNoGpuMeanRatioGPUAutographToNonautograph}{0.7768965080006982}
  \newcommand{\FindSurpAutographNoGpuGeomeanRatioGPUAutographToNonautograph}{0.7531146017327028}
  \newcommand{\FindSurpAutographNoGpuStdRatioGPUAutographToNonautograph}{0.21314061903167708}
  \newcommand{\FindSurpAutographInflatesInferenceMinRatioInferenceFrameworkAutographToGraph}{3.7917878755076933}
  \newcommand{\FindSurpAutographInflatesInferenceMaxRatioInferenceFrameworkAutographToGraph}{4.380135934106408}
  \newcommand{\FindSurpAutographInflatesInferenceMeanRatioInferenceFrameworkAutographToGraph}{4.085961904807051}
  \newcommand{\FindSurpAutographInflatesInferenceGeomeanRatioInferenceFrameworkAutographToGraph}{4.075358429392468}
  \newcommand{\FindSurpAutographInflatesInferenceStdRatioInferenceFrameworkAutographToGraph}{0.41602490193309166}
  \newcommand{\FindSurpAutographInflatesInferenceMeanRatioInferenceFrameworkDDPGAutographToGraph}{4.380135934106408}
  \newcommand{\FindSurpAutographInflatesInferenceMeanRatioInferenceFrameworkTDAutographToGraph}{3.7917878755076933}
  \newcommand{\FindSurpAutographInflatesPythonMaxRatioSimulationPythonAutographToEager}{2.448781101364494}
  \newcommand{\FindSurpAutographInflatesPythonMeanRatioSimulationPythonAutographToEagerDDPG}{2.448781101364494}
  \newcommand{\FindSurpAutographInflatesPythonMeanRatioSimulationPythonAutographToEagerTD}{1.4814400243206378}
\fi

%% file: tex/autogenerated/AlgoChoiceMetrics.tex
\ifx\TexMetricsEnabled\undefined
  \newcommand{\FindAlgoChoiceMinRatioPercentOnPolicyToOffPolicyBackpropagation}{a}
  \newcommand{\FindAlgoChoiceMaxRatioPercentOnPolicyToOffPolicyBackpropagation}{b}
  \newcommand{\FindAlgoChoiceMeanRatioPercentOnPolicyToOffPolicyBackpropagation}{c}
  \newcommand{\FindAlgoChoiceMinRatioPercentOnPolicyToOffPolicyInference}{d}
  \newcommand{\FindAlgoChoiceMaxRatioPercentOnPolicyToOffPolicyInference}{e}
  \newcommand{\FindAlgoChoiceMeanRatioPercentOnPolicyToOffPolicyInference}{f}
  \newcommand{\FindAlgoChoiceMinRatioPercentOnPolicyToOffPolicySimulation}{g}
  \newcommand{\FindAlgoChoiceMaxRatioPercentOnPolicyToOffPolicySimulation}{h}
  \newcommand{\FindAlgoChoiceMeanRatioPercentOnPolicyToOffPolicySimulation}{i}
  \newcommand{\FindAlgoChoiceACBackpropagationOpPercentMean}{j}
  \newcommand{\FindAlgoChoiceACInferenceOpPercentMean}{k}
  \newcommand{\FindAlgoChoiceACSimulationOpPercentMean}{l}
  \newcommand{\FindAlgoChoiceDDPGBackpropagationOpPercentMean}{m}
  \newcommand{\FindAlgoChoiceDDPGInferenceOpPercentMean}{n}
  \newcommand{\FindAlgoChoiceDDPGSimulationOpPercentMean}{o}
  \newcommand{\FindAlgoChoicePPOBackpropagationOpPercentMean}{p}
  \newcommand{\FindAlgoChoicePPOInferenceOpPercentMean}{q}
  \newcommand{\FindAlgoChoicePPOSimulationOpPercentMean}{r}
  \newcommand{\FindAlgoChoiceSACBackpropagationOpPercentMean}{s}
  \newcommand{\FindAlgoChoiceSACInferenceOpPercentMean}{t}
  \newcommand{\FindAlgoChoiceSACSimulationOpPercentMean}{u}
  \newcommand{\FindAlgoChoiceACCpuResourcePercentMean}{v}
  \newcommand{\FindAlgoChoiceACGpuResourcePercentMean}{w}
  \newcommand{\FindAlgoChoiceDDPGCpuResourcePercentMean}{x}
  \newcommand{\FindAlgoChoiceDDPGGpuResourcePercentMean}{y}
  \newcommand{\FindAlgoChoicePPOCpuResourcePercentMean}{z}
  \newcommand{\FindAlgoChoicePPOGpuResourcePercentMean}{A}
  \newcommand{\FindAlgoChoiceSACCpuResourcePercentMean}{B}
  \newcommand{\FindAlgoChoiceSACGpuResourcePercentMean}{C}
  \newcommand{\FindAlgoChoiceACBackpropagationCpuResourcePercentMean}{D}
  \newcommand{\FindAlgoChoiceACBackpropagationGpuResourcePercentMean}{E}
  \newcommand{\FindAlgoChoiceACInferenceCpuResourcePercentMean}{F}
  \newcommand{\FindAlgoChoiceACInferenceGpuResourcePercentMean}{G}
  \newcommand{\FindAlgoChoiceDDPGBackpropagationCpuResourcePercentMean}{H}
  \newcommand{\FindAlgoChoiceDDPGBackpropagationGpuResourcePercentMean}{I}
  \newcommand{\FindAlgoChoiceDDPGInferenceCpuResourcePercentMean}{J}
  \newcommand{\FindAlgoChoiceDDPGInferenceGpuResourcePercentMean}{K}
  \newcommand{\FindAlgoChoicePPOBackpropagationCpuResourcePercentMean}{L}
  \newcommand{\FindAlgoChoicePPOBackpropagationGpuResourcePercentMean}{M}
  \newcommand{\FindAlgoChoicePPOInferenceCpuResourcePercentMean}{N}
  \newcommand{\FindAlgoChoicePPOInferenceGpuResourcePercentMean}{O}
  \newcommand{\FindAlgoChoiceSACBackpropagationCpuResourcePercentMean}{P}
  \newcommand{\FindAlgoChoiceSACBackpropagationGpuResourcePercentMean}{Q}
  \newcommand{\FindAlgoChoiceSACInferenceCpuResourcePercentMean}{R}
  \newcommand{\FindAlgoChoiceSACInferenceGpuResourcePercentMean}{S}
  \newcommand{\FindAlgoChoiceACBackendCategoryPercentMean}{T}
  \newcommand{\FindAlgoChoiceACCUDACategoryPercentMean}{U}
  \newcommand{\FindAlgoChoiceACPythonCategoryPercentMean}{V}
  \newcommand{\FindAlgoChoiceACSimulatorCategoryPercentMean}{W}
  \newcommand{\FindAlgoChoiceDDPGBackendCategoryPercentMean}{X}
  \newcommand{\FindAlgoChoiceDDPGCUDACategoryPercentMean}{Y}
  \newcommand{\FindAlgoChoiceDDPGPythonCategoryPercentMean}{Z}
  \newcommand{\FindAlgoChoiceDDPGSimulatorCategoryPercentMean}{aa}
  \newcommand{\FindAlgoChoicePPOBackendCategoryPercentMean}{bb}
  \newcommand{\FindAlgoChoicePPOCUDACategoryPercentMean}{cc}
  \newcommand{\FindAlgoChoicePPOPythonCategoryPercentMean}{dd}
  \newcommand{\FindAlgoChoicePPOSimulatorCategoryPercentMean}{ee}
  \newcommand{\FindAlgoChoiceSACBackendCategoryPercentMean}{ff}
  \newcommand{\FindAlgoChoiceSACCUDACategoryPercentMean}{gg}
  \newcommand{\FindAlgoChoiceSACPythonCategoryPercentMean}{hh}
  \newcommand{\FindAlgoChoiceSACSimulatorCategoryPercentMean}{ii}
\else
  \newcommand{\FindAlgoChoiceMinRatioPercentOnPolicyToOffPolicyBackpropagation}{0.15691900907146536}
  \newcommand{\FindAlgoChoiceMaxRatioPercentOnPolicyToOffPolicyBackpropagation}{0.5423675235810014}
  \newcommand{\FindAlgoChoiceMeanRatioPercentOnPolicyToOffPolicyBackpropagation}{0.34168271503131087}
  \newcommand{\FindAlgoChoiceMinRatioPercentOnPolicyToOffPolicyInference}{0.8501801284784146}
  \newcommand{\FindAlgoChoiceMaxRatioPercentOnPolicyToOffPolicyInference}{2.53616784175325}
  \newcommand{\FindAlgoChoiceMeanRatioPercentOnPolicyToOffPolicyInference}{1.596789936530216}
  \newcommand{\FindAlgoChoiceMinRatioPercentOnPolicyToOffPolicySimulation}{3.483607505461769}
  \newcommand{\FindAlgoChoiceMaxRatioPercentOnPolicyToOffPolicySimulation}{6.1679465311102755}
  \newcommand{\FindAlgoChoiceMeanRatioPercentOnPolicyToOffPolicySimulation}{4.738697826267484}
  \newcommand{\FindAlgoChoiceACBackpropagationOpPercentMean}{0.12327482306988842}
  \newcommand{\FindAlgoChoiceACInferenceOpPercentMean}{0.20650977286962188}
  \newcommand{\FindAlgoChoiceACSimulationOpPercentMean}{0.6702154040604897}
  \newcommand{\FindAlgoChoiceDDPGBackpropagationOpPercentMean}{0.7181151109265606}
  \newcommand{\FindAlgoChoiceDDPGInferenceOpPercentMean}{0.17322386215305116}
  \newcommand{\FindAlgoChoiceDDPGSimulationOpPercentMean}{0.1086610269203884}
  \newcommand{\FindAlgoChoicePPOBackpropagationOpPercentMean}{0.3894823143593347}
  \newcommand{\FindAlgoChoicePPOInferenceOpPercentMean}{0.14727148538080817}
  \newcommand{\FindAlgoChoicePPOSimulationOpPercentMean}{0.4632462002598571}
  \newcommand{\FindAlgoChoiceSACBackpropagationOpPercentMean}{0.7855952175542069}
  \newcommand{\FindAlgoChoiceSACInferenceOpPercentMean}{0.08142590938573763}
  \newcommand{\FindAlgoChoiceSACSimulationOpPercentMean}{0.13297887306005549}
  \newcommand{\FindAlgoChoiceACCpuResourcePercentMean}{0.9738371239118612}
  \newcommand{\FindAlgoChoiceACGpuResourcePercentMean}{0.026162876088138786}
  \newcommand{\FindAlgoChoiceDDPGCpuResourcePercentMean}{0.8940238715970722}
  \newcommand{\FindAlgoChoiceDDPGGpuResourcePercentMean}{0.1059761284029278}
  \newcommand{\FindAlgoChoicePPOCpuResourcePercentMean}{0.9404249298208406}
  \newcommand{\FindAlgoChoicePPOGpuResourcePercentMean}{0.05957507017915938}
  \newcommand{\FindAlgoChoiceSACCpuResourcePercentMean}{0.9065233290444658}
  \newcommand{\FindAlgoChoiceSACGpuResourcePercentMean}{0.09347667095553423}
  \newcommand{\FindAlgoChoiceACBackpropagationCpuResourcePercentMean}{0.9515796774721544}
  \newcommand{\FindAlgoChoiceACBackpropagationGpuResourcePercentMean}{0.04842032252784568}
  \newcommand{\FindAlgoChoiceACInferenceCpuResourcePercentMean}{0.9022134927808918}
  \newcommand{\FindAlgoChoiceACInferenceGpuResourcePercentMean}{0.09778650721910821}
  \newcommand{\FindAlgoChoiceDDPGBackpropagationCpuResourcePercentMean}{0.883458854217872}
  \newcommand{\FindAlgoChoiceDDPGBackpropagationGpuResourcePercentMean}{0.11654114578212799}
  \newcommand{\FindAlgoChoiceDDPGInferenceCpuResourcePercentMean}{0.8713446848772162}
  \newcommand{\FindAlgoChoiceDDPGInferenceGpuResourcePercentMean}{0.12865531512278391}
  \newcommand{\FindAlgoChoicePPOBackpropagationCpuResourcePercentMean}{0.8751721343841706}
  \newcommand{\FindAlgoChoicePPOBackpropagationGpuResourcePercentMean}{0.12482786561582927}
  \newcommand{\FindAlgoChoicePPOInferenceCpuResourcePercentMean}{0.9256011837305879}
  \newcommand{\FindAlgoChoicePPOInferenceGpuResourcePercentMean}{0.07439881626941211}
  \newcommand{\FindAlgoChoiceSACBackpropagationCpuResourcePercentMean}{0.885034319095085}
  \newcommand{\FindAlgoChoiceSACBackpropagationGpuResourcePercentMean}{0.11496568090491498}
  \newcommand{\FindAlgoChoiceSACInferenceCpuResourcePercentMean}{0.9611894803802639}
  \newcommand{\FindAlgoChoiceSACInferenceGpuResourcePercentMean}{0.03881051961973612}
  \newcommand{\FindAlgoChoiceACBackendCategoryPercentMean}{0.11772080599803439}
  \newcommand{\FindAlgoChoiceACCUDACategoryPercentMean}{0.06830575148941499}
  \newcommand{\FindAlgoChoiceACPythonCategoryPercentMean}{0.5068558392124939}
  \newcommand{\FindAlgoChoiceACSimulatorCategoryPercentMean}{0.3071176033000567}
  \newcommand{\FindAlgoChoiceDDPGBackendCategoryPercentMean}{0.25714353475987584}
  \newcommand{\FindAlgoChoiceDDPGCUDACategoryPercentMean}{0.37204813315137797}
  \newcommand{\FindAlgoChoiceDDPGPythonCategoryPercentMean}{0.31122737228452985}
  \newcommand{\FindAlgoChoiceDDPGSimulatorCategoryPercentMean}{0.05958095980421628}
  \newcommand{\FindAlgoChoicePPOBackendCategoryPercentMean}{0.18334822468184173}
  \newcommand{\FindAlgoChoicePPOCUDACategoryPercentMean}{0.1931145322686568}
  \newcommand{\FindAlgoChoicePPOPythonCategoryPercentMean}{0.40630848016509913}
  \newcommand{\FindAlgoChoicePPOSimulatorCategoryPercentMean}{0.21722876288440235}
  \newcommand{\FindAlgoChoiceSACBackendCategoryPercentMean}{0.2359582614591876}
  \newcommand{\FindAlgoChoiceSACCUDACategoryPercentMean}{0.34189451293159445}
  \newcommand{\FindAlgoChoiceSACPythonCategoryPercentMean}{0.3679855117335336}
  \newcommand{\FindAlgoChoiceSACSimulatorCategoryPercentMean}{0.05416171387568431}
\fi

%% file: tex/autogenerated/uncorrected/FrameworkChoiceMetrics.tex
\ifx\TexMetricsEnabled\undefined
  \newcommand{\FindQualEagerMoreTransMinEagerSlowdown}{a}
  \newcommand{\FindQualEagerMoreTransMaxEagerSlowdown}{b}
  \newcommand{\FindQualEagerMoreTransMinRatioInferenceFrameworkTransTFToPyTorch}{c}
  \newcommand{\FindQualEagerMoreTransMaxRatioInferenceFrameworkTransTFToPyTorch}{d}
  \newcommand{\FindQualEagerMoreTransMeanRatioInferenceFrameworkTransTFToPyTorch}{e}
  \newcommand{\FindQualEagerMoreTransGeomeanRatioInferenceFrameworkTransTFToPyTorch}{f}
  \newcommand{\FindQualEagerMoreTransStdRatioInferenceFrameworkTransTFToPyTorch}{g}
  \newcommand{\FindQualEagerMoreTransMinRatioBackpropagationFrameworkTransTFToPyTorch}{h}
  \newcommand{\FindQualEagerMoreTransMaxRatioBackpropagationFrameworkTransTFToPyTorch}{i}
  \newcommand{\FindQualEagerMoreTransMeanRatioBackpropagationFrameworkTransTFToPyTorch}{j}
  \newcommand{\FindQualEagerMoreTransGeomeanRatioBackpropagationFrameworkTransTFToPyTorch}{k}
  \newcommand{\FindQualEagerMoreTransStdRatioBackpropagationFrameworkTransTFToPyTorch}{l}
  \newcommand{\FindQualEagerMoreTransMinRatioInferenceFrameworkTFToPyTorch}{m}
  \newcommand{\FindQualEagerMoreTransMaxRatioInferenceFrameworkTFToPyTorch}{n}
  \newcommand{\FindQualEagerMoreTransMeanRatioInferenceFrameworkTFToPyTorch}{o}
  \newcommand{\FindQualEagerMoreTransGeomeanRatioInferenceFrameworkTFToPyTorch}{p}
  \newcommand{\FindQualEagerMoreTransStdRatioInferenceFrameworkTFToPyTorch}{q}
  \newcommand{\FindQualEagerMoreTransMinRatioBackpropagationFrameworkTFToPyTorch}{r}
  \newcommand{\FindQualEagerMoreTransMaxRatioBackpropagationFrameworkTFToPyTorch}{s}
  \newcommand{\FindQualEagerMoreTransMeanRatioBackpropagationFrameworkTFToPyTorch}{t}
  \newcommand{\FindQualEagerMoreTransGeomeanRatioBackpropagationFrameworkTFToPyTorch}{u}
  \newcommand{\FindQualEagerMoreTransStdRatioBackpropagationFrameworkTFToPyTorch}{v}
  \newcommand{\FindQualAutographReducesPythonMinAutographPythonPercentOfOp}{a}
  \newcommand{\FindQualAutographReducesPythonMaxAutographPythonPercentOfOp}{b}
  \newcommand{\FindQualAutographReducesPythonMinGraphPythonPercentOfOp}{c}
  \newcommand{\FindQualAutographReducesPythonMaxGraphPythonPercentOfOp}{d}
  \newcommand{\FindQualAutographReducesPythonMinRatioPythonBackpropagationDDPG}{e}
  \newcommand{\FindQualAutographReducesPythonMaxRatioPythonBackpropagationDDPG}{f}
  \newcommand{\FindQualAutographReducesPythonMeanRatioPythonBackpropagationDDPG}{g}
  \newcommand{\FindQualAutographReducesPythonGeomeanRatioPythonBackpropagationDDPG}{h}
  \newcommand{\FindQualAutographReducesPythonStdRatioPythonBackpropagationDDPG}{i}
  \newcommand{\FindQualAutographReducesPythonMinRatioPythonInferenceDDPG}{j}
  \newcommand{\FindQualAutographReducesPythonMaxRatioPythonInferenceDDPG}{k}
  \newcommand{\FindQualAutographReducesPythonMeanRatioPythonInferenceDDPG}{l}
  \newcommand{\FindQualAutographReducesPythonGeomeanRatioPythonInferenceDDPG}{m}
  \newcommand{\FindQualAutographReducesPythonStdRatioPythonInferenceDDPG}{n}
  \newcommand{\FindQualAutographReducesPythonMinRatioPythonBackpropagationTD}{o}
  \newcommand{\FindQualAutographReducesPythonMaxRatioPythonBackpropagationTD}{p}
  \newcommand{\FindQualAutographReducesPythonMeanRatioPythonBackpropagationTD}{q}
  \newcommand{\FindQualAutographReducesPythonGeomeanRatioPythonBackpropagationTD}{r}
  \newcommand{\FindQualAutographReducesPythonStdRatioPythonBackpropagationTD}{s}
  \newcommand{\FindQualAutographReducesPythonMinRatioPythonInferenceTD}{t}
  \newcommand{\FindQualAutographReducesPythonMaxRatioPythonInferenceTD}{u}
  \newcommand{\FindQualAutographReducesPythonMeanRatioPythonInferenceTD}{v}
  \newcommand{\FindQualAutographReducesPythonGeomeanRatioPythonInferenceTD}{w}
  \newcommand{\FindQualAutographReducesPythonStdRatioPythonInferenceTD}{x}
  \newcommand{\FindSurpDdpgBackpropSlowMeanRatioAutographToGraphBackpropagationDDPG}{a}
  \newcommand{\FindSurpDdpgBackpropSlowMeanRatioAutographToGraphInferenceDDPG}{b}
  \newcommand{\FindSurpDdpgBackpropSlowMeanRatioAutographToGraphBackpropagationTD}{c}
  \newcommand{\FindSurpDdpgBackpropSlowMeanRatioAutographToGraphInferenceTD}{d}
  \newcommand{\FindSurpDdpgBackpropSlowMeanRatioAutographToGraphCUDABackpropagationDDPG}{e}
  \newcommand{\FindSurpDdpgBackpropSlowMeanRatioAutographToGraphPythonBackpropagationDDPG}{f}
  \newcommand{\FindSurpDdpgBackpropSlowMeanRatioAutographToGraphCUDAInferenceDDPG}{g}
  \newcommand{\FindSurpDdpgBackpropSlowMeanRatioAutographToGraphPythonInferenceDDPG}{h}
  \newcommand{\FindSurpDdpgBackpropSlowMeanRatioAutographToGraphCUDABackpropagationTD}{i}
  \newcommand{\FindSurpDdpgBackpropSlowMeanRatioAutographToGraphPythonBackpropagationTD}{j}
  \newcommand{\FindSurpDdpgBackpropSlowMeanRatioAutographToGraphCUDAInferenceTD}{k}
  \newcommand{\FindSurpDdpgBackpropSlowMeanRatioAutographToGraphPythonInferenceTD}{l}
  \newcommand{\FindQualPytorchEagerBetterMinPyTorchEagerSpeedup}{a}
  \newcommand{\FindQualPytorchEagerBetterMaxPyTorchEagerSpeedup}{b}
  \newcommand{\FindQualAutographGraphSimilarMaxAutographSpeedup}{a}
  \newcommand{\FindQualAutographGraphSimilarMaxGraphSpeedup}{b}
  \newcommand{\FindQualAutographGraphSimilarMaxSpeedup}{c}
  \newcommand{\FindSurpExecModelComparisonMinTFSpeedup}{a}
  \newcommand{\FindSurpExecModelComparisonMaxTFSpeedup}{b}
  \newcommand{\FindSurpTotalGpuTimeMinGPUPercent}{a}
  \newcommand{\FindSurpTotalGpuTimeMaxGPUPercent}{b}
  \newcommand{\FindSurpCudaApiDominatesMinRatioCUDAToGPU}{a}
  \newcommand{\FindSurpCudaApiDominatesMaxRatioCUDAToGPU}{b}
  \newcommand{\FindSurpCudaApiDominatesMeanRatioCUDAToGPU}{c}
  \newcommand{\FindSurpCudaApiDominatesGeomeanRatioCUDAToGPU}{d}
  \newcommand{\FindSurpCudaApiDominatesStdRatioCUDAToGPU}{e}
  \newcommand{\FindSurpAutographNoGpuMinRatioGPUAutographToNonautograph}{a}
  \newcommand{\FindSurpAutographNoGpuMaxRatioGPUAutographToNonautograph}{b}
  \newcommand{\FindSurpAutographNoGpuMeanRatioGPUAutographToNonautograph}{c}
  \newcommand{\FindSurpAutographNoGpuGeomeanRatioGPUAutographToNonautograph}{d}
  \newcommand{\FindSurpAutographNoGpuStdRatioGPUAutographToNonautograph}{e}
  \newcommand{\FindSurpAutographInflatesInferenceMinRatioInferenceFrameworkAutographToGraph}{a}
  \newcommand{\FindSurpAutographInflatesInferenceMaxRatioInferenceFrameworkAutographToGraph}{b}
  \newcommand{\FindSurpAutographInflatesInferenceMeanRatioInferenceFrameworkAutographToGraph}{c}
  \newcommand{\FindSurpAutographInflatesInferenceGeomeanRatioInferenceFrameworkAutographToGraph}{d}
  \newcommand{\FindSurpAutographInflatesInferenceStdRatioInferenceFrameworkAutographToGraph}{e}
  \newcommand{\FindSurpAutographInflatesInferenceMeanRatioInferenceFrameworkDDPGAutographToGraph}{f}
  \newcommand{\FindSurpAutographInflatesInferenceMeanRatioInferenceFrameworkTDAutographToGraph}{g}
  \newcommand{\FindSurpAutographInflatesPythonMaxRatioSimulationPythonAutographToEager}{a}
  \newcommand{\FindSurpAutographInflatesPythonMeanRatioSimulationPythonAutographToEagerDDPG}{b}
  \newcommand{\FindSurpAutographInflatesPythonMeanRatioSimulationPythonAutographToEagerTD}{c}
\else
  \newcommand{\FindQualEagerMoreTransMinEagerSlowdown}{1.7305056439463715}
  \newcommand{\FindQualEagerMoreTransMaxEagerSlowdown}{4.598368342263544}
  \newcommand{\FindQualEagerMoreTransMinRatioInferenceFrameworkTransTFToPyTorch}{3.155013832244816}
  \newcommand{\FindQualEagerMoreTransMaxRatioInferenceFrameworkTransTFToPyTorch}{3.155013832244816}
  \newcommand{\FindQualEagerMoreTransMeanRatioInferenceFrameworkTransTFToPyTorch}{3.155013832244816}
  \newcommand{\FindQualEagerMoreTransGeomeanRatioInferenceFrameworkTransTFToPyTorch}{3.155013832244816}
  \newcommand{\FindQualEagerMoreTransStdRatioInferenceFrameworkTransTFToPyTorch}{nan}
  \newcommand{\FindQualEagerMoreTransMinRatioBackpropagationFrameworkTransTFToPyTorch}{1.599702380952381}
  \newcommand{\FindQualEagerMoreTransMaxRatioBackpropagationFrameworkTransTFToPyTorch}{1.599702380952381}
  \newcommand{\FindQualEagerMoreTransMeanRatioBackpropagationFrameworkTransTFToPyTorch}{1.599702380952381}
  \newcommand{\FindQualEagerMoreTransGeomeanRatioBackpropagationFrameworkTransTFToPyTorch}{1.599702380952381}
  \newcommand{\FindQualEagerMoreTransStdRatioBackpropagationFrameworkTransTFToPyTorch}{nan}
  \newcommand{\FindQualEagerMoreTransMinRatioInferenceFrameworkTFToPyTorch}{11.762905935923435}
  \newcommand{\FindQualEagerMoreTransMaxRatioInferenceFrameworkTFToPyTorch}{11.762905935923435}
  \newcommand{\FindQualEagerMoreTransMeanRatioInferenceFrameworkTFToPyTorch}{11.762905935923435}
  \newcommand{\FindQualEagerMoreTransGeomeanRatioInferenceFrameworkTFToPyTorch}{11.762905935923435}
  \newcommand{\FindQualEagerMoreTransStdRatioInferenceFrameworkTFToPyTorch}{nan}
  \newcommand{\FindQualEagerMoreTransMinRatioBackpropagationFrameworkTFToPyTorch}{3.206866615000169}
  \newcommand{\FindQualEagerMoreTransMaxRatioBackpropagationFrameworkTFToPyTorch}{3.206866615000169}
  \newcommand{\FindQualEagerMoreTransMeanRatioBackpropagationFrameworkTFToPyTorch}{3.206866615000169}
  \newcommand{\FindQualEagerMoreTransGeomeanRatioBackpropagationFrameworkTFToPyTorch}{3.206866615000169}
  \newcommand{\FindQualEagerMoreTransStdRatioBackpropagationFrameworkTFToPyTorch}{nan}
  \newcommand{\FindQualAutographReducesPythonMinAutographPythonPercentOfOp}{0.05237954768430697}
  \newcommand{\FindQualAutographReducesPythonMaxAutographPythonPercentOfOp}{0.06309627838641839}
  \newcommand{\FindQualAutographReducesPythonMinGraphPythonPercentOfOp}{0.19297854945330573}
  \newcommand{\FindQualAutographReducesPythonMaxGraphPythonPercentOfOp}{0.5632848879878798}
  \newcommand{\FindQualAutographReducesPythonMinRatioPythonBackpropagationDDPG}{11.547763677104035}
  \newcommand{\FindQualAutographReducesPythonMaxRatioPythonBackpropagationDDPG}{11.547763677104035}
  \newcommand{\FindQualAutographReducesPythonMeanRatioPythonBackpropagationDDPG}{11.547763677104035}
  \newcommand{\FindQualAutographReducesPythonGeomeanRatioPythonBackpropagationDDPG}{11.547763677104035}
  \newcommand{\FindQualAutographReducesPythonStdRatioPythonBackpropagationDDPG}{nan}
  \newcommand{\FindQualAutographReducesPythonMinRatioPythonInferenceDDPG}{2.584217546193086}
  \newcommand{\FindQualAutographReducesPythonMaxRatioPythonInferenceDDPG}{2.584217546193086}
  \newcommand{\FindQualAutographReducesPythonMeanRatioPythonInferenceDDPG}{2.584217546193086}
  \newcommand{\FindQualAutographReducesPythonGeomeanRatioPythonInferenceDDPG}{2.584217546193086}
  \newcommand{\FindQualAutographReducesPythonStdRatioPythonInferenceDDPG}{nan}
  \newcommand{\FindQualAutographReducesPythonMinRatioPythonBackpropagationTD}{3.546300417853453}
  \newcommand{\FindQualAutographReducesPythonMaxRatioPythonBackpropagationTD}{3.546300417853453}
  \newcommand{\FindQualAutographReducesPythonMeanRatioPythonBackpropagationTD}{3.546300417853453}
  \newcommand{\FindQualAutographReducesPythonGeomeanRatioPythonBackpropagationTD}{3.5463004178534536}
  \newcommand{\FindQualAutographReducesPythonStdRatioPythonBackpropagationTD}{nan}
  \newcommand{\FindQualAutographReducesPythonMinRatioPythonInferenceTD}{3.8988723680744646}
  \newcommand{\FindQualAutographReducesPythonMaxRatioPythonInferenceTD}{3.8988723680744646}
  \newcommand{\FindQualAutographReducesPythonMeanRatioPythonInferenceTD}{3.8988723680744646}
  \newcommand{\FindQualAutographReducesPythonGeomeanRatioPythonInferenceTD}{3.898872368074465}
  \newcommand{\FindQualAutographReducesPythonStdRatioPythonInferenceTD}{nan}
  \newcommand{\FindSurpDdpgBackpropSlowMeanRatioAutographToGraphBackpropagationDDPG}{3.1455914542436902}
  \newcommand{\FindSurpDdpgBackpropSlowMeanRatioAutographToGraphInferenceDDPG}{0.7014258660919}
  \newcommand{\FindSurpDdpgBackpropSlowMeanRatioAutographToGraphBackpropagationTD}{0.8619596568959995}
  \newcommand{\FindSurpDdpgBackpropSlowMeanRatioAutographToGraphInferenceTD}{0.39768689368950083}
  \newcommand{\FindSurpDdpgBackpropSlowMeanRatioAutographToGraphCUDABackpropagationDDPG}{2.389004121179645}
  \newcommand{\FindSurpDdpgBackpropSlowMeanRatioAutographToGraphPythonBackpropagationDDPG}{11.547763677104035}
  \newcommand{\FindSurpDdpgBackpropSlowMeanRatioAutographToGraphCUDAInferenceDDPG}{0.7730556223566809}
  \newcommand{\FindSurpDdpgBackpropSlowMeanRatioAutographToGraphPythonInferenceDDPG}{2.584217546193086}
  \newcommand{\FindSurpDdpgBackpropSlowMeanRatioAutographToGraphCUDABackpropagationTD}{0.8817469934175527}
  \newcommand{\FindSurpDdpgBackpropSlowMeanRatioAutographToGraphPythonBackpropagationTD}{3.546300417853453}
  \newcommand{\FindSurpDdpgBackpropSlowMeanRatioAutographToGraphCUDAInferenceTD}{0.10319444996724406}
  \newcommand{\FindSurpDdpgBackpropSlowMeanRatioAutographToGraphPythonInferenceTD}{3.8988723680744646}
  \newcommand{\FindQualPytorchEagerBetterMinPyTorchEagerSpeedup}{1.8726833861689574}
  \newcommand{\FindQualPytorchEagerBetterMaxPyTorchEagerSpeedup}{1.8726833861689574}
  \newcommand{\FindQualAutographGraphSimilarMaxAutographSpeedup}{0.3525108512249642}
  \newcommand{\FindQualAutographGraphSimilarMaxGraphSpeedup}{0.41894766894746993}
  \newcommand{\FindQualAutographGraphSimilarMaxSpeedup}{0.41894766894746993}
  \newcommand{\FindSurpExecModelComparisonMinTFSpeedup}{1.7305056439463715}
  \newcommand{\FindSurpExecModelComparisonMaxTFSpeedup}{2.4554969495781442}
  \newcommand{\FindSurpTotalGpuTimeMinGPUPercent}{0.01535945516084529}
  \newcommand{\FindSurpTotalGpuTimeMaxGPUPercent}{0.10384931583912607}
  \newcommand{\FindSurpCudaApiDominatesMinRatioCUDAToGPU}{1.195889849143247}
  \newcommand{\FindSurpCudaApiDominatesMaxRatioCUDAToGPU}{10.752731571655387}
  \newcommand{\FindSurpCudaApiDominatesMeanRatioCUDAToGPU}{5.6524353682159925}
  \newcommand{\FindSurpCudaApiDominatesGeomeanRatioCUDAToGPU}{4.832639520309643}
  \newcommand{\FindSurpCudaApiDominatesStdRatioCUDAToGPU}{2.6554692038731345}
  \newcommand{\FindSurpAutographNoGpuMinRatioGPUAutographToNonautograph}{0.49360871145526874}
  \newcommand{\FindSurpAutographNoGpuMaxRatioGPUAutographToNonautograph}{1.2645218294738434}
  \newcommand{\FindSurpAutographNoGpuMeanRatioGPUAutographToNonautograph}{0.8826934214998957}
  \newcommand{\FindSurpAutographNoGpuGeomeanRatioGPUAutographToNonautograph}{0.843152637068627}
  \newcommand{\FindSurpAutographNoGpuStdRatioGPUAutographToNonautograph}{0.28302394416734444}
  \newcommand{\FindSurpAutographInflatesInferenceMinRatioInferenceFrameworkAutographToGraph}{2.212317755094514}
  \newcommand{\FindSurpAutographInflatesInferenceMaxRatioInferenceFrameworkAutographToGraph}{3.9326091608115323}
  \newcommand{\FindSurpAutographInflatesInferenceMeanRatioInferenceFrameworkAutographToGraph}{3.072463457953023}
  \newcommand{\FindSurpAutographInflatesInferenceGeomeanRatioInferenceFrameworkAutographToGraph}{2.9496069348831364}
  \newcommand{\FindSurpAutographInflatesInferenceStdRatioInferenceFrameworkAutographToGraph}{1.216429718599442}
  \newcommand{\FindSurpAutographInflatesInferenceMeanRatioInferenceFrameworkDDPGAutographToGraph}{2.212317755094514}
  \newcommand{\FindSurpAutographInflatesInferenceMeanRatioInferenceFrameworkTDAutographToGraph}{3.9326091608115323}
  \newcommand{\FindSurpAutographInflatesPythonMaxRatioSimulationPythonAutographToEager}{1.5836380826253553}
  \newcommand{\FindSurpAutographInflatesPythonMeanRatioSimulationPythonAutographToEagerDDPG}{1.5836380826253553}
  \newcommand{\FindSurpAutographInflatesPythonMeanRatioSimulationPythonAutographToEagerTD}{1.1591156215570124}
\fi

%% file: tex/autogenerated/uncorrected/AlgoChoiceMetrics.tex
\ifx\TexMetricsEnabled\undefined
  \newcommand{\FindAlgoChoiceMinRatioPercentOnPolicyToOffPolicyBackpropagation}{a}
  \newcommand{\FindAlgoChoiceMaxRatioPercentOnPolicyToOffPolicyBackpropagation}{b}
  \newcommand{\FindAlgoChoiceMeanRatioPercentOnPolicyToOffPolicyBackpropagation}{c}
  \newcommand{\FindAlgoChoiceMinRatioPercentOnPolicyToOffPolicyInference}{d}
  \newcommand{\FindAlgoChoiceMaxRatioPercentOnPolicyToOffPolicyInference}{e}
  \newcommand{\FindAlgoChoiceMeanRatioPercentOnPolicyToOffPolicyInference}{f}
  \newcommand{\FindAlgoChoiceMinRatioPercentOnPolicyToOffPolicySimulation}{g}
  \newcommand{\FindAlgoChoiceMaxRatioPercentOnPolicyToOffPolicySimulation}{h}
  \newcommand{\FindAlgoChoiceMeanRatioPercentOnPolicyToOffPolicySimulation}{i}
  \newcommand{\FindAlgoChoiceACBackpropagationOpPercentMean}{j}
  \newcommand{\FindAlgoChoiceACInferenceOpPercentMean}{k}
  \newcommand{\FindAlgoChoiceACSimulationOpPercentMean}{l}
  \newcommand{\FindAlgoChoiceDDPGBackpropagationOpPercentMean}{m}
  \newcommand{\FindAlgoChoiceDDPGInferenceOpPercentMean}{n}
  \newcommand{\FindAlgoChoiceDDPGSimulationOpPercentMean}{o}
  \newcommand{\FindAlgoChoicePPOBackpropagationOpPercentMean}{p}
  \newcommand{\FindAlgoChoicePPOInferenceOpPercentMean}{q}
  \newcommand{\FindAlgoChoicePPOSimulationOpPercentMean}{r}
  \newcommand{\FindAlgoChoiceSACBackpropagationOpPercentMean}{s}
  \newcommand{\FindAlgoChoiceSACInferenceOpPercentMean}{t}
  \newcommand{\FindAlgoChoiceSACSimulationOpPercentMean}{u}
  \newcommand{\FindAlgoChoiceACCpuResourcePercentMean}{v}
  \newcommand{\FindAlgoChoiceACGpuResourcePercentMean}{w}
  \newcommand{\FindAlgoChoiceDDPGCpuResourcePercentMean}{x}
  \newcommand{\FindAlgoChoiceDDPGGpuResourcePercentMean}{y}
  \newcommand{\FindAlgoChoicePPOCpuResourcePercentMean}{z}
  \newcommand{\FindAlgoChoicePPOGpuResourcePercentMean}{A}
  \newcommand{\FindAlgoChoiceSACCpuResourcePercentMean}{B}
  \newcommand{\FindAlgoChoiceSACGpuResourcePercentMean}{C}
  \newcommand{\FindAlgoChoiceACBackpropagationCpuResourcePercentMean}{D}
  \newcommand{\FindAlgoChoiceACBackpropagationGpuResourcePercentMean}{E}
  \newcommand{\FindAlgoChoiceACInferenceCpuResourcePercentMean}{F}
  \newcommand{\FindAlgoChoiceACInferenceGpuResourcePercentMean}{G}
  \newcommand{\FindAlgoChoiceDDPGBackpropagationCpuResourcePercentMean}{H}
  \newcommand{\FindAlgoChoiceDDPGBackpropagationGpuResourcePercentMean}{I}
  \newcommand{\FindAlgoChoiceDDPGInferenceCpuResourcePercentMean}{J}
  \newcommand{\FindAlgoChoiceDDPGInferenceGpuResourcePercentMean}{K}
  \newcommand{\FindAlgoChoicePPOBackpropagationCpuResourcePercentMean}{L}
  \newcommand{\FindAlgoChoicePPOBackpropagationGpuResourcePercentMean}{M}
  \newcommand{\FindAlgoChoicePPOInferenceCpuResourcePercentMean}{N}
  \newcommand{\FindAlgoChoicePPOInferenceGpuResourcePercentMean}{O}
  \newcommand{\FindAlgoChoiceSACBackpropagationCpuResourcePercentMean}{P}
  \newcommand{\FindAlgoChoiceSACBackpropagationGpuResourcePercentMean}{Q}
  \newcommand{\FindAlgoChoiceSACInferenceCpuResourcePercentMean}{R}
  \newcommand{\FindAlgoChoiceSACInferenceGpuResourcePercentMean}{S}
  \newcommand{\FindAlgoChoiceACBackendCategoryPercentMean}{T}
  \newcommand{\FindAlgoChoiceACCUDACategoryPercentMean}{U}
  \newcommand{\FindAlgoChoiceACPythonCategoryPercentMean}{V}
  \newcommand{\FindAlgoChoiceACSimulatorCategoryPercentMean}{W}
  \newcommand{\FindAlgoChoiceDDPGBackendCategoryPercentMean}{X}
  \newcommand{\FindAlgoChoiceDDPGCUDACategoryPercentMean}{Y}
  \newcommand{\FindAlgoChoiceDDPGPythonCategoryPercentMean}{Z}
  \newcommand{\FindAlgoChoiceDDPGSimulatorCategoryPercentMean}{aa}
  \newcommand{\FindAlgoChoicePPOBackendCategoryPercentMean}{bb}
  \newcommand{\FindAlgoChoicePPOCUDACategoryPercentMean}{cc}
  \newcommand{\FindAlgoChoicePPOPythonCategoryPercentMean}{dd}
  \newcommand{\FindAlgoChoicePPOSimulatorCategoryPercentMean}{ee}
  \newcommand{\FindAlgoChoiceSACBackendCategoryPercentMean}{ff}
  \newcommand{\FindAlgoChoiceSACCUDACategoryPercentMean}{gg}
  \newcommand{\FindAlgoChoiceSACPythonCategoryPercentMean}{hh}
  \newcommand{\FindAlgoChoiceSACSimulatorCategoryPercentMean}{ii}
\else
  \newcommand{\FindAlgoChoiceMinRatioPercentOnPolicyToOffPolicyBackpropagation}{0.115270974748685}
  \newcommand{\FindAlgoChoiceMaxRatioPercentOnPolicyToOffPolicyBackpropagation}{0.5480925369272535}
  \newcommand{\FindAlgoChoiceMeanRatioPercentOnPolicyToOffPolicyBackpropagation}{0.3176344050148199}
  \newcommand{\FindAlgoChoiceMinRatioPercentOnPolicyToOffPolicyInference}{0.8209371165030651}
  \newcommand{\FindAlgoChoiceMaxRatioPercentOnPolicyToOffPolicyInference}{2.591661152637387}
  \newcommand{\FindAlgoChoiceMeanRatioPercentOnPolicyToOffPolicyInference}{1.6168360509757207}
  \newcommand{\FindAlgoChoiceMinRatioPercentOnPolicyToOffPolicySimulation}{3.4267244965196086}
  \newcommand{\FindAlgoChoiceMaxRatioPercentOnPolicyToOffPolicySimulation}{5.161615929856052}
  \newcommand{\FindAlgoChoiceMeanRatioPercentOnPolicyToOffPolicySimulation}{4.290052048847504}
  \newcommand{\FindAlgoChoiceACBackpropagationOpPercentMean}{0.09042924978769398}
  \newcommand{\FindAlgoChoiceACInferenceOpPercentMean}{0.19748697610972946}
  \newcommand{\FindAlgoChoiceACSimulationOpPercentMean}{0.7120837741025766}
  \newcommand{\FindAlgoChoiceDDPGBackpropagationOpPercentMean}{0.6782135328510118}
  \newcommand{\FindAlgoChoiceDDPGInferenceOpPercentMean}{0.18382893916173343}
  \newcommand{\FindAlgoChoiceDDPGSimulationOpPercentMean}{0.13795752798725483}
  \newcommand{\FindAlgoChoicePPOBackpropagationOpPercentMean}{0.3717237757987062}
  \newcommand{\FindAlgoChoicePPOInferenceOpPercentMean}{0.15091199924525084}
  \newcommand{\FindAlgoChoicePPOSimulationOpPercentMean}{0.4773642249560429}
  \newcommand{\FindAlgoChoiceSACBackpropagationOpPercentMean}{0.7844928004196096}
  \newcommand{\FindAlgoChoiceSACInferenceOpPercentMean}{0.07620092461113528}
  \newcommand{\FindAlgoChoiceSACSimulationOpPercentMean}{0.13930627496925513}
  \newcommand{\FindAlgoChoiceACCpuResourcePercentMean}{0.9868115765868581}
  \newcommand{\FindAlgoChoiceACGpuResourcePercentMean}{0.013188423413141902}
  \newcommand{\FindAlgoChoiceDDPGCpuResourcePercentMean}{0.9316787860351122}
  \newcommand{\FindAlgoChoiceDDPGGpuResourcePercentMean}{0.06832121396488774}
  \newcommand{\FindAlgoChoicePPOCpuResourcePercentMean}{0.9669810818460296}
  \newcommand{\FindAlgoChoicePPOGpuResourcePercentMean}{0.03301891815397045}
  \newcommand{\FindAlgoChoiceSACCpuResourcePercentMean}{0.9323911957551164}
  \newcommand{\FindAlgoChoiceSACGpuResourcePercentMean}{0.06760880424488361}
  \newcommand{\FindAlgoChoiceACBackpropagationCpuResourcePercentMean}{0.966780697406984}
  \newcommand{\FindAlgoChoiceACBackpropagationGpuResourcePercentMean}{0.033219302593015886}
  \newcommand{\FindAlgoChoiceACInferenceCpuResourcePercentMean}{0.948429881292393}
  \newcommand{\FindAlgoChoiceACInferenceGpuResourcePercentMean}{0.051570118707606916}
  \newcommand{\FindAlgoChoiceDDPGBackpropagationCpuResourcePercentMean}{0.9204480048320827}
  \newcommand{\FindAlgoChoiceDDPGBackpropagationGpuResourcePercentMean}{0.07955199516791737}
  \newcommand{\FindAlgoChoiceDDPGInferenceCpuResourcePercentMean}{0.9218405200931563}
  \newcommand{\FindAlgoChoiceDDPGInferenceGpuResourcePercentMean}{0.07815947990684373}
  \newcommand{\FindAlgoChoicePPOBackpropagationCpuResourcePercentMean}{0.9278692449398025}
  \newcommand{\FindAlgoChoicePPOBackpropagationGpuResourcePercentMean}{0.07213075506019746}
  \newcommand{\FindAlgoChoicePPOInferenceCpuResourcePercentMean}{0.958875360721341}
  \newcommand{\FindAlgoChoicePPOInferenceGpuResourcePercentMean}{0.04112463927865907}
  \newcommand{\FindAlgoChoiceSACBackpropagationCpuResourcePercentMean}{0.9168183252447575}
  \newcommand{\FindAlgoChoiceSACBackpropagationGpuResourcePercentMean}{0.0831816747552425}
  \newcommand{\FindAlgoChoiceSACInferenceCpuResourcePercentMean}{0.9691161323230669}
  \newcommand{\FindAlgoChoiceSACInferenceGpuResourcePercentMean}{0.03088386767693316}
  \newcommand{\FindAlgoChoiceACBackendCategoryPercentMean}{0.09743167385931133}
  \newcommand{\FindAlgoChoiceACCUDACategoryPercentMean}{0.0827962568560451}
  \newcommand{\FindAlgoChoiceACPythonCategoryPercentMean}{0.5338958489939756}
  \newcommand{\FindAlgoChoiceACSimulatorCategoryPercentMean}{0.28587622029066795}
  \newcommand{\FindAlgoChoiceDDPGBackendCategoryPercentMean}{0.27053612194711674}
  \newcommand{\FindAlgoChoiceDDPGCUDACategoryPercentMean}{0.36851833948624235}
  \newcommand{\FindAlgoChoiceDDPGPythonCategoryPercentMean}{0.29628420023537266}
  \newcommand{\FindAlgoChoiceDDPGSimulatorCategoryPercentMean}{0.06466133833126828}
  \newcommand{\FindAlgoChoicePPOBackendCategoryPercentMean}{0.17409155354135128}
  \newcommand{\FindAlgoChoicePPOCUDACategoryPercentMean}{0.21494529463320539}
  \newcommand{\FindAlgoChoicePPOPythonCategoryPercentMean}{0.4128270457928353}
  \newcommand{\FindAlgoChoicePPOSimulatorCategoryPercentMean}{0.19813610603260806}
  \newcommand{\FindAlgoChoiceSACBackendCategoryPercentMean}{0.2638748187586117}
  \newcommand{\FindAlgoChoiceSACCUDACategoryPercentMean}{0.35068845199229054}
  \newcommand{\FindAlgoChoiceSACPythonCategoryPercentMean}{0.3268570078458032}
  \newcommand{\FindAlgoChoiceSACSimulatorCategoryPercentMean}{0.058579721403294704}
\fi

%% file: tex/hardcoded_numbers.tex
\newcommand{\HRDCalibValidRatioInflationMax}{1.902439}
\newcommand{\HRDCalibValidPercentInflationMax}{0.90243902}
\newcommand{\HRDCalibValidPercentLeft}{0.16}

\newcommand{\HRDSimulatorComparisonWalkerDGpuPercent}{0.027908293}
\newcommand{\HRDSimulatorComparisonPongGpuPercent}{0.06948511}

\newcommand{\HRDSimulatorComparisonAirLearningSimulationPercent}{0.995655817}

\newcommand{\HRDSimulatorComparisonPongSimulationPercent}{0.741525424}
\newcommand{\HRDSimulatorComparisonWalkerDSimulationPercent}{0.711864407}
\newcommand{\HRDSimulatorComparisonAntSimulationPercent}{0.478813559}
\newcommand{\HRDSimulatorComparisonHalfCheetahSimulationPercent}{0.381355932}

\newcommand{\HRDSimulatorComparisonAtLeastSimulationBoundPercent}{\HRDSimulatorComparisonHalfCheetahSimulationPercent}
\newcommand{\HRDSimulatorComparisonAtMostSimulationBoundPercent}{\HRDSimulatorComparisonPongSimulationPercent}

%% file: tex/abstract.tex
\begin{abstract}

Deep reinforcement learning (RL) has made groundbreaking advancements in robotics, data center management and other applications.
Unfortunately, system-level bottlenecks in RL workloads are poorly understood; we observe fundamental structural differences in RL workloads that make them inherently less GPU-bound than supervised learning (SL).
To explain where training time is spent in RL workloads, we propose \rlscope, a cross-stack profiler that scopes low-level CPU/GPU resource usage to high-level algorithmic operations, and provides accurate insights by correcting for profiling overhead.
Using \rlscope, we survey RL workloads across its major dimensions including ML backend, RL algorithm, and simulator.
For ML backends, we explain a \FCAsTimes{\FindQualPytorchEagerBetterMinPyTorchEagerSpeedup} difference in runtime between equivalent PyTorch and TensorFlow algorithm implementations, and identify a bottleneck rooted in overly abstracted algorithm implementations.
For RL algorithms and simulators, we show that on-policy algorithms are at least \FCAsTimes{\FindAlgoChoiceMinRatioPercentOnPolicyToOffPolicySimulation} more simulation-bound than off-policy algorithms.
Finally, we profile a scale-up workload and demonstrate that GPU utilization metrics reported by commonly used tools dramatically inflate GPU usage, whereas \rlscope reports true GPU-bound time.
\rlscope is an open-source tool available at \githubURL.

\ifx\AddPageCount\undefined
\else
{\hfill\color{red} $\Rightarrow$ \textbf{Num pages = \pageref*{page:last-page}} $\Leftarrow$}
\fi

\end{abstract}

%% file: tex/intro.tex
\section{Introduction}
\label{sec:intro}

Deep Reinforcement Learning (RL) has made many algorithmic advancements in the past
decade\CiteExtra{mnih2016asynchronous,wang2016sample,wu2017scalable,lillicrap2015continuous,ho2016generative,andrychowicz2017hindsight,schulman2017proximal,schulman2015trust},  
such as 
learning to play Atari games from raw pixels~\cite{mnih2015human}, 
surpassing human performance on games with intractably large state spaces~\cite{nature-agz},
as well as in diverse industrial applications including robotics~\cite{brockman2016openai}, data center management~\cite{datacenterRL}, autonomous driving~\cite{dosovitskiy2017carla}, and drone tasks~\cite{krishnan2019air}.

Despite their promise, RL models are notoriously slow to train with AlphaGoZero taking 40 days to train~\cite{nature-agz}; 
to make matters worse, 
most benchmark studies of RL training focus only on model accuracy as a function of training steps~\cite{duan2016benchmarking,fujimoto2019benchmarking}, not 
training runtime
leading the community to \emph{believe} they behave similarly to supervised learning (SL) workloads.
In SL training workloads like image recognition, 
neural networks are large (e.g., 152 layers~\cite{resnet}), 
and runtime is largely accelerator-bound,  
spending 
at least \asPercent{0.5368} 
of total training time executing on the GPU~\cite{li2020xsp}.
In contrast to SL workloads, we observe that RL workloads are fundamentally different in structure, spending a large portion of their training time on the CPU collecting training data through simulation, running high-level language code (e.g., Python) inside the training loop, and executing relatively small neural networks (e.g., AlphaGoZero~\cite{nature-agz} has only 39 layers).

Identifying the precise reasons for poor RL training runtime remains a challenge since existing profiling tools are designed for GPU-bound SL workloads and are not suitable for RL.
General purpose GPU profiling tools~\cite{nvidia-nsight-systems} are unsuitable for two reasons.  First, they mainly focus on GPU kernel metrics, and low-level system calls with little context about which operation in high-level code they originate from.
Second, they make no effort to correct for CPU profiling overhead that can inflate RL workloads significantly; we observe up to a \FCAsTimes{\HRDCalibValidRatioInflationMax} inflation of total training time when profiling RL workloads.
Even specialized machine learning (ML) profiling tools only target GPU-bound SL workloads and limit their analysis to total training time spent in neural network layers and bottleneck kernels within each layer~\cite{nvidia-dlprof,li2020xsp}.
In contrast, deep RL training workloads execute a diverse software stack that includes algorithms, simulators, ML backends, accelerator APIs, and GPU kernels, that are written in a mix of high- and low-level languages, and are executed asynchronously across single and even multiple processes resulting in CPU/GPU overlap.
Hence, identifying bottlenecks in RL workloads requires a \emph{cross-stack} profiler that can accurately separate GPU and CPU resource usage at each different level, and correlate low-level CPU/GPU execution time with RL algorithmic operations.

We present 
\emph{\rlscope}\ -- an open-source, accurate, cross-stack, cross-backend tool for profiling RL workloads to address this challenge.
\rlscope provides ML researchers and developers with a simple Python API for annotating their high-level code with
operations.
\rlscope collects cross-stack profiling information by scoping time spent in low-level GPU kernels, CUDA API calls, simulators, and ML backends to user annotations in high-level code.
To ensure accurate insights during offline analysis, \rlscope calibrates and corrects CPU time inflation introduced by book-keeping code, correcting within 
\asPercentNoDecimal{\HRDCalibValidPercentLeft} of profiling overhead.
\rlscope easily supports multiple ML backends (TensorFlow~\cite{abadi2016tensorflow} and PyTorch~\cite{pytorch}) and simulators, as it requires no recompilation of ML backends or simulators.

In contrast to prior studies that have been limited to GPU-bound SL training workloads~\cite{li2020xsp} and RL model accuracy as a function of training steps~\cite{duan2016benchmarking,fujimoto2019benchmarking}, we use \rlscope to perform the first cross-stack study of RL training runtime.
In our first case study, we use \rlscope to 
compare RL frameworks and analyze bottlenecks.
\rlscope can help practitioners decide between a myriad of potential RL frameworks spanning multiple ML backends with diverse execution models
ranging from developer-friendly \eager execution popularized by PyTorch to the more sophisticated \autograph execution in TensorFlow
that converts high-level code to in-graph operators.
\rlscope collects fine-grained metrics that quantify the intuition of why different execution models like \eager perform up to \FCAsTimes{\FindQualEagerMoreTransMaxEagerSlowdown} worse than \graph/\autograph\IntroRefFnd{find:qual-eager-more-trans}, and uses these metrics to explain further a \FCAsTimes{\FindQualPytorchEagerBetterMinPyTorchEagerSpeedup} difference in runtime between a TensorFlow implementation and PyTorch implementation of the \eager execution model\IntroRefFnd{find:qual-pytorch-eager-better}.
\rlscope can detect and analyze bottlenecks in RL algorithm implementations; \rlscope's metrics detect a \FCAsTimes{\FindSurpDdpgBackpropSlowMeanRatioAutographToGraphBackpropagationDDPG} inflation in backpropagation rooted in an overly abstracted MPI-friendly but very GPU-unfriendly Adam optimizer\IntroRefFnd{find:surp-ddpg-backprop-slow}.
Finally, \rlscope illustrates that GPU usage is low across \emph{all} RL frameworks regardless of ML backend or execution model\IntroRefFnd{find:surp-total-gpu-time}.

In our next case study, 
we 
survey RL workloads across different RL algorithms and simulators, 
which can help researchers decide where to devote their efforts in optimizing RL workloads by surveying how training bottlenecks change.
We observe that simulation time is non-negligible, taking up 
at least \asPercent{\HRDSimulatorComparisonAtLeastSimulationBoundPercent} and 
at most \asPercent{\HRDSimulatorComparisonAtMostSimulationBoundPercent} of training runtime, 
with simulation time being higher for on-policy RL algorithms typically used in robotics tasks\IntroRefFnd{find:algo-choice}.
Operations considered GPU-heavy in SL workloads (i.e., inference, backpropagation) only spend at most \FCAsPercent{\FindAlgoChoiceDDPGInferenceGpuResourcePercentMean} of their time executing GPU kernels, with the rest spent on the CPU in ML backend and CUDA API calls\IntroRefFnd{find:software-overhead}.  

Our final case study
examines a 
scale-up RL workload
that increases GPU utilization by parallelizing inference operations.  
We find that coarse-grained GPU utilization metrics collected using common tools (e.g., \texttt{nvidia-smi}) can be a poor indicator of actual time spent executing GPU kernels on the GPU in the RL context, whereas \rlscope is able to identify the true GPU-bound time\IntroRefFnd{find:gpu-util}.

\iftrue
{

To summarize, our contributions are:
\begin{itemize}[nosep, leftmargin=*]

    \item 
    We observe fundamental structural differences between RL and SL training workloads 
    that leads to more time spent CPU-bound in RL 
    and makes them poorly suited to existing profiling tools that target GPU-bound SL workloads (Section~\ref{sec:background}).

    \item 
    We propose \rlscope: the first profiler designed specifically for RL workloads, featuring: (i) scoping of high- and low-level GPU and CPU time across the stack to algorithmic operations, 
    and (ii) low CPU profiling overhead, and (iii) support for multiple ML backends and simulators (Section~\ref{sec:iml-profiler}).
    To help the community measure their own RL workloads, we have open-sourced \rlscope at \githubURL.

    \item 
    The first cross-stack study
    of RL training runtime (i) across multiple RL frameworks and ML backends, (ii) across multiple RL algorithms and simulators, (iii) in scale-up RL workloads.  
    Prior studies have been limited to SL training workloads, and RL model accuracy as a function of training steps (Section~\ref{sec:visualization}).

\end{itemize}
}
\fi

%% file: tex/background.tex
\section{Background and Motivation}
\label{sec:background}

\begin{figure*}[t]
\captionsetup[subfigure]{position=b}
\centering
\subcaptionbox{Supervised learning (SL) training loop (e.g., image recognition) 
\label{fig:15_regular_workload}}
{\centering
\includegraphics[align=c,width=.35\textwidth]{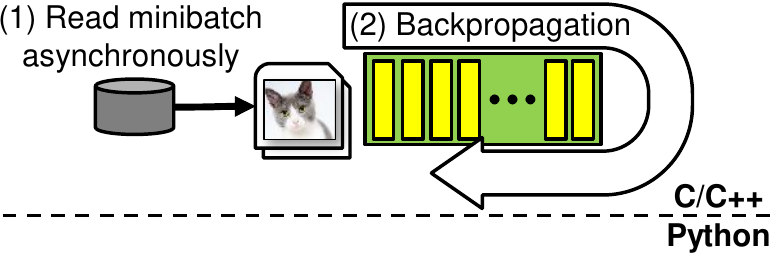}}
\hspace{1em}
\subcaptionbox{Reinforcement learning (RL) training loop (e.g., Atari Pong trained with DQN)
\label{fig:16_irregular_workload}}
{\centering
\includegraphics[align=c,width=.55\textwidth]{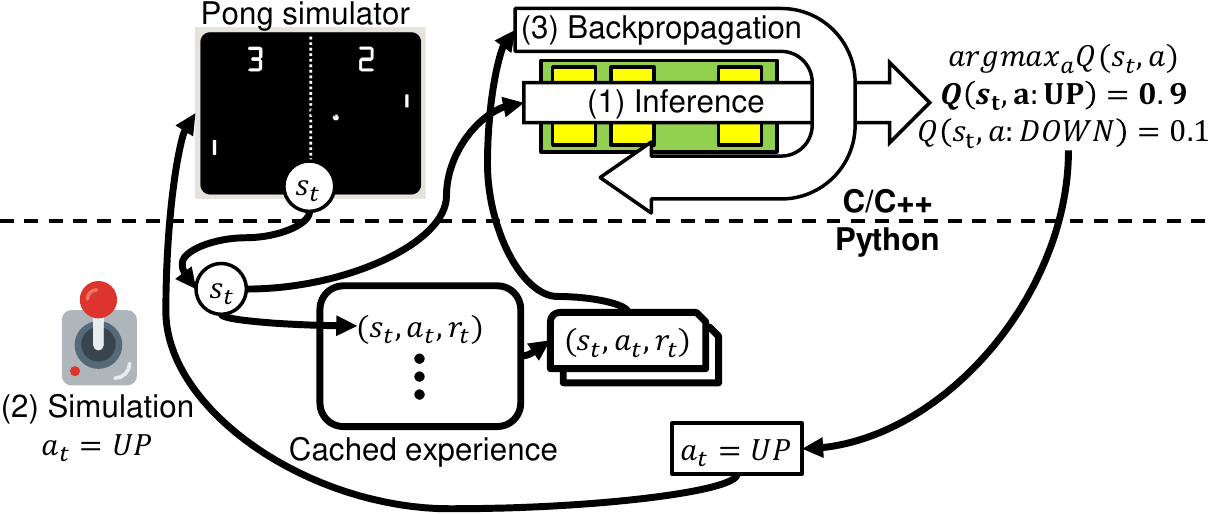}}
\caption{
\textit{Training loop comparison:} the SL training loop is more GPU-bound, whereas RL computation is spread across the software and hardware stack.
Supervised learning workloads (\ref{fig:15_regular_workload}) spend most of their time GPU-bound: they stream training data into large-scale neural networks on the GPU, while high-level language is kept out of the training loop. 
Conversely, RL workloads frequently transition between high-level $\leftrightarrow$ native code in their training loop, collect training data at runtime from a simulator, and use small neural networks.
}

\label{fig:reg-vs-irreg-workloads}
\end{figure*}

Using a simplified
version of DQN~\cite{mnih2015human} on the Atari Pong simulator as an example RL workload, we outline the typical training loop of an RL workload, and showcase key differences between RL and supervised learning (SL) workloads.
These differences inform our design of the \rlscope profiling toolkit for RL workloads.

\subsection{DQN -- an Example RL Workload}
\label{sec:dqn}

The DQN algorithm's goal is to learn to estimate the Q-value function $Q(s, a)$: the expected \emph{reward} if at state $s$ an agent takes action $a$, and repeats this until the simulation terminates. 
Pong's reward would be 1 if the agent won the game or 0 if it lost.  
DQN learns $Q(s, a)$ by constructing training minibatches sampled from a cache of \emph{experience tuples} -- past states, actions taken, estimated rewards, and resulting rewards -- and applying the standard backpropagation training algorithm.
Once learned, a deployed model runs inference to obtain $Q(s, a)$ and greedily selects the action that maximizes the expected reward.

Code for training RL algorithms is often implemented in a high-level language such as Python, with calls to a native simulator (e.g., Atari emulator) and ML backend library (e.g., TensorFlow).
Hence, several steps require \emph{transitioning} from the high-level language to a native library and marshaling data between them.
Other RL algorithms have a similar overall structure and can be grouped into the same components as DQN.
Figure~\ref{fig:16_irregular_workload} illustrates the \textit{high-level algorithmic operations} of the DQN training loop:%

\setlist{nolistsep}
\begin{enumerate}[noitemsep]
\item \textbf{Inference:} Given the latest state $s_t$ as input (e.g., rendered pixels from the Atari emulator), perform an inference operation to predict $Q(s_t, a)$ for each possible action $a$.
To ensure adequate exploration and convergence of $Q(s, a)$, we occasionally select a random action with probability $\epsilon$ or the greedy action with the highest reward $Q(s, a)$ with probability $1-\epsilon$.

\item \textbf{Simulation:} Call the simulator (e.g., Atari emulator C++ library) using the selected action (e.g., $a_t = UP$) as input, and receive the next state $s_{t+1}$. 

\item \textbf{Backpropagation:} Form a minibatch of $Q(s,a)$ samples with predicted and actual rewards from cached experience tuples, and perform the forward, background, and gradient updates of the Q-value network by calling into the C++ ML backend.

\end{enumerate}

\subsection{Comparing SL and RL Training}
\label{sec:compare-sl-and-rl}

Comparing the training loop of supervised learning (SL) (Figure~\ref{fig:15_regular_workload}) to the training loop of RL (Figure~\ref{fig:16_irregular_workload}), clear differences in CPU/GPU execution become apparent.

First, RL workloads collect training data at runtime by running a simulator inside the training loop, dramatically increasing CPU runtime.
In contrast, SL workloads are more GPU-bound since they use large pre-labeled training datasets\CiteExtra{deng2009imagenet} that can be streamed asynchronously onto the GPU while forward/backward GPU compute passes are in progress.

Second, RL workloads run high-level language code inside the training loop.
RL workloads use a high-level language (e.g., Python) to orchestrate collecting training data from simulators and store them in a data structure (\emph{replay buffer}), to be later sampled from by high-level code at each backpropagation training step.
In contrast, SL workloads statically define a computational graph ahead of time, then offload the execution of forward/backward passes to the ML backend without involving high-level code.

Finally, the deep neural networks in most RL workloads are smaller than typically used for SL workloads like image recognition.
SL workloads consist of an extremely large number of layers, making their forward/backward training loop heavily GPU-bound. 
For instance, ResNet-152~\cite{resnet}, a popular image recognition model, has 152 layers, whereas the model used in AlphaGoZero~\cite{nature-agz} only contains 39 layers.
Hence, while the sizes of networks in RL workloads are increasing, they have not reached near the capacity of SL workloads, and as a result, they are not usually bottlenecked by GPU runtime.

Due to these characteristics, the RL training loop is less GPU-bound and time spent on the CPU  plays a crucial role on execution time. 
The inevitable question is how much of an RL workload is CPU-bound, and how much is GPU-bound?  
How does this breakdown depend on the choice of algorithm and simulator?
Which particular high-level algorithmic operation is responsible for the most CPU latency, or the most GPU latency? 
Do excessive language \emph{transitions} contribute to increased training time?
\rlscope is designed to answer these types of questions.

%% file: tex/rlscope_profiler.tex
\section{\rlscope Cross-Stack Profiler}
\label{sec:iml-profiler}

The \rlscope profiler satisfies three high-level design goals.
First, \rlscope provides a cross-stack view of CPU/GPU resource usage by collecting information at each layer of the RL software stack and scoping fine-grained latency information to high-level algorithmic operations.
Second, to ease adoption, \rlscope does not require recompilation or modification of ML backends or simulators, and as a result, can be ported to new simulators and future ML backends.
Finally, \rlscope corrects for profiling overhead to ensure accurate critical path latency measurements.

To achieve these goals, \rlscope adopts several key implementation choices:

\begin{enumerate}
\item
\textbf{High-level algorithmic annotations:} \rlscope provides the developer with a simple API for annotating their code with high-level algorithmic operations.%

\item
\textbf{Transparent event interception:} \rlscope uses transparent hooks to intercept profiling events across all other stack levels, requiring no additional effort from the user.  

\item
\textbf{Cross-stack event overlap:} \rlscope sums up regions of overlap between cross-stack events, which simultaneously measures overlap between CPU and GPU resources, and scopes training time to high-level algorithmic operations.

\item
\textbf{Profiling calibration and overhead correction:}  
\rlscope runs the RL training workload multiple times to calibrate for 
the average time spent in profiler book-keeping code, which it uses during offline analysis to correct for CPU overhead caused by profiling.
This calibration only needs to be done once per workload and can be reused in future profiling runs.

\end{enumerate}

Our current implementation targets Python as the high-level language and supports both TensorFlow and PyTorch as ML backends since these are the most common tools used by RL developers. 
However, our design ensures that porting \rlscope to other high-level languages and ML backends is straightforward. 
\ifx\PaperFormatWithAppendix\undefined
\else
Additional implementation details such as avoiding sampling profilers and storing trace files asynchronously are provided in Appendix~\ref{sec:rlscope-impl}. 
\fi

\subsection{High-level Algorithmic Annotations}
\label{sec:high-level-algorithmic-annotations}

\begin{figure}[t]
\center{\includegraphics[width=\columnwidth]{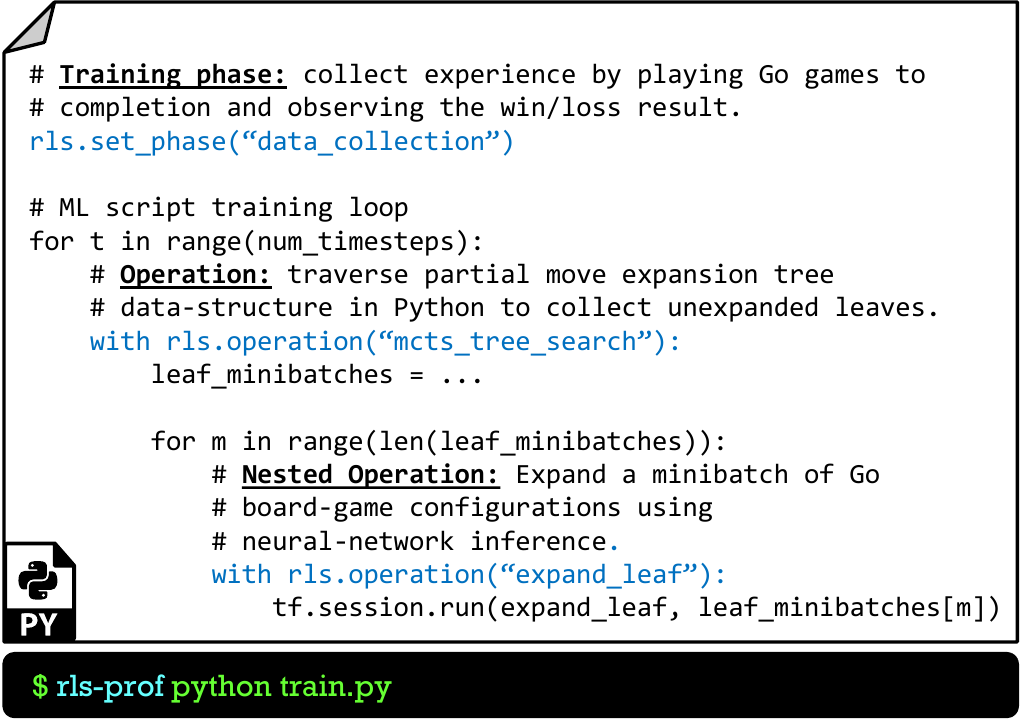}}
\caption{\textit{High-level algorithmic annotations:} \rlscope provides developers with a simple API for annotating their code with training phases and operations; these annotations are used to scope CPU/GPU critical path analysis to high-level algorithmic operations.}
\label{fig:02_ml_script}
\end{figure}

\rlscope exposes a simple high-level API to the developer for annotating their code with \textit{training phases} comprised of individual \textit{operations}.
We opt for manual annotations over automatic annotations derived from symbols since we have anecdotally observed that RL developers tend to become overwhelmed by results from language-level profilers (e.g., the Python profiler), but are quick to recognize high-level algorithmic operations.

Figure~\ref{fig:02_ml_script} shows part of a simplified Python script for training an agent to play the game Go~\cite{nature-agz}.  \rlscope leverages the Python \texttt{with} language construct to allow the developer to annotate regions of their code with high-level operation labels.  Executing a \texttt{with} block results in a start/end timestamps being recorded. 
Operations can be arbitrarily nested.  For example, \texttt{mcts\_tree\_search} uses pure high-level Python code to traverse a tree data-structure of possible board-game states randomly and collects minibatches of unexpanded Go board-game states.  Upon forming a minibatch, it executes \texttt{expand\_leaf}, which passes the minibatch of board-game states through a neural network to decide how to traverse the tree next.  This nesting of operations attributes all neural network inferences (in this example, CPU/GPU TensorFlow time) to \texttt{expand\_leaf}, and all data-structure traversal (Python CPU time) to \texttt{mcts\_tree\_search}.

\subsection{Transparent Event Interception}
\label{sec:transparent-event-interception}

To satisfy \rlscope's goals of collecting cross-stack information and avoiding simulator and ML library recompilation, \rlscope uses transparent hooks to intercept and record the start/end timestamps of profiling events.  \rlscope uses two techniques to collect events across the stack transparently: (i)~NVIDIA CUPTI profiling library and (ii)~High-level language $\leftrightarrow$ C interception:

\paragraph{NVIDIA CUPTI Profiling Library}  We use the NVIDIA CUPTI profiling library~\cite{cupti} to register hooks at program startup transparently.  
When the user launches \rlscope (bottom of Figure~\ref{fig:02_ml_script}), 
we transparently prepend \librlscope to the \codeword{LD_PRELOAD} environment variable to force hooks to be registered before loading any ML libraries.  
Using these hooks, we can collect \textit{CUDA API time}, CPU time spent in CUDA API calls such as \texttt{cudaLaunchKernel} that asynchronously queue kernels for execution on the GPU, and \textit{GPU kernel time}: time spent executing CUDA kernel code on the GPU.

\paragraph{High-level language $\leftrightarrow$ C interception:} To collect time spent in high-level language (e.g., Python)
and C libraries (e.g. ML backends and simulators) we  intercept calls to (high-level $\rightarrow$ C) and returns from C libraries (C $\rightarrow$ high-level).
To perform this interception while avoiding recompilation, we use Python to dynamically generate function wrappers around native library bindings for both  simulator and ML libraries.
Using timestamps at these interception points, we can collect time spent in Python, ML backends, and simulators.

\subsection{Cross-Stack Event Overlap}
\label{sec:event-overlap}
\begin{figure}[t]
\center{\includegraphics[width=0.95\columnwidth]{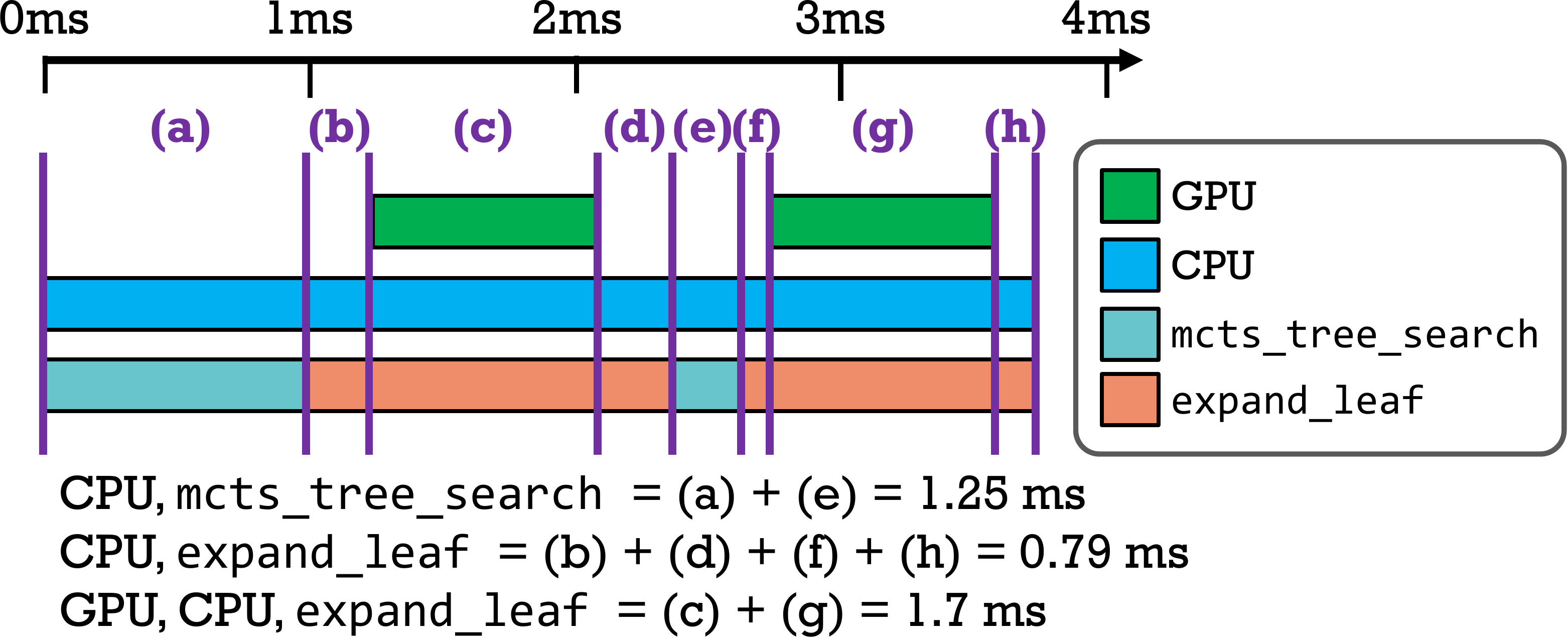}}
\caption{\textit{Cross-stack event overlap:} \rlscope accounts for the overlap between cross-stack events, which simultaneously measures overlap between CPU and GPU resources, and scopes training time to high-level algorithmic operations.
}
\label{fig:03_overlap_regions}
\end{figure}

Unfortunately, visualizations showing raw event traces
tend to overwhelm RL users since they do not provide a clear summary of the percentage of CPU and GPU time on the critical path of a program scoped to high-level algorithmic operations. \rlscope solves this issue by using the raw event trace to compute CPU/GPU overlap scoped to high-level user operations.  Figure~\ref{fig:03_overlap_regions} illustrates how CPU/GPU usage is computed. Offline, \rlscope linearly walks left-to-right through the events in the trace sorted by start-time and identifies regions of overlap between CPU events (e.g., Python, simulator, and TensorFlow C), GPU events (i.e., kernel execution times, memory copies), and high-level operations (e.g., \codeword{mcts_tree_search} and \codeword{expand_leaf}). Resource utilization in these overlapping regions is then summed to obtain the total CPU/GPU critical path latency, scoped to the user's operations. For example, 
\codeword{expand_leaf} spends $0.79 ms$ of its time purely CPU-bound, while $1.7 ms$ is spent executing on both the CPU and the GPU.
For more fine-grained CPU information, CPU time can be further divided into ML backend and CUDA API time.

\ifx\PaperFormatWithAppendix\undefined

\input{tex/no_appendix_profiling_calib_and_overhead_correction}
\else

\input{tex/with_appendix_profiling_calib_and_overhead_correction}
\fi

%% file: tex/no_appendix_profiling_calib_and_overhead_correction.tex
\subsection{Profiling Calibration and Overhead Correction}
\label{sec:profiling-calibration-and-overhead-correction}
Profilers inflate CPU-side time due to additional book-keeping code in the critical path.
This overhead can reach up to  \asPercent{\HRDCalibValidPercentInflationMax} of runtime in our experiments. 
\rlscope achieves accurate cross-stack critical path measurements by correcting for any CPU time inflation during offline analysis.
To correct profiling overhead, \rlscope calibrates for the average duration of book-keeping code paths and subtracts this time at the precise point when it occurs.  Since \rlscope already knows when book-keeping occurs from the information it collects for critical path analysis (Sections~\ref{sec:high-level-algorithmic-annotations}~and~\ref{sec:transparent-event-interception}), the challenge lies in accurately calibrating for the average duration of book-keeping code.

We identify two types of profiler overhead that \rlscope must calibrate. 
First, the overhead of most \rlscope book-keeping code only depends on the type of intercepted event.
For example, the extra CPU time incurred by \rlscope intercepting Python $\leftrightarrow$ C/C++ transitions is the same regardless of which part of the code it occurs in.
Similarly, the overhead of our interception of CUDA API calls does not depend on which CUDA API was used, and tracking algorithmic annotations does not depend on which operation was annotated.
For these cases, we find \rlscope's overhead by dividing the increase in total runtime when enabling profiling by the number of times the book-keeping code was called.
The second type of overhead is incurred by internal closed-source profiling code inside the CUDA library.
This code inflates runtime by different amounts, depending on which CUDA API is called.
Since profiling cannot be enabled separately for different APIs, accounting for it requires tracking the number and duration of each individual API call separately.

\input{tex/overhead_eval}

%% file: tex/overhead_eval.tex
\subsection{Validating Accuracy of Overhead Correction}
\label{sec:validate-overhead-correction}

\begin{figure}[t]
\captionsetup[subfigure]{position=b}
\centering
\subcaptionbox{\textit{Algorithm choice.} The same Walker2D simulator is used for different choices of RL algorithm.
\label{fig:13_overhead_correction_algorithm_choice}}
{\centering
\includegraphics[width=.30\textwidth]{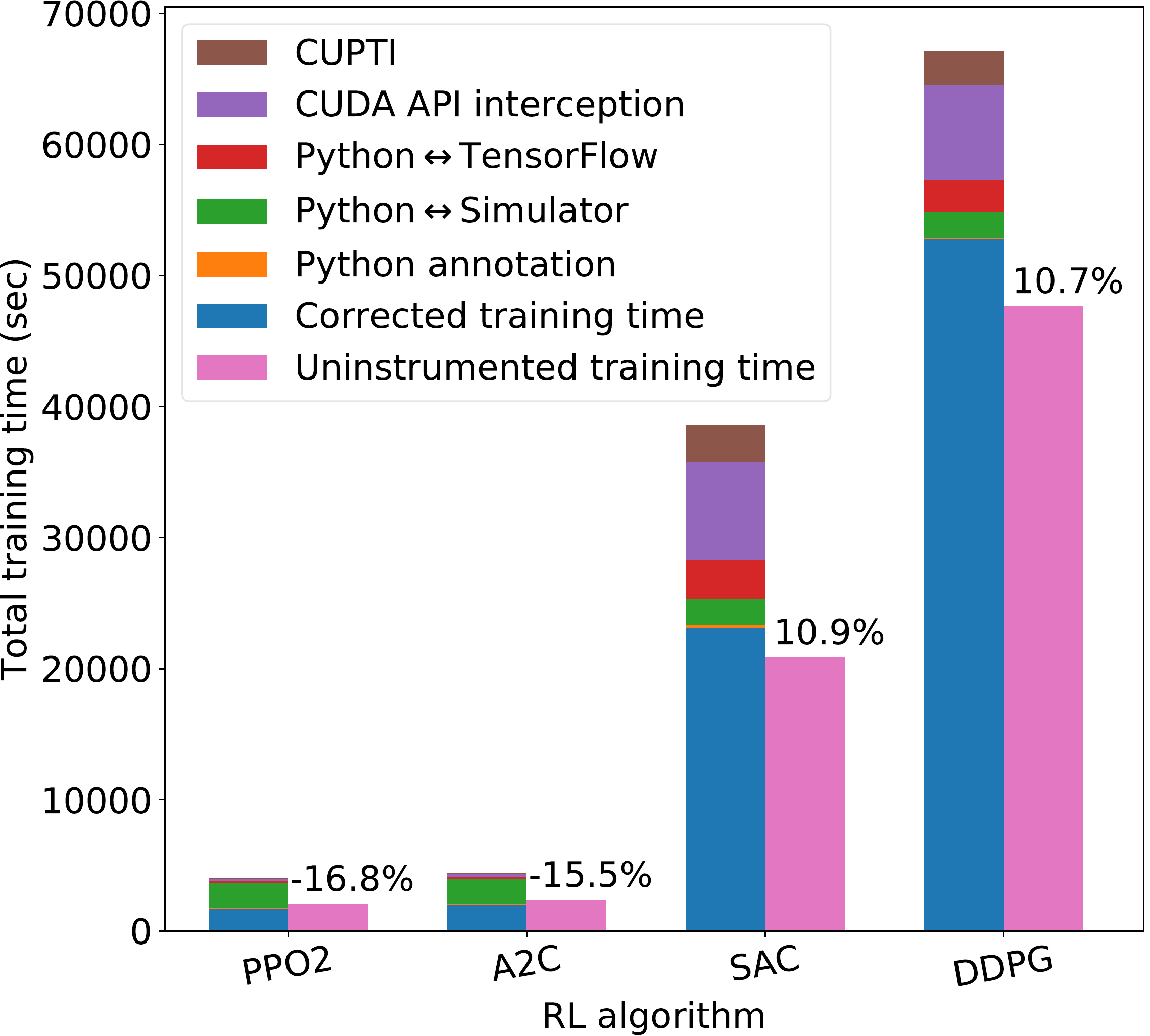}}
\hspace{1em}
\subcaptionbox{\textit{Simulator choice.} The same PPO2 RL algorithm is used for different choices of simulators. 
\label{fig:13_overhead_correction_environment_choice}}
{\centering
\includegraphics[width=.30\textwidth]{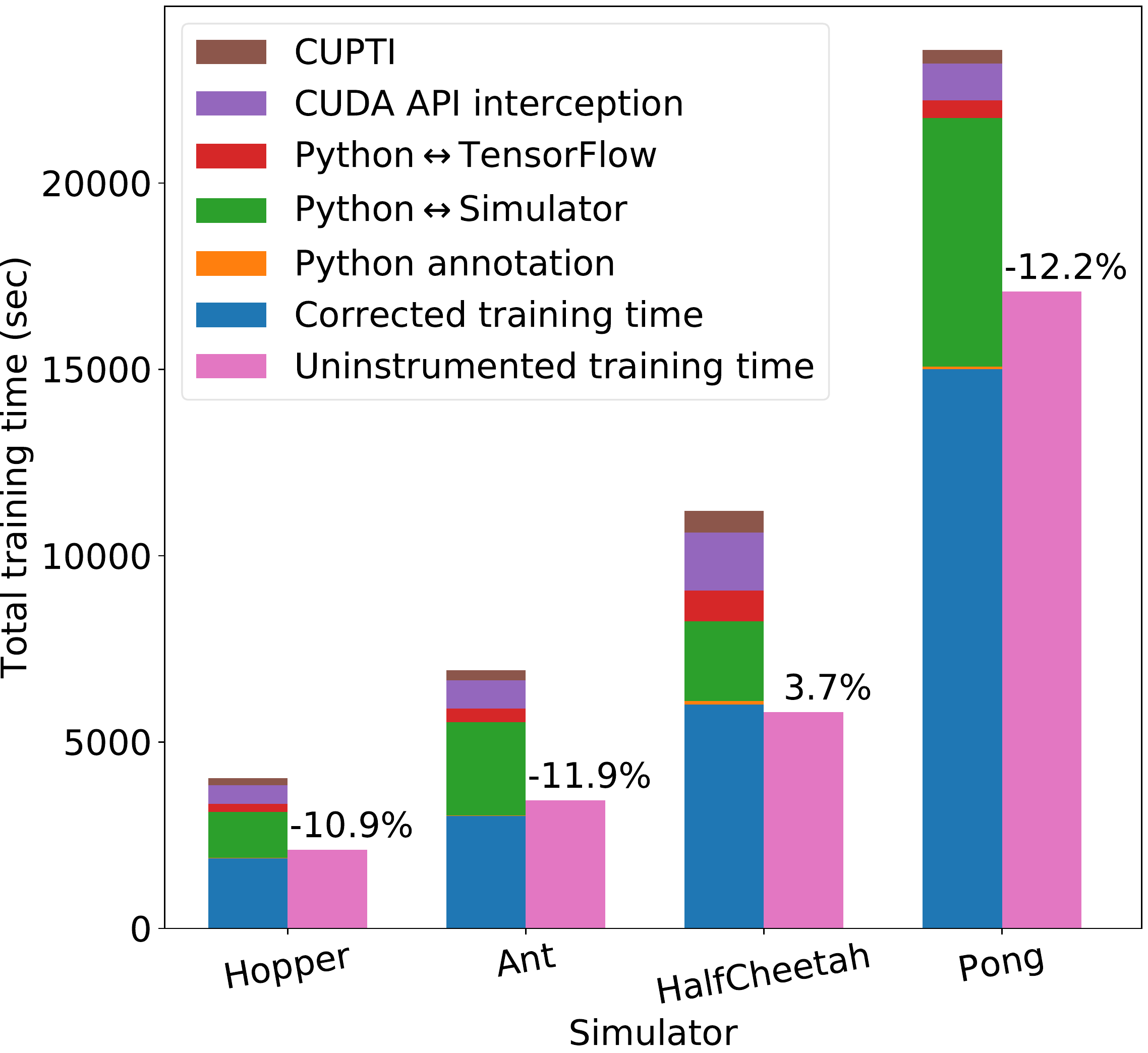}}
\caption{\textit{Low-bias overhead correction:} The left stack of bars shows \textit{Full \rlscope}, and the right bar shows an \textit{Uninstrumented} run.  The bias (\%) shows the deviation of the \textit{Corrected training time} (\barbox{blue}{blue}, left) from the \textit{Uninstrumented time} (\barbox{magenta}{pink}, right). \rlscope is able to achieve low-bias correction of critical path latency, regardless of simulator or algorithm.
}
\label{fig:13_overhead_correction}
\end{figure}

To evaluate the accuracy of overhead correction, we run each workload twice: once without \rlscope (uninstrumented run), and once with full \rlscope with all book-keeping code enabled.  We compare the \textit{corrected training time} reported by \rlscope to the \textit{uninstrumented training time}.  
If our overhead correction has low bias, then the corrected training time will be close to the uninstrumented training time.
Figure~\ref{fig:13_overhead_correction} shows corrected (\barbox{blue}{blue}) and uninstrumented (\barbox{magenta}{pink}) run times for different choices of environment (e.g., Walker2D, Ant, Pong) and RL algorithm (e.g., PPO2, A2C, DDPG).
Regardless of the choice of RL algorithm and or simulator, \rlscope is able to achieve accurate overhead correction of within $\pm 16\%$ of the actual uninstrumented training time, down from up to \FCAsPercent{\HRDCalibValidPercentInflationMax} without overhead correction.

\ifx\RebuttalEffectOfOverheadCorrection\undefined %
\else
\subsection{Effect of Overhead Correction}
\label{sec:effect-overhead-correction}

To understand the impact of overhead correction on the case studies performed throughout this paper, we re-computed the RL framework case study (Section~\ref{sec:framework_choice}) without overhead correction, and observed how results of our analysis changed. 
First, we observed bottleneck shift with TensorFlow Eager DDPG (Figure~\ref{fig:framework_choice_ddpg}), where the largest contributor to CPU-bound ML backend time shifts from Inference \textit{with} correction to Backpropagation \textit{without} correction, which is due to overhead from a larger number of ML backend transitions in Backpropagation compared to Inference (Figure~\ref{fig:framework_choice_ddpg_transitions}). 
Second, findings related to GPU usage (Section~\ref{sec:gpu-usage}) become incorrect since the total GPU-bound time becomes dwarfed by inflation in CPU time.  
For example, 
time spent in CUDA API calls exceeds GPU kernels by \FCAsTimes{\UncorrectedFindSurpCudaApiDominatesMeanRatioCUDAToGPU} (up from \FCAsTimes{\FindSurpCudaApiDominatesMeanRatioCUDAToGPU}).
Third, the total training time of the workloads we measured become inflated  \FCAsTimes{\FindUncorrectedTrainingTimeInflationMinRatio} - \FCAsTimes{\FindUncorrectedTrainingTimeInflationMaxRatio}, and \FCAsTimes{\FindUncorrectedTrainingTimeInflationMeanRatio} on average.
Given that \rlscope is a profiling tool for heterogeneous CPU/GPU RL workloads, it is important that \rlscope provides accurate CPU-bound and GPU-bound metrics, otherwise developers will spend their time optimizing artificial bottlenecks.

\fi %

%% file: tex/with_appendix_profiling_calib_and_overhead_correction.tex
\subsection{Profiling Calibration and Overhead Correction}
\label{sec:profiling-calibration-and-overhead-correction}
Profilers inflate CPU-side time due to additional book-keeping code in the critical path.
This overhead can reach up to  \asPercent{\HRDCalibValidPercentInflationMax} of runtime in our experiments (Appendix~\ref{sec:validate-overhead-correction}). %
\rlscope achieves accurate cross-stack critical path measurements by correcting for any CPU time inflation during offline analysis.
The full details of how we calibrate and verify the overhead correction are available in Appendix~\ref{sec:app-profiler-overhead-correction}, and below, we provide a high-level summary.

To correct profiling overhead, \rlscope calibrates for the average duration of book-keeping code paths and subtracts this time at the precise point when it occurs.  Since \rlscope already knows when book-keeping occurs from the information it collects for critical path analysis (Sections~\ref{sec:high-level-algorithmic-annotations}~and~\ref{sec:transparent-event-interception}), the challenge lies in accurately calibrating for the average duration of book-keeping code.

We identify two types of profiler overhead that \rlscope must calibrate. 
First, the overhead of most \rlscope book-keeping code only depends on the type of intercepted event.
For example, the extra CPU time incurred by \rlscope intercepting Python $\leftrightarrow$ C/C++ transitions is the same regardless of which part of the code it occurs in.
Similarly, the overhead of our interception of CUDA API calls does not depend on which CUDA API was used, and tracking algorithmic annotations does not depend on which operation was annotated.
For these cases, we find \rlscope's overhead by dividing the increase in total runtime when enabling profiling by the number of times the book-keeping code was called; see Appendix~\ref{sec:isolated-book-keeping-event-calibration} for details.
The second type of overhead is incurred by internal closed-source profiling code inside the CUDA library.
This code inflates runtime by different amounts, depending on which CUDA API is called.
Since profiling cannot be enabled separately for different APIs, accounting for it requires tracking the number of and duration of each individual API call separately; see Appendix~\ref{sec:grouped-book-keeping-event-calibration} for details.

We validate overhead correction accuracy on a diverse set of RL training workloads with different combinations of environment and RL algorithms.
We run each workload without profiling, and again with full \rlscope with all book-keeping code enabled.
Across all algorithms and simulators, \rlscope's error is within $\pm 16\%$ of the actual uninstrumented training time (see Appendix~\ref{sec:validate-overhead-correction} for details).

%% file: tex/case_studies.tex
\section{\rlscope Case Studies}
\label{sec:visualization}

We illustrate the usefulness of \rlscope's accurate cross-stack profiling metrics through in-depth case studies.
The hardware configuration we used throughout all experiments is an AMD EPYC 7371 CPU running at 3.1 GHz with 128 GB of RAM and an NVIDIA 2080Ti GPU.
For software configuration, we used Ubuntu 18.04, TensorFlow v2.2.0, PyTorch v1.6.0, and Python 3.6.9.  The exact RL framework(s), simulator(s), RL algorithm(s), and ML backend(s) used varies by case study, where it is stated explicitly.

\subsection{Case Study: Selecting an RL Framework}
\label{sec:framework_choice}

One of the first major problems faced by RL developers is which RL framework to use since there are so many spanning both
TensorFlow and PyTorch ML backends~\cite{blog2-choose-rl}.
In the absence of apples-to-apples comparisons amongst available RL frameworks, users fall back on rule-of-thumb approaches~\cite{blog2-choose-rl,blog-choose-rl} to selecting an RL framework: 
PyTorch for ease-of-use, 
frameworks with a particular algorithm, %
frameworks built on a specific ML backend.
This suggests developers often do not understand the exact training time trade-off between using PyTorch- and TensorFlow-based RL frameworks, and how this trade-off is influenced by the underlying \emph{execution model} of the ML backend.

TensorFlow 1.0 initially only supported a \graph execution model, where symbolic computations are declared during program initialization, then run all at once using initial graph inputs.
PyTorch, and more recently TensorFlow 2.0, supports an \eager execution model where operators are run as they are defined, making it easy to use with language level debuggers.  
With TensorFlow 2.0, developers can use the \autograph execution model to automatically convert their \eager program to an optimized TensorFlow graph, so long as developers adhere to coding practices~\cite{autograph-python-coding}.
While developers generally prefer to only use the \eager execution API for productivity, \rlscope reveals that the execution model used for a given ML backend can subtly impact training performance.

\begin{table}[t]
\caption{\textit{RL frameworks:} we consider popular $\langle$\textit{execution model, ML backend}$\rangle$ combinations.
}
\begin{tabular}{@{}llll@{}}
\toprule
\textbf{RL framework}&  \textbf{Execution model} & \textbf{ML backend} \\ \midrule
stable-baselines & \graph & TensorFlow 2.2.0 \\ \midrule
tf-agents & \autograph & TensorFlow 2.2.0 \\ \midrule
tf-agents & \eager & TensorFlow 2.2.0 \\ \midrule
ReAgent & \eager & PyTorch 1.6.0 \\ \bottomrule
\end{tabular}
\label{table:frameworks}
\end{table}

To ensure an apples-to-apples comparison across RL frameworks, we compare the same underlying RL algorithms (TD3/DDPG), the same simulator (Walker2D), and identical hyperparameters (e.g., batch size, network architecture, learning rate, %
etc.) which were pre-tuned in the stable-baselines RL framework~\cite{stable-baselines,rl-zoo}.  This allows us to attribute differences in training time breakdowns to an RL framework's ML backend and execution model.  We consider four representative RL frameworks that span popular 
$\langle$\textit{execution model, ML backend}$\rangle$ combinations\footnote{ReAgent does not yet make use of PyTorch's \autograph analog TorchScript for training, only for model serving.}, shown in Table~\ref{table:frameworks}.

\begin{figure*}[t]
    \centering
    \begin{subfigure}[b]{0.475\textwidth}
        \centering
        \includegraphics[width=\textwidth]{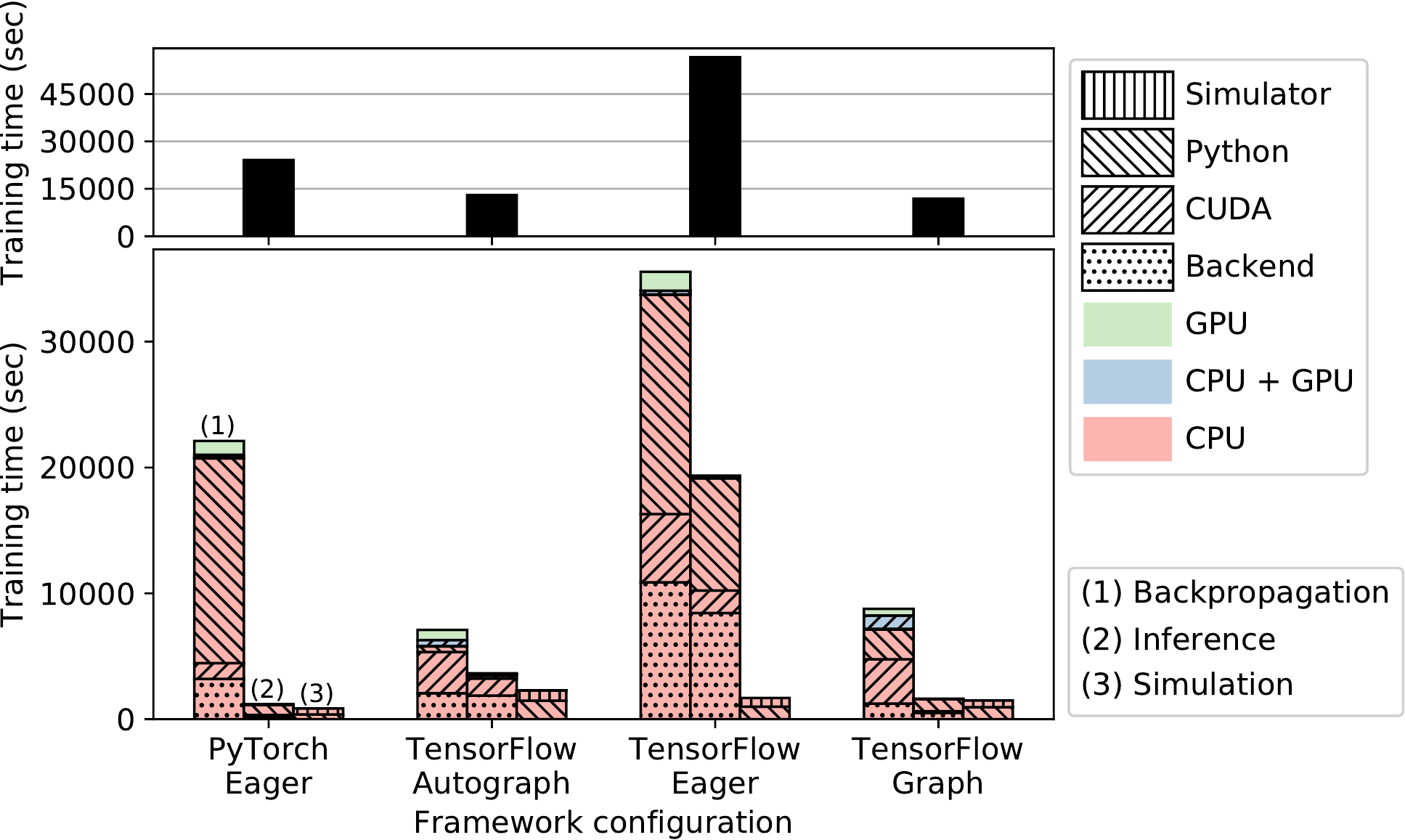}
        \caption{{\small (TD3, Walker2D) - time breakdown}}    
        \label{fig:framework_choice}
    \end{subfigure}
    \hfill
    \begin{subfigure}[b]{0.475\textwidth}  
        \centering 
        \includegraphics[width=\textwidth]{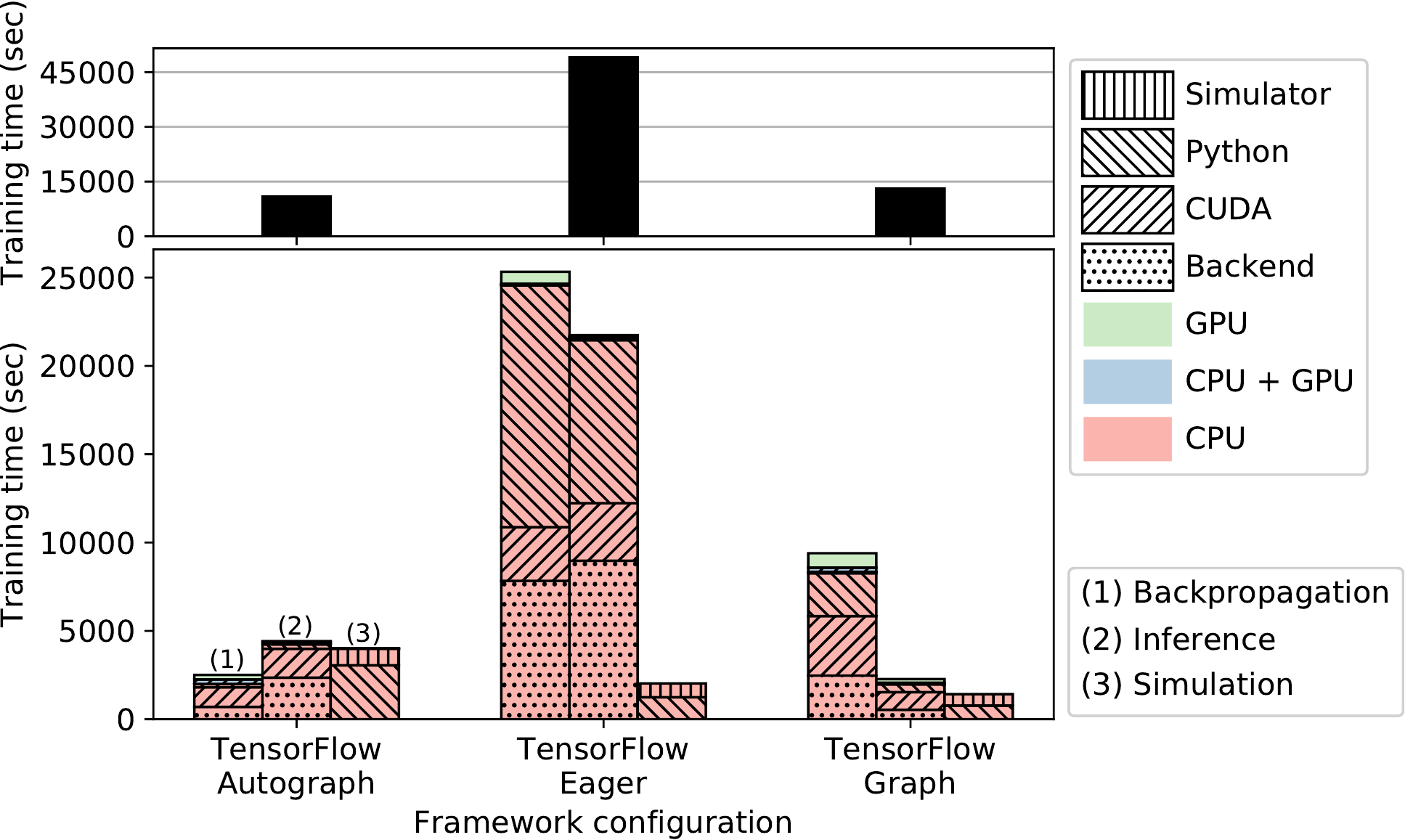}
        \caption{{\small (DDPG, Walker2D) - time breakdown}}    
        \label{fig:framework_choice_ddpg}
    \end{subfigure}
    \vskip\baselineskip
    \begin{subfigure}[b]{0.475\textwidth}   
        \centering 
        \includegraphics[width=\textwidth]{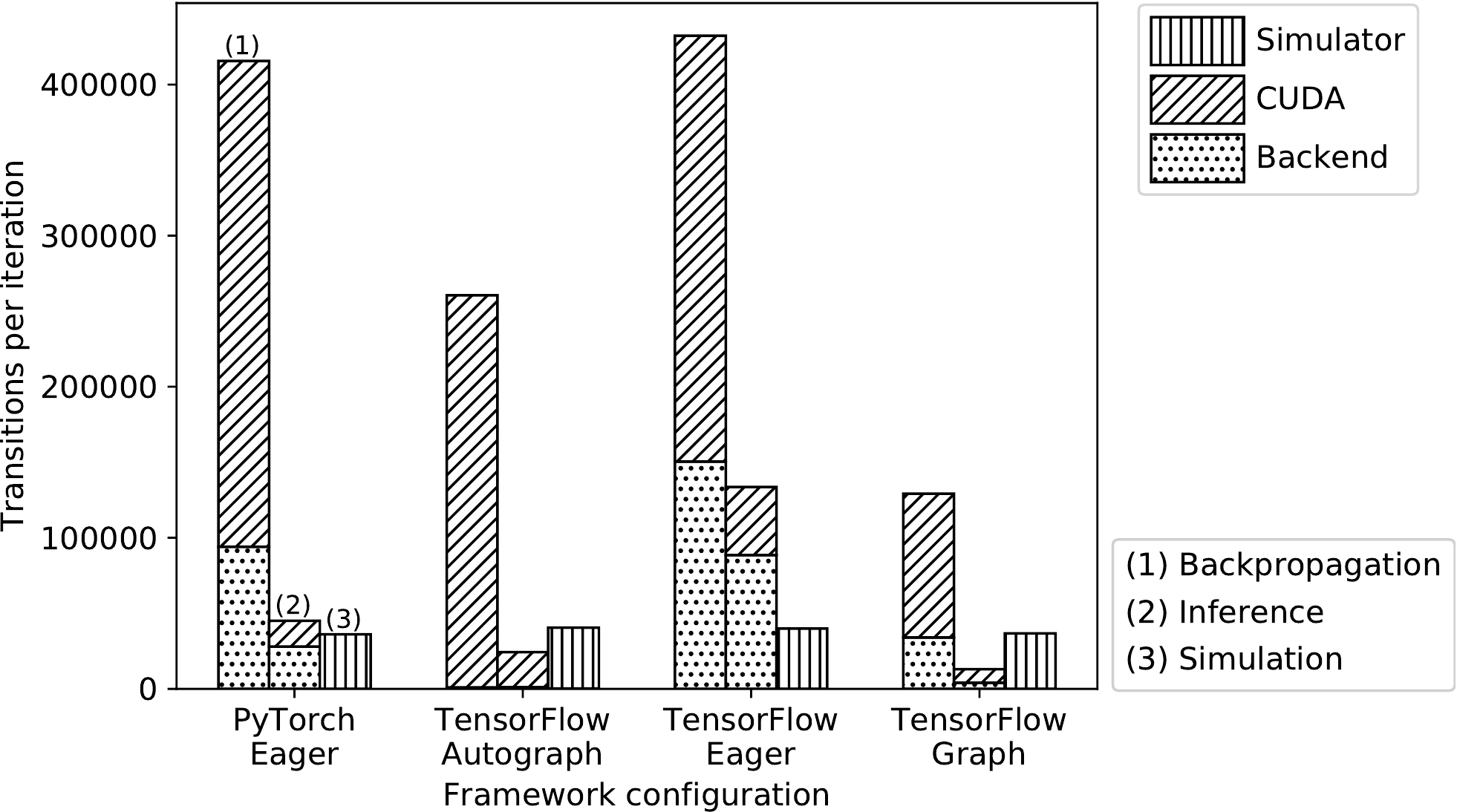}
        \caption{{\small (TD3, Walker2D) - language transitions}}    
        \label{fig:framework_choice_transitions}
    \end{subfigure}
    \hfill
    \begin{subfigure}[b]{0.475\textwidth}   
        \centering 
        \includegraphics[width=\textwidth]{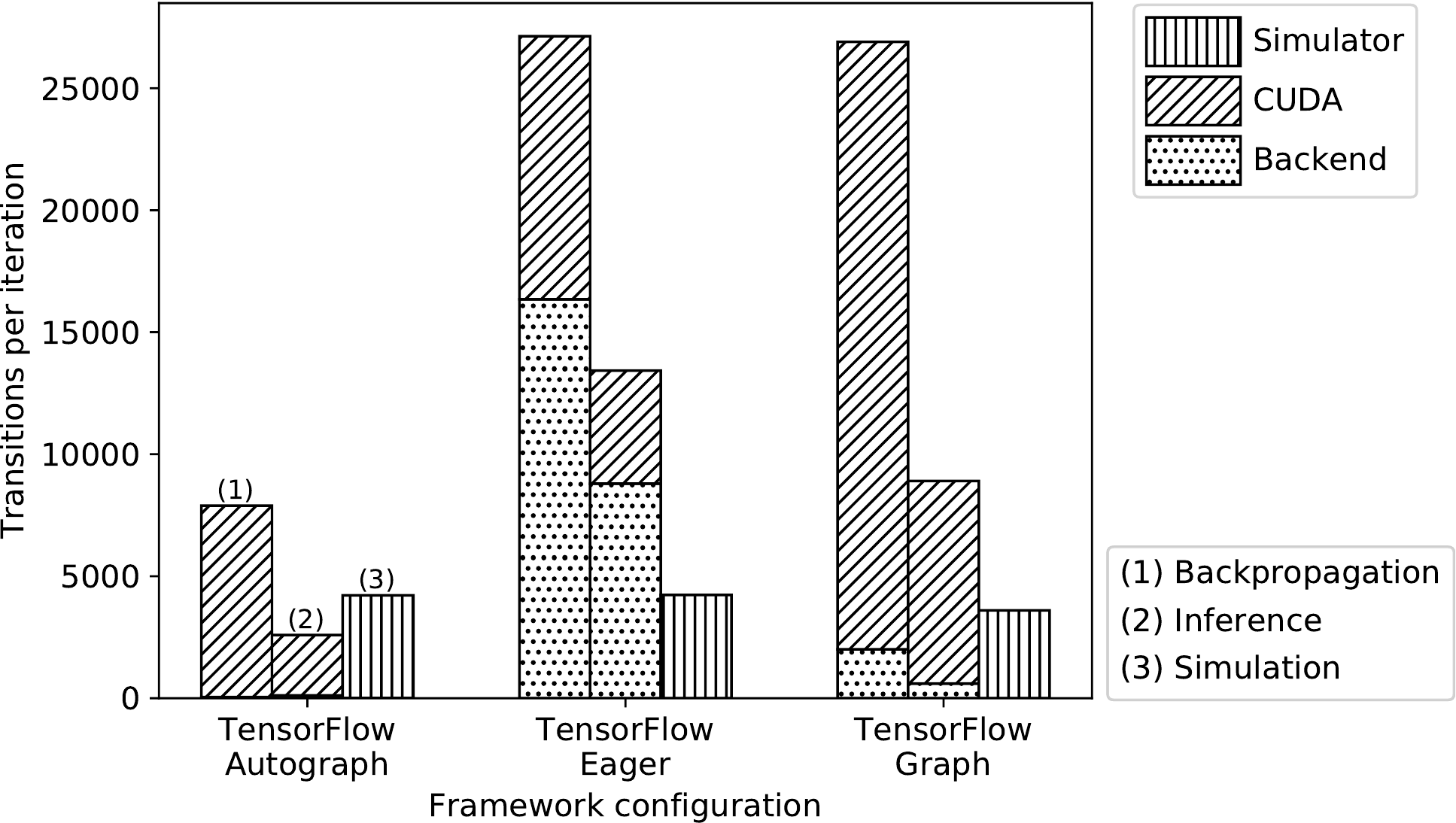}
        \caption{{\small (DDPG, Walker2D) - language transitions}}    
        \label{fig:framework_choice_ddpg_transitions}
    \end{subfigure}
    \caption{\small \textit{RL framework comparison:} We used identical RL algorithm (left: TD3, right: DDPG), simulator (Walker2D), and tuned hyperparameters; differences in execution between RL frameworks are strictly due to RL algorithm implementation and ML backend.  Differences in \textit{time breakdown} (top) across RL frameworks can be explained by higher number of \textit{language transitions} (bottom) between the Python and the ML backend (\textit{Backend}), and between the ML backend and the accelerator API calls (\textit{CUDA}).} 
    \label{fig:all_framework_choice}
\end{figure*}

Figure~\ref{fig:framework_choice} shows the training time breakdown across the RL frameworks for TD3.
The top black bars show the total training time for each workload.
The bottom bars show the same training time split between the different operations of the RL training loop (i.e., backpropagation, inference, simulation).
Colors indicate whether the time was spent in \BoxGpu, \BoxCpu, or overlapping both \BoxCpuPlusGpu,
while patterns show the software stack level (\emph{Simulator}, \emph{Python}, ML \emph{Backend}, or \emph{CUDA} API calls). 
Figure~\ref{fig:framework_choice_transitions} below shows the number of native library transitions from Python$\rightarrow$ML \emph{Backend},  Python$\rightarrow$\emph{Simulator}, and ML Backend$\rightarrow$\textit{CUDA} API calls.  
As we will see, these transitions are a major contributor to differences in execution time breakdowns. 
Similarly, Figures~\ref{fig:framework_choice_ddpg} and~\ref{fig:framework_choice_ddpg_transitions} shows the training time breakdown and transitions for DDPG.
We will frequently refer to these figures to support our findings.

\subsubsection{Execution model}
\label{sec:exec-model}

\rlscope's metrics are able to quantify our intuition of how execution models affect training time by correlating it with reduced Python$\rightarrow$Backend transitions.
This correlation further explains a \FCAsTimes{\FindQualPytorchEagerBetterMinPyTorchEagerSpeedup} slowdown between the same algorithm implemented in two different ML backends.

\begin{rlscope-finding}{find:qual-eager-more-trans}
\eager execution is \FCAsTimes{\FindQualEagerMoreTransMinEagerSlowdown} to \FCAsTimes{\FindQualEagerMoreTransMaxEagerSlowdown} slower than both \autograph and \graph execution, whereas neither \autograph nor \graph always outperform the other and are within  \FCAsPercent{\FindQualAutographGraphSimilarMaxSpeedup} 
of one another.
\end{rlscope-finding}

The total training time of Figure~\ref{fig:framework_choice} shows that \eager execution implementations generally perform worse than \graph and \autograph.  TensorFlow's older \graph API is still a more efficient API than the newer \eager API.  Hence, developers should be careful to ensure their code-base can run with \autograph by adhering to coding practices~\cite{autograph-python-coding} and not rely on \eager performance alone.
Moreover, even \autograph does not always outperform the older \graph API.  
In particular,
for TD3 (Figure~\ref{fig:framework_choice}) \graph is \FCAsPercent{\FindQualAutographGraphSimilarMaxGraphSpeedup} faster than \autograph;
conversely,
for DDPG (Figure~\ref{fig:framework_choice_ddpg}) \autograph is \FCAsPercent{\FindQualAutographGraphSimilarMaxAutographSpeedup} faster than \graph.

\begin{rlscope-finding}{find:qual-autograph-reduces-python}
By reducing Python$\rightarrow$Backend transitions, \autograph substantially reduces Python time from \FCAsPercent{\FindQualAutographReducesPythonMaxGraphPythonPercentOfOp} in Graph to at most \FCAsPercent{\FindQualAutographReducesPythonMaxAutographPythonPercentOfOp} for inference/backpropagation.
\end{rlscope-finding}

Comparing TensorFlow \autograph to TensorFlow \graph for TD3 in Figure~\ref{fig:framework_choice} demonstrates substantial reductions in Python time both for  backpropagation (\FCAsTimes{\FindQualAutographReducesPythonMeanRatioPythonBackpropagationTD}) and for inference (\FCAsTimes{\FindQualAutographReducesPythonMeanRatioPythonInferenceTD}); similar results are observed for backpropagation (\FCAsTimes{\FindQualAutographReducesPythonMeanRatioPythonBackpropagationDDPG}) and inference (\FCAsTimes{\FindQualAutographReducesPythonMeanRatioPythonInferenceDDPG}) for DDPG in Figure~\ref{fig:framework_choice_ddpg}.   
These reductions in Python time are consistent with \autograph's design, which automatically converts high-level Python code (e.g., conditionals, while loops) into in-graph TensorFlow operators.  Figure~\ref{fig:framework_choice_transitions}/\ref{fig:framework_choice_ddpg_transitions} explains this reduction in Python time for TD3/DDPG since the number of Backend  transitions are close to zero in \autograph compared to other execution models.

\begin{rlscope-finding}{find:qual-pytorch-eager-better}
For TD3, PyTorch \eager is \FCAsTimes{\FindQualPytorchEagerBetterMinPyTorchEagerSpeedup} faster than TensorFlow \eager since it minimizes Python$\rightarrow$Backend transitions more effectively.  The \graph/\autograph implementation is \FCAsTimes{\FindSurpExecModelComparisonMaxTFSpeedup} faster than PyTorch \eager.
\end{rlscope-finding}
For TD3 (Figure~\ref{fig:framework_choice}), the difference in the \eager execution model's performance is markedly different across PyTorch and TensorFlow ML backends. 
PyTorch \eager performs up to 
\FCAsTimes{\FindSurpExecModelComparisonMaxTFSpeedup} 
slower than \graph/\autograph, whereas TensorFlow \eager is up to 
\FCAsTimes{\FindQualEagerMoreTransMaxEagerSlowdown} 
slower than \graph/\autograph.  This difference in \eager performance can be explained by Figure~\ref{fig:framework_choice_transitions}, where we observe a larger number of transitions to the Backend from Python in TensorFlow \eager than in PyTorch \eager during backpropagation (\FCAsTimes{\FindQualEagerMoreTransMeanRatioBackpropagationFrameworkTransTFToPyTorch}) and inference (\FCAsTimes{\FindQualEagerMoreTransMeanRatioInferenceFrameworkTransTFToPyTorch}).

\subsubsection{Bottleneck analysis}
\label{sec:bottleneck-analysis}

\rlscope's fine-grained time breakdown is useful for explaining subtle performance bottlenecks that are a result of an overly abstracted RL algorithm implementation\EvalRefFnd{find:surp-ddpg-backprop-slow}, sensitivity to particular hyperparameter settings\EvalRefFnd{find:surp-autograph-inflates-python}, and even the result of anomalous behaviour in the underlying implementation of an ML backend\EvalRefFnd{find:surp-autograph-inflates-inference}.

\begin{rlscope-finding}{find:surp-ddpg-backprop-slow}
\rlscope's metrics identify inefficient abstractions in high-level code as 
responsible for a \FCAsTimes{\FindSurpDdpgBackpropSlowMeanRatioAutographToGraphBackpropagationDDPG} inflation in backpropagation in \graph compared to \autograph for DDPG.
\end{rlscope-finding}

Backpropagation in \autograph is \FCAsTimes{\FindSurpDdpgBackpropSlowMeanRatioAutographToGraphBackpropagationDDPG} faster than \graph for DDPG (Figure~\ref{fig:framework_choice_ddpg}), whereas backpropagation in \autograph is only \FCAsTimes{\FindSurpDdpgBackpropSlowMeanRatioAutographToGraphBackpropagationTD} faster than \graph for TD3 (Figure~\ref{fig:framework_choice}); these inefficiencies in backpropagation for DDPG \graph are correlated with high CUDA API time inflation (\FCAsTimes{\FindSurpDdpgBackpropSlowMeanRatioAutographToGraphCUDABackpropagationDDPG}), 
and high Python inflation (\FCAsTimes{\FindSurpDdpgBackpropSlowMeanRatioAutographToGraphPythonBackpropagationDDPG}) relative to DDPG \autograph.  The inflation in CUDA API time suggests a greater number of kernel executions and greater Python time suggests a larger number of calls to the ML backend.
Inspecting the DDPG and TD3 implementations reveals that these inefficiencies in DDPG backpropagation are caused by two factors.  
First, DDPG uses inefficient abstractions leading to more GPU kernel launches; DDPG uses an MPI-friendly, but GPU-unfriendly Python implementation of the Adam optimizer that copies GPU weights to the CPU then writes back results to the GPU even during single-node training.  
Second, DDPG performs frequent backend transitions leading to more Python time; several operations (e.g., copying network weights to a ``target'' network, applying gradients to actor and critic networks) execute in separate Backend calls that could be bundled into a single call. 

\begin{rlscope-finding}{find:surp-autograph-inflates-python}
\autograph can inflate Python time by as much as \FCAsTimes{\FindSurpAutographInflatesPythonMaxRatioSimulationPythonAutographToEager} during simulation; training time (not just model accuracy) is highly sensitive to small differences in hyperparameter choices.
\end{rlscope-finding}

\rlscope is able to identify a performance anomaly in the DDPG \autograph implementation in Figure~\ref{fig:framework_choice_ddpg}; interestingly, the Python time of simulation in \autograph is inflated \FCAsTimes{\FindSurpAutographInflatesPythonMaxRatioSimulationPythonAutographToEager} when compared to executing in \eager mode.  
Even more surprisingly, both DDPG and TD3 share the same data collection code, yet TD3 does not suffer from this inflation (Figure~\ref{fig:framework_choice}).  
Hence, the only difference that would explain the anomaly in DDPG is a difference in hyperparameter settings from TD3.  
Inspecting the algorithm implementations reveals that TD3 performs 1000 consecutive simulator steps before performing a gradient update, whereas DDPG performs only 100 consecutive simulator steps.  
Hence, TD3 is better able to amortize overheads associated will calling into \autograph's in-graph data collection loop.  
To confirm our hypothesis, we adjusted DDPG's hyperparameter to 1000 to match TD3, and the inflation dropped down to  \FCAsTimes{\DDPGHyperParamFindSurpAutographInflatesPythonMeanRatioSimulationPythonAutographToEagerDDPG}, similar to what we see for TD3 (\FCAsTimes{\FindSurpAutographInflatesPythonMeanRatioSimulationPythonAutographToEagerTD3}).

\begin{rlscope-finding}{find:surp-autograph-inflates-inference}
Inference time in \autograph is inflated due to a \FCAsTimes{\FindSurpAutographInflatesInferenceMeanRatioInferenceFrameworkAutographToGraph} inflation in Backend time compared to \graph. This inflation is \textit{not} due to extra Python$\rightarrow$Backend transitions, and is instead a performance anomaly within the ML backend itself.
\end{rlscope-finding}

\rlscope's scoping capabilities can identify performance anomalies in RL framework implementations within specific high-level algorithmic operations, at precisely the level of the stack they originate from.  In TD3 (Figure~\ref{fig:framework_choice}), there is a 
\FCAsTimes{\FindSurpAutographInflatesInferenceMeanRatioInferenceFrameworkTDAutographToGraph}
increase in inference Backend time in \autograph compared to \graph; 
similarly, in DDPG (Figure~\ref{fig:framework_choice_ddpg}), there is a 
\FCAsTimes{\FindSurpAutographInflatesInferenceMeanRatioInferenceFrameworkDDPGAutographToGraph}
increase.  However, this inflation is not due to increased Python$\rightarrow$Backend transitions; in fact, \autograph clearly minimizes Backend transitions (Figures~\ref{fig:framework_choice_transitions} and \ref{fig:framework_choice_ddpg_transitions}).  Hence, this consistent anomalous behaviour indicates a performance issue within the \autograph implementation of inference, which is not present in the \graph API implementation.

\subsubsection{GPU usage}
\label{sec:gpu-usage}

\rlscope confirms that GPU usage is poor across \emph{all} RL frameworks, showing this issue is not isolated to any one ML backend.  In all ML backends, poor GPU usage is rooted in greater CPU time spent in CUDA API calls than spent in GPU kernels.

\begin{rlscope-finding}{find:surp-total-gpu-time}
Total GPU time is similar across all RL frameworks regardless of ML backend, and consistently low across all RL frameworks, making up at most \FCAsPercent{\FindSurpTotalGpuTimeMaxGPUPercent} of total training time.
\end{rlscope-finding}

In all RL frameworks the total GPU execution time makes up only a tiny portion of the total training time of both TD3 (Figure~\ref{fig:framework_choice}) and DDPG (Figure~\ref{fig:framework_choice_ddpg}), which is immediately apparent by the small total GPU time (i.e., \BoxGpu\ + \BoxCpuPlusGpu).  This suggests that poor GPU usage is not an implementation quirk in any one ML backend.  Instead, it is a widespread problem affecting \emph{all} ML backends.

\begin{rlscope-finding}{find:surp-cuda-api-dominates}
In all RL frameworks, CPU-side CUDA API time dominates total GPU kernel execution time, taking up on average \FCAsTimes{\FindSurpCudaApiDominatesMeanRatioCUDAToGPU} as much time as GPU kernel execution.
\end{rlscope-finding}

The surprising insight from \rlscope that CUDA API time dominates GPU kernel execution time emphasizes the drastic differences between supervised learning workloads and RL workloads established in Section~\ref{sec:compare-sl-and-rl}: (1) \textit{Training data collection:} data must be collected at run-time by repeatedly alternating between selecting an action (inference) and taking an action (simulation) -- the training loop includes more than just backpropagation, (2) \textit{Smaller neural networks:} RL algorithms make use of smaller neural networks than state-of-art models from supervised learning tasks, leading to shorter GPU kernel execution times.

\input{tex/case_study_algorithm_survey}

\ifx\PaperFormatWithAppendix\undefined
\input{tex/case_study_simulator_survey}

\else
\fi

\ifx\PaperFormatWithAppendix\undefined
\input{tex/no_appendix_case_study_scale_up_workload}
\else
\input{tex/with_appendix_case_study_scale_up_workload}

\fi

%% file: tex/case_study_algorithm_survey.tex
\ifx\PaperFormatWithAppendix\undefined
\subsection{Case Study: RL Algorithm Survey}
\else
\subsection{Case Study: RL Algorithm and Simulator Survey}
\fi
\label{sec:algorithm_choice}

\begin{figure}[t]
\center{\includegraphics[width=\columnwidth]{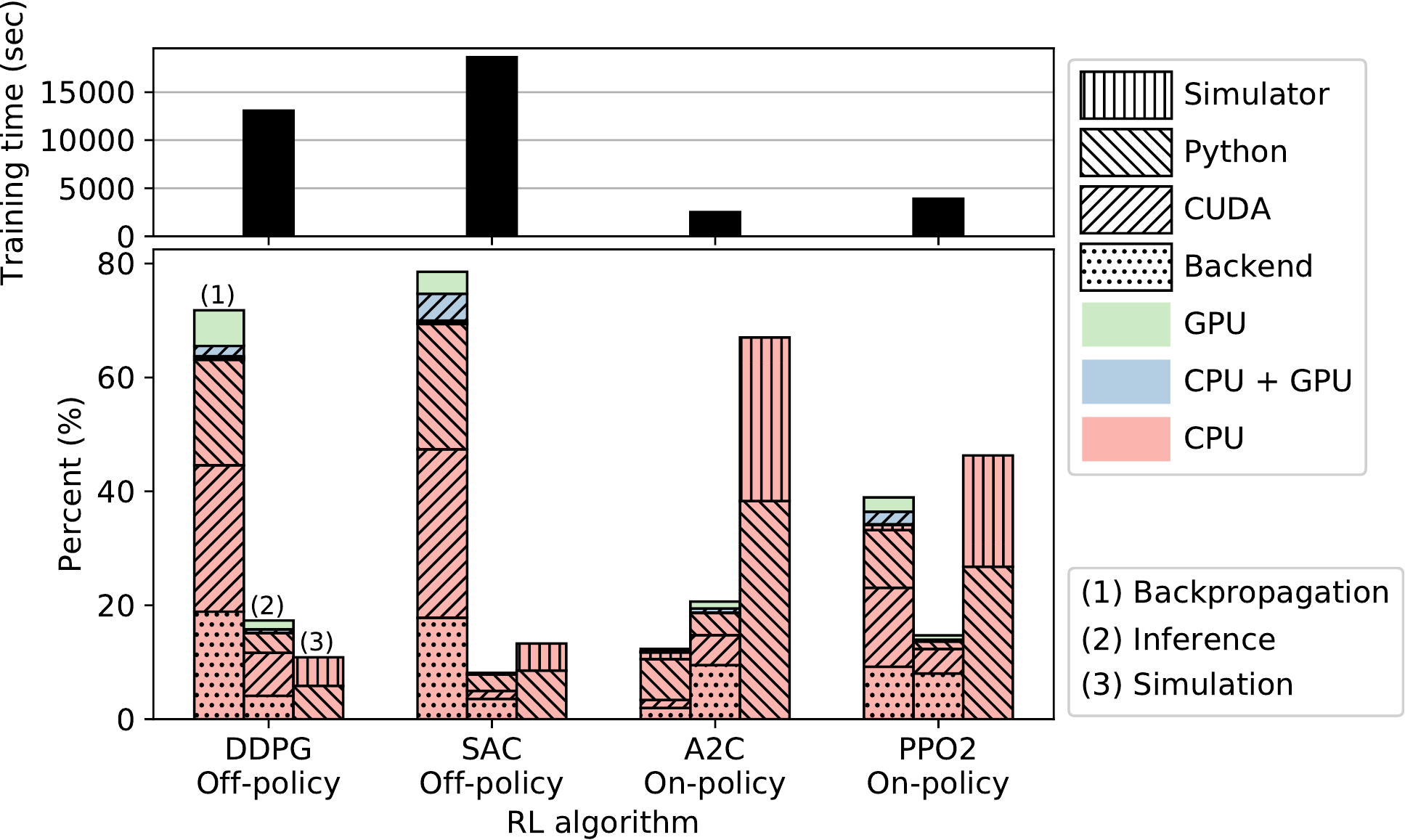}}
\caption{\textit{Algorithm choice:} We used a popular RL environment (Walker2D: robotics task of a walking humanoid) and measured how the stages (backpropagation, inference, simulation) of each measured algorithm change in relation to algorithm choice.
All tested RL workloads spend about 90\% of their runtime purely in the CPU.
}
\label{fig:algorithm_choice}
\end{figure}

Practitioners must evaluate a large number of RL algorithms for each application. For example, OpenAI's repository has implemented 9 RL algorithms~\cite{baselines}.
We investigate how algorithm choice affects the workload profile (Figure~\ref{fig:algorithm_choice}).  
We chose the popular Walker2D ``walking bipedal humanoid'' simulated task~\cite{erez2012infinite}, where the agent must learn to walk forward.

\begin{rlscope-finding}{find:software-overhead}
Most of the training time across RL algorithms is spent CPU-bound executing the software stack -- CUDA API calls, backend code, high-level code -- suggesting the entire hardware/software stack is poorly optimized for the RL use case.
Even inference and backpropagation, which are GPU-heavy in SL workloads, spend at most \FCAsPercent{\FindAlgoChoiceDDPGInferenceGpuResourcePercentMean} executing GPU kernels.
\end{rlscope-finding}

Of the surveyed RL algorithms, DDPG has the largest portion of time spent GPU-bound, with \FCAsPercent{\FindAlgoChoiceDDPGGpuResourcePercentMean} of total training time spent executing GPU kernels (i.e., \BoxGpu\ + \BoxCpuPlusGpu); conversely, A2C spends the least time on the GPU with only \FCAsPercent{\FindAlgoChoiceACGpuResourcePercentMean} total training time spent executing GPU kernels.  
If we consider backpropagation only, at most \FCAsPercent{\FindAlgoChoicePPOBackpropagationGpuResourcePercentMean} percent of a backpropagation operation is spent executing GPU kernels (for PPO2).  
Similarly, at most \FCAsPercent{\FindAlgoChoiceDDPGInferenceGpuResourcePercentMean} of an inference operation is spent executing GPU kernels (for DDPG). 
For the most GPU-heavy operation (backpropagation) of the most GPU-heavy algorithm (DDPG), 
\FCAsPercent{\FindAlgoChoiceDDPGSimulatorCategoryPercentMean} of CPU time is spent in simulation and \FCAsPercent{\FindAlgoChoiceDDPGPythonCategoryPercentMean} in Python while, with a non-negligible amount of CPU time spent in CUDA API calls (\FCAsPercent{\FindAlgoChoiceDDPGCUDACategoryPercentMean}) and the Backend (\FCAsPercent{\FindAlgoChoiceDDPGBackendCategoryPercentMean}); a similar spread of CPU time across the RL software stack is seen in all RL algorithms.

\begin{rlscope-finding}{find:algo-choice}
On-policy RL algorithms are at least \FCAsTimes{\FindAlgoChoiceMinRatioPercentOnPolicyToOffPolicySimulation} more simulation-bound than off-policy RL algorithms.
\end{rlscope-finding}

A2C and PPO2 spend most of their execution in simulation, with A2C spending \FCAsPercent{\FindAlgoChoiceACSimulationOpPercentMean} and PPO2 spending \FCAsPercent{\FindAlgoChoicePPOSimulationOpPercentMean}.  
Conversely, DDPG and SAC spend only \FCAsPercent{\FindAlgoChoiceDDPGSimulationOpPercentMean} and \FCAsPercent{\FindAlgoChoiceSACSimulationOpPercentMean} of their execution in simulation, respectively, which are instead dominated by backpropagation.
\textit{Off-policy algorithms} (e.g., DDPG, SAC) 
are said to be more ``sample-efficient'' since they are learning $Q(s,a)$ or $V(s)$ by sampling rewards obtained from the simulator; as a result, they can re-use data collected from the simulator when forming a minibatch to learn from, allowing them to spend a smaller percentage of their training time collecting data from the simulator.  \textit{On-policy algorithms} (e.g., A2C, PPO2) must learn and update a policy network $\pi$ directly.  As a result, the minibatches for updating the policy network must consist of experience collected using the same policy.  This means entire episodes of experience must be collected from the simulator before a minibatch, and subsequent gradient update can be performed; hence, the on-policy algorithm is sample-\textit{in}efficient.  Interestingly, this disparity has been qualitatively explained as the ``sample-efficiency'' problem in the RL research community~\cite{wang2016sample}, but it has never been systematically quantified.

\ifx\PaperFormatWithAppendix\undefined
\else
Due to space constraints, our survey of how simulators affect total RL training time can be found in Appendix~\ref{sec:environment_choice}.
\fi

%% file: tex/case_study_simulator_survey.tex
\subsection{Case Study: Simulator Survey}
\label{sec:environment_choice}

\ifx\PaperFormatWithAppendix\undefined
\else
\begin{figure}[t]
    \centering
    \includegraphics[width=0.95\columnwidth]{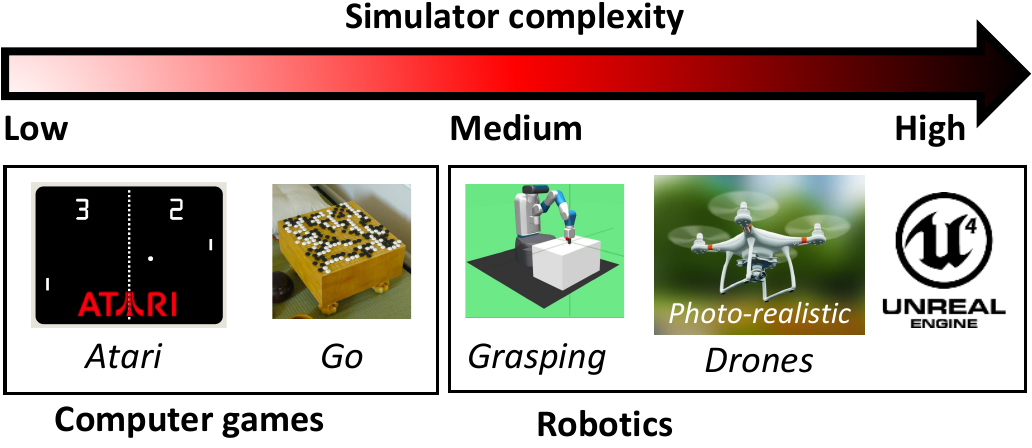}
    \caption{\textit{A representative sample of RL simulators} organized by computational complexity. Computer games such as arcade games have simple graphics and hence have lower complexity. Emerging robotics applications often have photo-realistic simulators and have higher complexity.
    }
    \label{fig:01_simulators}
\end{figure}
\fi

\begin{figure}[t]
\center{\includegraphics[width=\columnwidth]{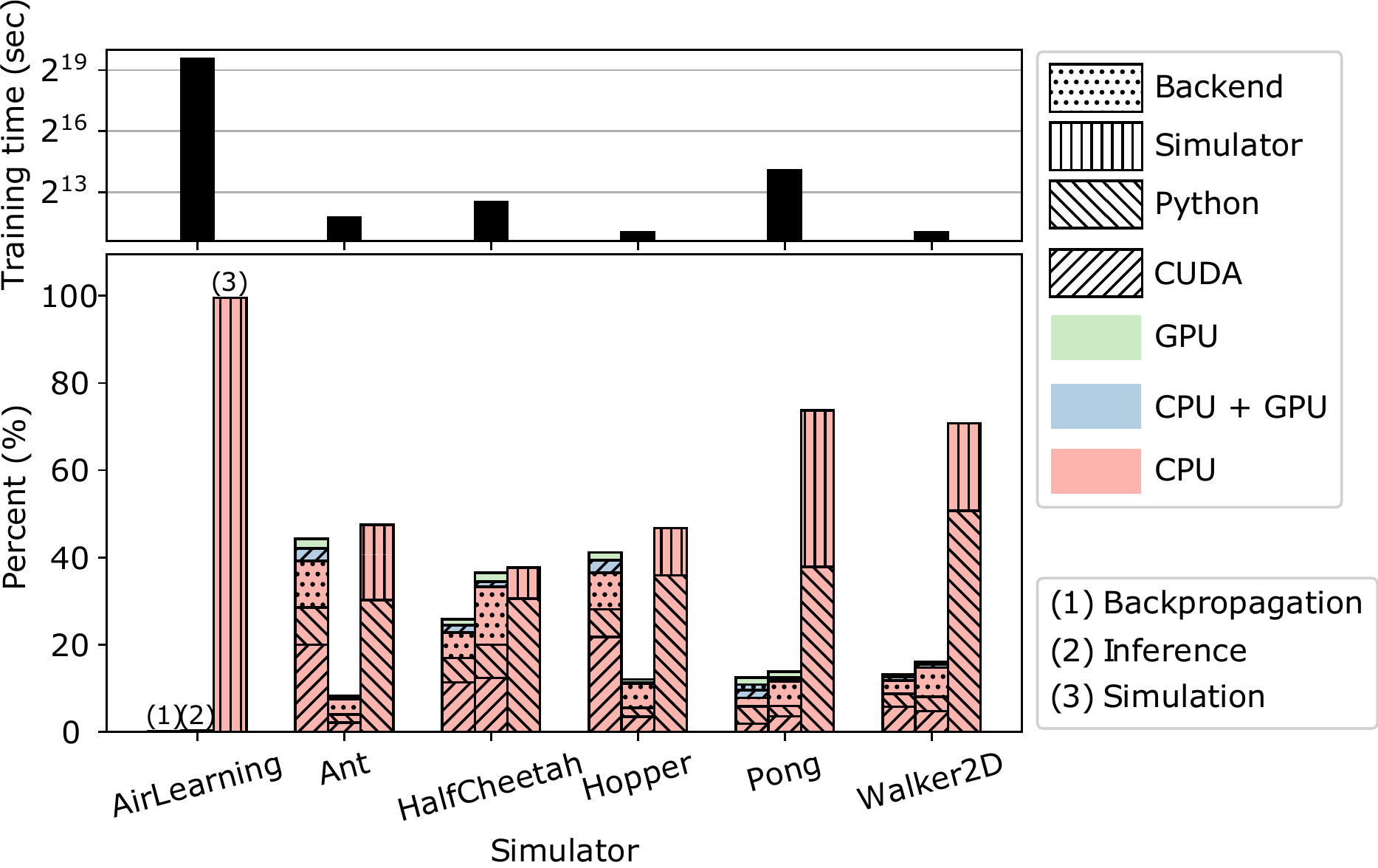}}
\caption{\textit{Simulator choice:} We used a top-performing RL algorithm (PPO) and measured how each stage of the algorithm changes with respect to environment choice.
GPU accounts for 5\% or less of the runtime across all simulators.
}
\label{fig:environment_choice}
\end{figure}

To observe how RL training time varies with the choice of simulator, we focus on three application domains, 
\ifx\PaperFormatWithAppendix\undefined
organized by simulator complexity, 
\else
organized by simulator complexity (Figure~\ref{fig:01_simulators}), 
\fi
described in detail below:

\paragraph{ML algorithm research:}  
These applications use simple, quick-to-execute simulators such as games (i.e. Atari Pong, Go board game), where the notion of reward is well-defined (e.g., 1 or 0 if a game is won or lost).  RL researchers often use such applications when exploring enhancements to core RL algorithms, since it is easier to understand the behaviour of the learned policy since rules of the game and common human strategies are known.  Atari games are provided by the OpenAI gym repository~\cite{brockman2016openai}.
    
\paragraph{Robotics:}
These applications use physics simulators to emulate a robot interacting with its environment, in tasks such as locomotion and object grasping.  The physics simulation is done on the CPU, whereas training is done on the GPU.  These medium complexity robotics tasks are provided by the OpenAI gym repository~\cite{brockman2016openai}.
    
\paragraph{Drone applications:}
These applications use realistic physics simulators with photo-realistic rendering implemented using a popular video game engine~\cite{ue4}.  Sufficiently realistic simulators allow a researcher to determine how an RL algorithm will behave in a real-world deployment without the high cost overhead associated with actually deploying it~\cite{dosovitskiy2017carla}. These simulators are computationally intensive, and make use the GPU to perform graphics rendering.  We use the point-to-point navigation task from the AirLearning toolkit for unmanned aerial vehicle (UAV) robotics tasks~\cite{krishnan2019air}.

We chose a top-performing (i.e., highest final average episodic reward) algorithm from the OpenAI stable-baselines repository~\cite{stable-baselines,rl-zoo} and ran it across a representative sampling of simulators ranging from low, medium, to high complexity; Figure~\ref{fig:environment_choice} provides a breakdown of training time for each different simulator.

\begin{rlscope-finding}{find:simulator-choice}
Simulation time varies, but is always a \textit{large} training bottleneck, taking up at least \asPercent{\HRDSimulatorComparisonAtLeastSimulationBoundPercent} of training time.
\end{rlscope-finding}

Excluding AirLearning, the CPU/GPU breakdown is similar across all low/medium complexity simulators, with as little as \asPercent{\HRDSimulatorComparisonWalkerDGpuPercent} GPU kernel time for Walker2D and as high as \asPercent{\HRDSimulatorComparisonPongGpuPercent} for Pong.  
For high-complexity simulators like AirLearning that simulate a drone in a realistic video game engine, simulation dominates total training time at \asPercent{\HRDSimulatorComparisonAirLearningSimulationPercent}.  In the presence of such large simulation times, GPU time spent in inference or backpropagation makes practically no contribution to total training time. 
Excluding high-complexity simulators (AirLearning), even in low and medium complexity environments, simulation is always the majority contributor to total training time compared to inference and backpropagation, accounting for anywhere between 
\asPercent{\HRDSimulatorComparisonHalfCheetahSimulationPercent} in HalfCheetah, to \asPercent{\HRDSimulatorComparisonPongSimulationPercent} in the Pong simulator.

Even though Pong is a low-complexity simulator, the tuned hyperparameter configuration we used for (PPO, Pong) performs a small number of gradient updates compared to the number of simulator invocations, which explains why it has a large simulation time; this is also the case for the Walker2D configuration.  Further, even though backpropagation has a different relative contribution to training time depending on the choice of simulator, it always has a consistent breakdown that is mostly dominated by CUDA API call time, and a small fraction of GPU kernel execution time; the same can be said of inference.

%% file: tex/no_appendix_case_study_scale_up_workload.tex
\subsection{Case Study: Scale-up RL Workload}
\label{sec:minigo}
As shown above in our RL algorithm and simulator survey, an RL workload running on one worker is unable to fully saturate a modern GPU.
One way practitioners try to increase GPU usage is by taking advantage of CPU and GPU hardware parallelism.
RL training workloads feature a data collection phase, where the worker repeatedly switches between inference to decide on an action to take and running the simulator on the chosen action.
Since the model is not being updated during data collection, one common way to increase GPU usage is to have multiple workers collecting data at the same time with the hope that inference minibatches will be concurrently processed by the GPU to keep it occupied.

To decide whether a given number of parallel workers fully occupies the GPU, developers resort to overly coarse-grained tools (e.g., \texttt{nvidia-smi}) that report a ``GPU utilization'' metric, and increase the number of parallel workers until it reaches $100\%$.

\rlscope reveals that coarse-grained tools misleadingly report $100\%$ utilization even though the total GPU usage is, in fact, extremely low.
To illustrate this problem in practice, we use \rlscope to profile the \minigo workload, a popular re-implementation of AlphaGoZero~\cite{nature-agz} adopted by the MLPerf training benchmark suite~\cite{mattson2019mlperf}. 
\minigo parallelizes data collection by having 16 ``self-play'' worker processes play games of Go in parallel, thereby overlapping inference minibatches when evaluating board game states.
\ifx\PaperFormatWithAppendix\undefined
\else
For brevity, we summarize the finding and supporting metrics provided by \rlscope; figures containing these metrics and additional details of the \minigo training loop can be found in Appendix~\ref{sec:app-minigo}.
\fi

\begin{rlscope-finding}{find:gpu-util}
The GPU utilization reported by \texttt{nvidia-smi} for short inference tasks, common in RL workloads, gives a drastically inflated measure of GPU use.
In reality, the workload is hardly using the GPU. Scaling-up by running more workers can exacerbate this issue.
\end{rlscope-finding}

\begin{figure}[!t]
\center{\includegraphics[width=0.9\columnwidth]{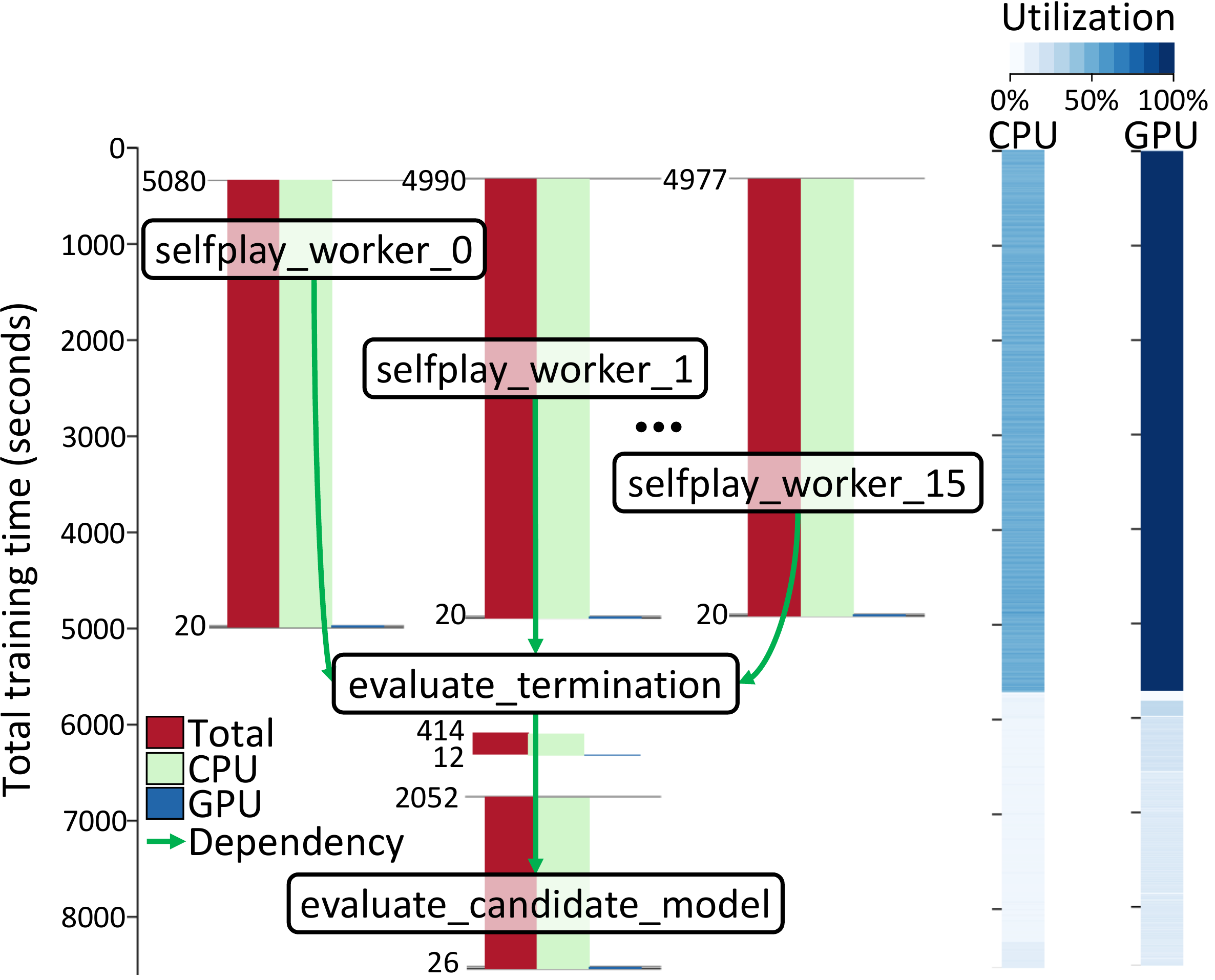}}
\caption{\textit{\minigo\ -- multi-process view:} \texttt{nvidia-smi} misleadingly reports $100\%$ GPU utilization (rightmost bar), even though \rlscope's GPU execution time metric reveals low GPU usage in each self-play worker (20 seconds).
}
\label{fig:06_comp_graph}
\end{figure}

Figure~\ref{fig:06_comp_graph} shows \rlscope's multi-process view of training time for one round of the \minigo training loop: data collection through self-play across 16 parallel worker processes (\texttt{selfplay\_worker\_*}), followed by a model update and evaluation. 
Each parallel self-play worker process appears as a separate ``node'' (barplot) in the ``computational graph'', with dependencies generated from process fork/join relationships.
\rlscope's execution time metrics reveal that \texttt{selfplay\_worker\_0} contributes the most to training time, taking 5080 seconds in total; however, only a meager 20 seconds are spent executing GPU kernels, with the other workers having nearly identical breakdowns.
Conversely, \texttt{nvidia-smi} misleadingly reports $100\%$ GPU utilization during parallel data collection.

The official documentation for \texttt{nvidia-smi}~\cite{nvidia_smi} states that GPU utilization is measured by looking to see if one or more kernels are executing over the sample period, and that the sample period is between $1/6$ seconds and $1$ second.  
Hence, if an extremely small kernel runs that does not fully occupy the sample period, that entire sample period will still have $100\%$ GPU utilization. 
Given that RL inference operations are short kernels, but numerous since 16 processes generate them in parallel, we suspect that all sample periods include at least one such short GPU kernel. 

%% file: tex/with_appendix_case_study_scale_up_workload.tex
\subsection{Case Study: Scale-up RL Workload}
\label{sec:minigo}
As shown above, in our RL algorithm and simulator survey, an RL workload run on one worker cannot fully saturate a modern GPU.
One way practitioners try to increase GPU usage is by taking advantage of CPU and GPU hardware parallelism.
RL training workloads feature a data collection phase, where the worker repeatedly switches between inference to decide on an action to take and running the simulator on the chosen action.
Since the model is not being updated during data collection, one common way to increase GPU usage is to have multiple workers collecting data at the same time with the hope that inference minibatches will be concurrently processed by the GPU to keep it occupied.

To decide whether a given number of parallel workers fully occupy the GPU, developers resort to overly coarse-grained tools (e.g., \texttt{nvidia-smi}) that report a ``GPU utilization'' metric, and increase the number of parallel workers until it reaches $100\%$.

\rlscope reveals that coarse-grained tools misleadingly report $100\%$ utilization even though the total GPU usage is, in fact, extremely low.
To illustrate this problem in practice, we use \rlscope to profile the \minigo workload, a popular re-implementation of AlphaGoZero~\cite{nature-agz} adopted by the MLPerf training benchmark suite~\cite{mattson2019mlperf}. 
\minigo parallelizes data collection by having 16 ``self-play'' worker processes play games of Go in parallel, thereby overlapping inference minibatches when evaluating board game states.
\ifx\PaperFormatWithAppendix\undefined
\else
For brevity, we summarize the finding and supporting metrics provided by \rlscope; figures containing these metrics and additional details of the \minigo training loop can be found in Appendix~\ref{sec:app-minigo}.
\fi

\begin{rlscope-finding}{find:gpu-util}
The GPU utilization reported by \texttt{nvidia-smi} for short inference tasks, common in RL workloads, gives a drastically inflated measure of GPU use.
In reality, the workload is hardly using the GPU. Scaling-up by running more workers can exacerbate this issue.
\end{rlscope-finding}

\rlscope's time breakdown metrics reveal that self-play workers take at most
5080 seconds in total, and only a scant 20 seconds of each worker is spent executing GPU kernels.
Conversely, \texttt{nvidia-smi} misleadingly reports $100\%$ GPU utilization during parallel data collection.
\rlscope's metrics reveal the danger of using coarse-grained metrics like GPU utilization for measuring GPU usage.
The official documentation for \texttt{nvidia-smi}~\cite{nvidia_smi} states that GPU utilization is measured by looking to see if one or more kernels are executing over the sample period and that the sample period is between $1/6$ seconds and $1$ second.  
Hence, if an extremely small kernel runs that does not fully occupy the sample period, that entire sample period will still have $100\%$ GPU utilization. 
Given that RL inference operations are short kernels but numerous since 16 processes generate them in parallel, we suspect that all sample periods include at least one such short GPU kernel.

%% file: tex/related_work.tex
\section{Related work}
\label{sec:related-work}

We begin with a high-level comparison of \rlscope to groups of related works; detailed \rlscope comparisons to each work are given below.  
Existing \emph{ML profiling tools} target GPU-bound supervised learning (SL) workloads and limit their analysis to bottleneck layers and kernels.  
In contrast, with \rlscope we observe the majority of the time is spent CPU-bound throughout the software stack, even for operations considered GPU-heavy (e.g., backpropagation).
\emph{GPU profiling tools} collect GPU kernel execution time with some user-level control over scoping, but unlike \rlscope they cannot correct for profiling overhead.
Finally, \emph{distributed RL training and frameworks} attempt to increase GPU usage by parallelizing inference during data collection, but we observe this to be ineffective with \minigo (Section~\ref{sec:minigo}).

\textbf{ML Profiling Tools:} 
In TBD~\cite{tbd}, the authors survey ML training workloads (including one RL workload) by collecting high-level CPU/GPU utilization metrics; 
\rlscope illustrates that GPU utilization can be a misleading indicator of GPU usage and empowers developers to explain why GPU utilization is a poor metric by providing an accurate CPU/GPU time breakdown.
XSP~\cite{li2020xsp} is an ``across-stack'' profiler that combines profiling information from the model/layer/GPU-kernel levels to provide a drill-down interface for exploring neural network accelerator bottlenecks (e.g., a specific kernel in a specific layer) that dominate SL workloads.
The authors avoid some sources of profiling overhead by collecting profiling information from different stack levels at different training iterations, and correlating them into a single view.  
The two key differences with our approach are that (1) XSP limits its timing analysis to bottleneck layers and kernels in GPU-bound SL workloads whereas \rlscope targets RL workloads which are more CPU-bound due to time spent in high-level code, CUDA APIs, simulators, and from excessive \emph{transitions} to ML backends, and (2) \rlscope's calibration approach allows it to correct for additional sources of book-keeping overhead that XSP cannot correct such as inflation in closed-source CUDA API calls; we observe that RL workloads are particularly 
sensitive to profiling overhead and can experience training time inflation of up to \FCAsTimes{\HRDCalibValidRatioInflationMax} (Figure~\ref{fig:13_overhead_correction}).

\textbf{GPU Profiling Tools:}
NVIDIA Nsight systems~\cite{nvidia-nsight-systems} visualizes a timeline of GPU kernels, CUDA API calls, and system calls and allows users to annotate their code with high-level operations using NVTX~\cite{nvidia-nvtx}.
DLProf~\cite{nvidia-dlprof} focuses on deep neural networks by extending Nsight systems to provide layer-wise scoping and recognizing training iterations to enable summary statistics.
\rlscope avoids using both Nsight systems and DLProf since their analysis backend is closed-source, and as a result, cannot provide calibration correction of CPU profiling overhead which is needed for accurately attributing time spent throughout the RL software stack.

\textbf{Distributed RL training and frameworks:}
Distributed RL frameworks~\cite{moritz2018ray,nair2015massively} have been proposed to scale exploring large hyperparameter search spaces to entire clusters of machines.  
These frameworks have been used to reduce the training time of 49 Atari games from 14 days on a single-machine to 6 days on 100 machines by exploiting the embarrassingly parallel nature of some RL training jobs~\cite{nair2015massively}. 
Unfortunately, little attention has been paid to the performance of individual machines within a cluster. 
Today's distributed RL frameworks are built off of pre-existing ML backends (e.g., TensorFlow, PyTorch), and as a result, inherit all of the poor GPU usage characteristics more commonly observed in single-machine frameworks. 
In particular, scaling GPU usage up by parallelizing inference requests is still insufficient to fully occupy GPU resource usage (Section~\ref{sec:minigo}).
In future work, we hope to extend \rlscope to support distributed RL frameworks, though we expect many of the findings to remain consistent with single-machine frameworks.

%% file: tex/conclusion.tex
\section{Conclusions}
\label{sec:conclusions}

While deep Reinforcement Learning (RL) models are notoriously slow to train, the exact reasons for this are not well-understood.
We observe that RL workloads are fundamentally different from supervised learning workloads, meaning current GPU and ML profilers are a poor fit for this setting.

\rlscope is the first profiler specifically designed for RL workloads: 
it explains where time is spent across the software stack by scoping CPU/GPU usage to high-level algorithmic annotations, corrects for profiling overhead, captures language transitions, and supports multiple ML backends and simulators.
Using \rlscope, we studied the effect of RL framework and ML backend on training time, identified bottlenecks caused by inefficient abstractions in RL frameworks, quantified how bottlenecks change as we vary the RL algorithm and simulator,  and showed how coarse-grained GPU utilization metrics can be misleading in 
a popular scale-up RL workload.  
\rlscope is an open-source tool available at \githubURL, and we look forward to enabling future researchers and practitioners to benefit from \rlscope's insights in their own RL code-bases.

%% file: tex/acknowledgements.tex
\section*{Acknowledgements}

We want to thank Xiaodan (Serina) Tan, Anand Jayarajan, and the entire \ecosys for their continued feedback and support during the development of this work. 
We want to thank reviewers for providing valuable constructive feedback for improving the quality of this paper.
We want to thank Jimmy Ba for allowing me to attend their group meetings to better understand important but less talked about challenges facing RL training.
Finally, we want to thank Vector Institute, Natural Sciences and Engineering Research Council of Canada (NSERC), Canada Foundation for Innovation (CFI), and Amazon Machine Learning Research Award for continued financial support that has made this research possible.

%% file: tex/appendix.tex
\clearpage
\section*{Summary of Appendices}
\label{sec:appendix-summary}
In Section~\ref{sec:rlscope-impl}, we provide information essential for developers looking to implement profiling tools analogous to \rlscope.  
In Section~\ref{sec:rlscope-additional-case-studies}, we provide a simulator survey to show how RL training bottlenecks vary with simulator choice, and we provide supporting figures and additional training loop details for the \minigo RL scale-up workload.  
In Section~\ref{sec:app-profiler-overhead-correction}, we provide additional details on how \rlscope corrects for profiling overhead from transparent interception hooks and closed-source CUDA profiling libraries, and show how \rlscope is able to correct within $\pm 16\%$ of profiling overhead.
In Section~\ref{sec:rlscope-artifact-eval}, we link to detailed online artifact evaluation instructions for reproducing some of the figures found in this paper.

\section{\rlscope implementation details}
\label{sec:rlscope-impl}
We provide additional information for developers looking to implement profiling tools analogous to \rlscope.

\subsection{Storing Traces Asynchronously}
\label{sec:storing-traces-async}

To avoid adding profiling overhead as much as possible, \rlscope stores trace files to disk asynchronously, off the critical path of training.  Traces are aggregated in a stand-alone C++ library (\librlscope), are saved to disk once they reach 20MB, and are stored using the Protobuf binary serialization library~\cite{protobuf}. We explicitly avoid collecting and dumping trace files from Python since asynchronous dumping in python requires using the \texttt{multiprocessing} module~\cite{python_multiprocessing}, which is implemented using a process fork. This would lead to undefined behaviour in multi-threaded libraries (e.g., TensorFlow), and would also require serialization of 20MB of data on the critical path. \librlscope allows us to access shared memory multi-threading in C++ without overheads.

\subsection{Avoiding Sampling Profilers} 
\label{sec:avoiding-sampling-profilers}
During development, we considered the possibility of collecting information using a sampling profiler, which aims to minimize profiling overhead by periodically sampling events at regular intervals.  
Examples of sampling profilers for CPU-only workloads include VTune~\cite{vtune} and the Linux \texttt{perf} utility.  Analogously, NVIDIA's CUPTI library has a PC Sampling API~\cite{gpu-pc-sampling} for periodically sampling the GPU-side program-counter of executing GPU kernels.  However after experimenting with the CUPTI PC Sampling API, we chose to avoid sampling profilers due to several shortcomings.
First, GPU PC sampling adds significant overhead: we observed large overheads when using the CUPTI PC Sampling API, which exceeded those of collecting GPU kernel start/end timestamps\footnote{We confirmed with NVIDIA employees on their developer forums that overheads above $2\times$ are to be expected with the CUPTI PC Sampling API~\cite{forums-pc-sampling-slowdown}.}.
Additionally, given that overhead is inevitable,  sampling profilers offer no convenient way to correct for their overhead during offline analysis. 
Lastly, sampling provides incomplete GPU event information, since some GPU kernels execute very quickly ($\leq 1ms$). A sampling profiler would need to account for loss of accuracy due to missed GPU events.

\section{Extended \rlscope Case Studies}
\label{sec:rlscope-additional-case-studies}
We provide an additional case study relevant to RL practitioners, showing how RL training time breakdowns vary significantly with different choices of simulator.
We also provide more details on our case study of scale-up RL workloads (Section~\ref{sec:minigo}), with supporting figures and additional details on the \minigo training loop.

\input{tex/case_study_simulator_survey}
\subsection{Case Study: Scale-up RL Workload}
\label{sec:app-minigo}

First, we provide \rlscope's multi-process analysis of \minigo which reveals that \texttt{nvidia-smi} reports $100\%$ GPU utilization even though \rlscope reveals that overall GPU usage is still poor. 
Finally, for interested readers we provide additional details of the full \minigo training loop.

\subsubsection{\minigo \rlscope Analysis}
\label{sec:app-minigo-analysis}

In \minigo, each parallel self-play worker plays a game of Go against itself (i.e., each player uses identical model weights), with each worker running an inference minibatch of potential board game states to estimate the probability of winning the match $Q(s, a)$ given the current board game state $s$ and move selection $a$.

\begin{figure}[!t]
\center{\includegraphics[width=0.9\columnwidth]{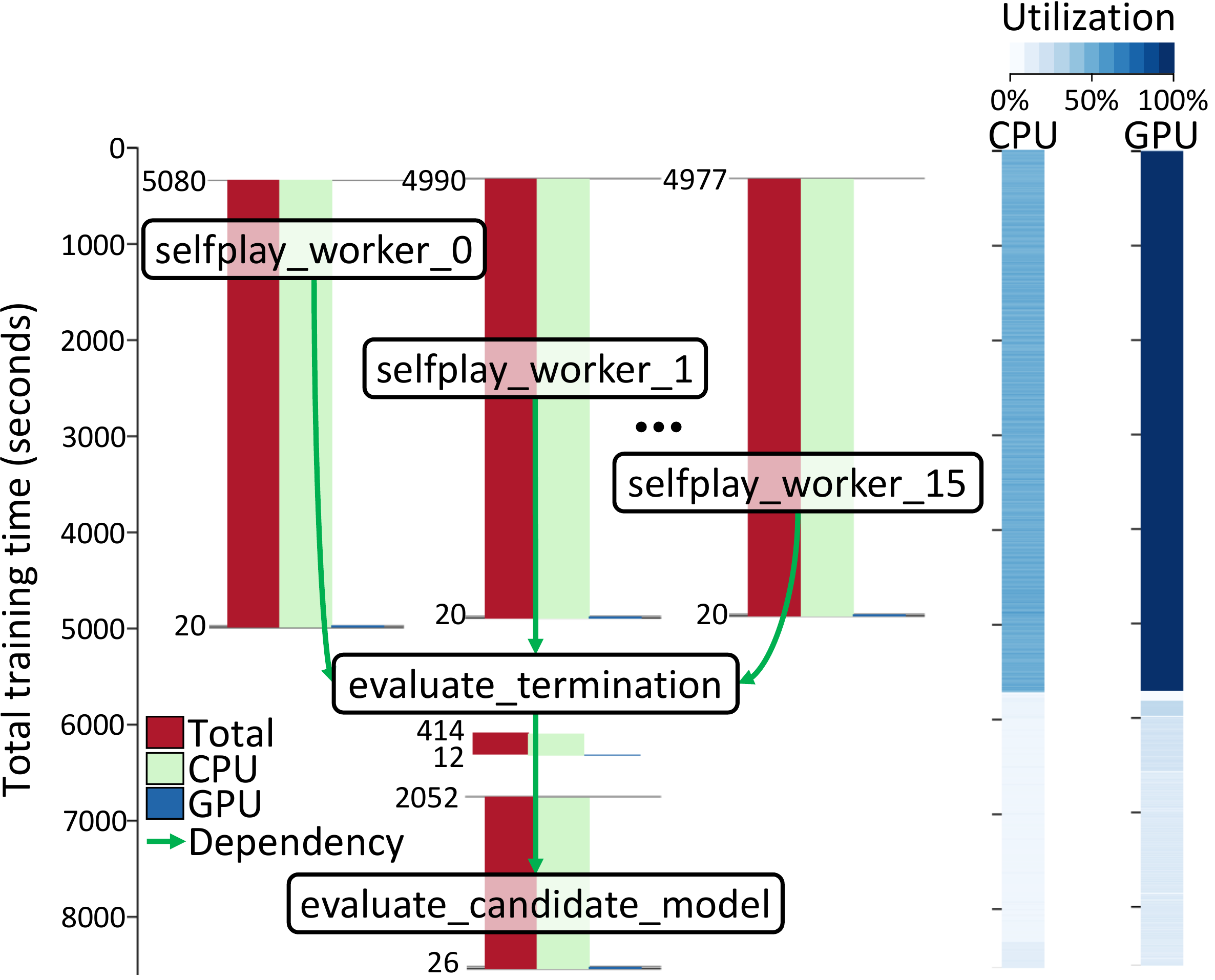}}
\caption{\textit{\minigo\ -- multi-process view:} \texttt{nvidia-smi} misleadingly reports $100\%$ GPU utilization (rightmost bar), even though \rlscope's GPU execution time metric reveals low GPU usage in each self-play worker (20 seconds).
}
\label{fig:06_comp_graph}
\end{figure}

Figure~\ref{fig:06_comp_graph} shows \rlscope's multi-process view of training time for one round of the \minigo training loop: data collection through self-play across 16 parallel worker processes (\texttt{selfplay\_worker\_*}), followed by a model update and evaluation. 
Each parallel self-play worker process appears as a separate ``node'' (barplot) in the ``computational graph'', with dependencies generated from process fork/join relationships.
\rlscope's execution time metrics reveal that \texttt{selfplay\_worker\_0} contributes the most to training time, taking 5080 seconds in total; however, only a meager 20 seconds are spent executing GPU kernels, with the other workers having nearly identical breakdowns.
Conversely, \texttt{nvidia-smi} misleadingly reports $100\%$ GPU utilization during parallel data collection.

\subsubsection{\minigo Training Loop}
\label{sec:app-minigo-train-loop}

A naive computer program for playing Go decides the next move to make by exhaustively expanding all potential sequences of moves until game completion, and following the move that leads to a subtree that maximizes the occurrence of winning outcomes.  
Unfortunately, such an exhaustive expansion is intractable in the game of Go since there are an exponential number of board-game states to explore ($2 \times 10^{170}$ to be precise~\cite{tromp2006combinatorics}).  
Monte-carlo-tree-search (MCTS) provides a strategy for approximating this approach of move-tree expansion; 
the move-tree is truncated using an approximation $Q(s, a)$ of the probability of winning starting from board position $s$ and placing a game-piece at position $a$.  
In order for this MCTS strategy to be effective, we must learn a value function $Q(s, a)$.  The RL approach to this problem is to learn $Q(s, a)$ by collecting moves from games of self-play, and labelling moves with the eventual outcome of each game.

The entire training process for \minigo is split into three distinct \textit{training phases}.  
At the end of the three phases, a new \textit{generation} $i$ of model $Q(s, a)$ is obtained for evaluating Go board game moves.
The phases are repeated a fixed number of times (default $1000$) to obtain successive generations of models, where each generation of model performs as well as or better than the last.  
The training phases are as follows:
\begin{enumerate}
    \item \textbf{Self-play - collect data:} 
    The current generation of model plays against itself for 2000 games.  
    Self-play games can be played in parallel by a pool of worker processes, where the size of the pool is chosen in advance to maximize CPU and GPU hardware utilization (we use $16$ workers, which is the the number of processors on our AMD CPU).

    \item \textbf{SGD-updates - propose candidate model:} 
    data collected during self-play is used to update the $Q(s, a)$ network.

    \item \textbf{Evaluation - choose best model for generation $(i+1)$:} 
    the current model and the candidate model play $100$ games against each other, and the winner becomes the new model for the next generation\footnote{We use the \minigo implementation from MLPerf~\cite{mattson2019mlperf}. 
    The \minigo implementation does not optimize evaluation by parallelizing this portion of training, but we believe there is nothing preventing this optimization in the future.}.

\end{enumerate}

\input{tex/appendix_profiler_overhead_correction}
\section{\rlscope Artifact Evaluation}
\label{sec:rlscope-artifact-eval}
In our open-source documentation, we provide detailed instructions for reproducing some of the figures from this paper in a reproducible docker container.  You can find these instructions here: \url{https://rl-scope.readthedocs.io/en/latest/artifacts.html}.

%% file: tex/appendix_profiler_overhead_correction.tex
\section{Profiler Overhead Correction}
\label{sec:app-profiler-overhead-correction}

As discussed in Section~\ref{sec:profiling-calibration-and-overhead-correction},
profilers inflate CPU-side time due to additional book-keeping code.
To achieve accurate cross-stack critical path measurements, \rlscope corrects for any CPU time inflation during offline analysis by calibrating for the average duration of book-keeping code paths, and subtracting this time at the precise point when it occurs.  Since \rlscope already knows when book-keeping occurs from the information it collects for critical path analysis (Sections~\ref{sec:high-level-algorithmic-annotations}~and~\ref{sec:transparent-event-interception}), the challenge lies in \emph{accurately calibrating} for the average duration of book-keeping code.

\begin{figure}[t]
\centering
\includegraphics[width=0.75\columnwidth]{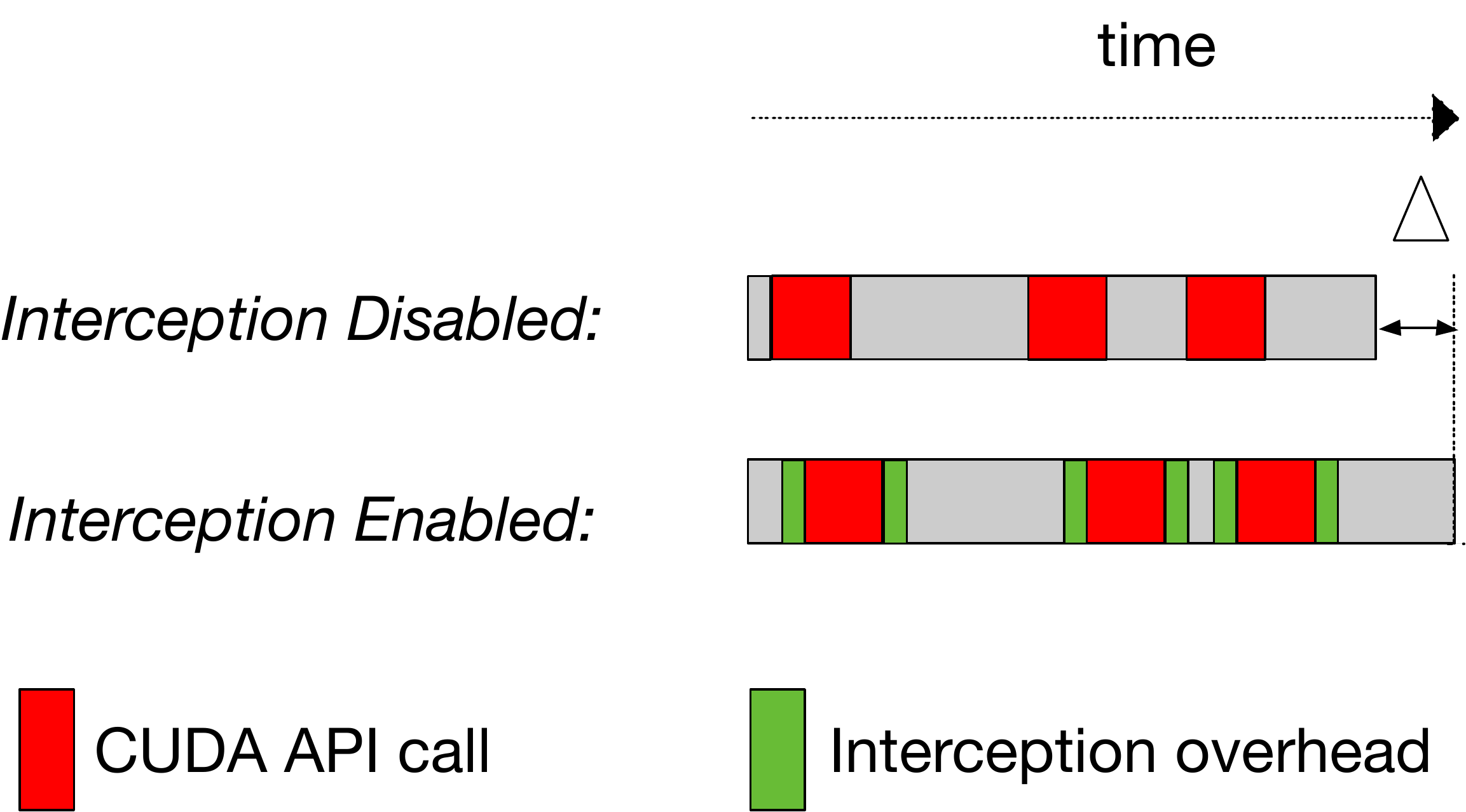}
\caption{\textit{Delta calibration of profiling overhead:}
The difference in runtime, $\Delta$, can be attributed to interception of CUDA API calls. In this example there are 3 CUDA API calls, so the average overhead for calibration is $\Delta / 3$.
}
\label{fig:05_interception_overhead}
\end{figure}

\subsection{Delta Calibration}
\label{sec:isolated-book-keeping-event-calibration}
To measure average book-keeping overhead, we run the training script twice: with book-keeping enabled and disabled.
Assuming that additional runtime ($\Delta$) is strictly due to the enabled book-keeping code\footnote{ML code is designed to be deterministic given the same random seed.}, the average book-keeping time is $\Delta$ divided by the total number of times this book-keeping code executed.
Figure~\ref{fig:05_interception_overhead} illustrates how \rlscope uses this technique to measure the average book-keeping duration of intercepting CUDA API calls. 
Besides CUDA API interception overhead, \rlscope uses delta calibration to correct for high-level algorithm annotations and Python $\leftrightarrow$ C transitions.

\subsection{Difference of Average Calibration}
\label{sec:grouped-book-keeping-event-calibration}

\begin{figure}[t]
\centering
\input{tex/cupti_overhead.tex}
\caption{\textit{Difference of average calibration of profiling overhead:} \texttt{cudaLaunchKernel} and \texttt{cudaMemcpyAsync} experience different CPU time inflations when CUPTI~\cite{cupti} is enabled; we measure precise durations of CUDA API calls to compute book-keeping overheads for each CUDA API.}
\label{fig:04_cupti_overhead}
\end{figure}

As discussed in Section~\ref{sec:profiling-calibration-and-overhead-correction},
not all sources of overhead can be handled by delta calibration. 
Besides CUDA API interception overhead, the CUDA library has internal closed-source profiling code paths that inflate both \texttt{cudaLaunchKernel} and \texttt{cudaMemcpyAsync} API calls by different amounts, and the inflation of each CUDA API cannot be enabled in isolation which is required for delta calibration.  
\rlscope handles this using \textit{difference of average calibration} which measures the average time of individual CUDA APIs with/without profiling enabled, and computes the overhead for each CUDA API using the difference of these averages.  
Figure~\ref{fig:04_cupti_overhead} shows how \rlscope calibrates for different profiling overheads in \texttt{cudaLaunchKernel} ($3\ us$) \texttt{cudaMemcpyAsync} ($1\ us$) induced by enabling CUDA's CUPTI profiling library.

\input{tex/overhead_eval}

%% file: tex/cupti_overhead.tex
\begin{tikzpicture}[every node/.style = {font=\footnotesize, align=left}]

\tikzstyle{timebox} = [draw,minimum height=0.3cm,inner sep=0]
\tikzstyle{launch}=[fill=blue!25, timebox, minimum width=0.6cm]
\tikzstyle{launchprofstyle}=[fill=yellow!50!white, timebox]

\tikzstyle{mem}=[fill=red!50, timebox,minimum width=0.4cm]
\tikzstyle{memprofstyle}=[fill=green!75!blue!50!white, timebox]

\tikzset{launchprof/.pic={
	\draw 
   		++(-0.1,0) node[launch, minimum width=0.4cm] (cl) {}
   		++(0.35,0) node[launchprofstyle, minimum width=0.3cm] {}
   		++(0.25,0) node[launch, minimum width=0.2cm] (cr) {}
   		++(0.1,-0.3) edge [<->] node [midway,below] {\footnotesize #1} ++(-0.9,0);
   	}
}
\tikzset{memprof/.pic={
	\draw 
   		++(-0.15,0) node[mem, minimum width=0.1cm] (cl) {}
   		++(0.1,0) node[memprofstyle, minimum width=0.1cm] {}
   		++(0.2,0) node[mem, minimum width=0.3cm] (cr) {}
   		++(0.15,-0.3) edge [<->] node [midway,below] {\footnotesize #1} ++(-0.5,0);
   	}
}

\node [anchor=west] at (-4.4,1.85) {CUPTI enabled:};
\node [anchor=west] at (-4.4,2.75) {CUPTI disabled:};

\draw  [very thin, fill=gray!20] (-1.2,2.9) rectangle (2.7,2.6);
\draw  [very thin, fill=gray!20] (-1.2,2.0) rectangle (3.5,1.7);

\node [mem, label=below:{\footnotesize 4us}] at (-0.8,2.75) {};
\node [launch, label=below:{\footnotesize 7us}] at (0,2.75) {};
\node [mem, label=below:{\footnotesize 5us}] at (1.4,2.75) {};
\node [launch, label=below:{\footnotesize 6us}] at (2.3,2.75) {};

\draw (-0.8,1.85) pic {memprof=5us};
\draw (0.1,1.85) pic {launchprof=10us};
\draw (1.8,1.85) pic {memprof=6us};
\draw (2.8,1.85) pic {launchprof=9us};

\draw [-stealth,dashed] (-1.2,3.1) -- (3.5,3.1) node [midway, above] {time};

\node[mem, minimum width=0.4cm, label=right:{\texttt{cudaMemcpyAsync}}] at (-4.1,0.8) {};
\node[launch, minimum width=0.4cm, label=right:{\texttt{cudaLaunchKernel}}] at (-4.1,0.4) {};
\node[memprofstyle, minimum width=0.4cm, label=right:{CUPTI overhead for memcpy}] at (-4.1,0.0) {};
\node[launchprofstyle, minimum width=0.4cm, label=right:{CUPTI overhead for launch}] at (-4.1,-0.4) {};

\node [anchor=west] at (0.6,0.8) {\footnotesize (average 4.5us)};
\node [anchor=west] at (0.6,0.4) {\footnotesize (average 6.5us)};
\node [anchor=west] at (0.6,0.0) {\footnotesize (average 1us = 5.5us - 4.5us)};
\node [anchor=west] at (0.6,-0.4) {\footnotesize (average 3us = 9.5us - 6.5us)};
\end{tikzpicture}